%% file: main.tex
\documentclass{article}

\usepackage{microtype}
\usepackage{graphicx}
\usepackage{booktabs} % for professional tables

\usepackage{xcolor}
\usepackage{amsmath}
\usepackage{booktabs}
\usepackage{caption}
\usepackage{subcaption}

\usepackage{hyperref}
\usepackage{url}
\usepackage{multirow}

\hypersetup{colorlinks,
	linkcolor=blue,
	citecolor=blue,
	urlcolor=magenta,
	linktocpage,
	plainpages=false}

\input{tex_zoo/math_commands.tex}

\usepackage[accepted]{icml2020}
\icmltitlerunning{Disentangling Trainability and Generalization in Deep Neural Networks}

\begin{document}

\twocolumn[
\icmltitle{Disentangling Trainability and Generalization in Deep Neural Networks}



\icmlsetsymbol{equal}{*}

\begin{icmlauthorlist}
\icmlauthor{Lechao Xiao}{to}
\icmlauthor{Jeffrey Pennington}{to}
\icmlauthor{Samuel S. Schoenholz}{to}
\end{icmlauthorlist}

\icmlaffiliation{to}{Google Research, Brain Team}
% \icmlaffiliation{goo}{Googol ShallowMind, New London, Michigan, USA}
% \icmlaffiliation{ed}{School of Computation, University of Edenborrow, Edenborrow, United Kingdom}

\icmlcorrespondingauthor{Lechao Xiao}{xlc@google.com}
\icmlcorrespondingauthor{Samuel S. Schoenholz}{schsam@google.com}

\icmlkeywords{Machine Learning, ICML}

\vskip 0.3in
]



\printAffiliationsAndNotice{}  % leave blank if no need to mention equal contribution

\begin{abstract}
A longstanding goal in the theory of deep learning is to characterize the conditions under which a given neural network architecture will be trainable, and if so, how well it might generalize to unseen data. In this work, we provide such a characterization in the limit of very wide and very deep networks, for which the analysis simplifies considerably. For wide networks, the trajectory under gradient descent is governed by the Neural Tangent Kernel (NTK), and for deep networks the NTK itself maintains only weak data dependence. By analyzing the spectrum of the NTK, we formulate necessary conditions for trainability and generalization across a range of architectures, including Fully Connected Networks (FCNs) and Convolutional Neural Networks (CNNs). We identify large regions of hyperparameter space for which networks can memorize the training set but completely fail to generalize. We find that CNNs without global average pooling behave almost identically to FCNs, but that CNNs with pooling have markedly different and often better generalization performance. These theoretical results are corroborated experimentally on CIFAR10 for a variety of network architectures
and we include a \href{https://colab.research.google.com/github/google/neural-tangents/blob/master/notebooks/Disentangling_Trainability_and_Generalization.ipynb}{colab}\footnote{Available at: \href{https://colab.research.google.com/github/google/neural-tangents/blob/master/notebooks/Disentangling_Trainability_and_Generalization.ipynb}{https://tinyurl.com/ybsfxk5y}.} notebook that reproduces the essential results of the paper.
\end{abstract}

\input{tex_zoo/intro.tex}

\section{Related Work}\label{sec:related work}
Recent work \citet{Jacot2018ntk, du2018gradient1, allen2018convergence-fc, du2018gradient, zou2018stochastic} and many others proved global convergence of over-parameterized deep networks by showing that the NTK essentailly remains a constant over the course of training. However, in a different scaling limit the NTK changes over the course of training and global convergence is much more difficult to obtain and is known for neural networks with one hidden layer \citet{mei2018mean,chizat2018global,sirignano2018mean,rotskoff2018neural}. Therefore, understanding the training and generalization properties in this scaling limit remains a very challenging open question.

Another two excellent recent works ~\citep{hayou2019meanfield, jacot2019freeze} also study the dynamics of $\Thetal(x, x')$ for FCNs (and deconvolutions in \cite{jacot2019freeze}) as a function of depth and variances of the weights and biases. \cite{hayou2019meanfield} investigates role of activation functions (smooth v.s. non-smooth) and skip-connection. \cite{jacot2019freeze} demonstrate that batch normalization helps remove the ``ordered phase'' (as in~\cite{yang2019mean}) and a layer-dependent learning rate allows every layer in a network to contribute to learning.

\input{tex_zoo/body.tex}

\input{tex_zoo/theory_experiments.tex}

\section{Conclusion and Future Work}
In this work, we identify several quantities ($\lmax$, $\lrest$, $\kappa$, and $\Deltal$) related to the spectrum of the NTK that control trainability and generalization of deep networks. We offer a precise characterization of these quantities and provide substantial experimental evidence supporting their role in predicting the training and generalization performance of deep neural networks. Future work might extend our framework to other architectures (for example, residual networks with batch-norm or attention architectures). Understanding the role of the nonuniform Fourier modes in the NTK in determining the test performance of CNNs is also an important research direction.

In practice, the correspondence between the NTK and neural networks is often broken due to, e.g., insufficient width, using a large learning rate, or changing the parameterization. Our theory does not directly apply to this setting. As such, developing an understanding of training and generalization away from the NTK regime remains an important research direction.

\section*{Acknowledgements}
We thank Jascha Sohl-dickstein, Greg Yang, Ben Adlam, Jaehoon Lee, Roman Novak and Yasaman Bahri for useful discussions and feedback. 
We also thank anonymous reviewers for feedback that helped improve
the manuscript. 
\nocite{langley00}

\bibliography{reference}
\bibliographystyle{icml2020}


%%%%%%%%%%%%%%%%%%%%%%%%%%%%%%%%%%%%%%%%%%%%%%%%%%%%%%%%%%%%%%%%%%%%%%%%%%%%%%%
%%%%%%%%%%%%%%%%%%%%%%%%%%%%%%%%%%%%%%%%%%%%%%%%%%%%%%%%%%%%%%%%%%%%%%%%%%%%%%%
\onecolumn

\input{tex_zoo/appendix.tex}
\end{document}

%% file: tex_zoo/math_commands.tex
\usepackage{amsmath,amsfonts,bm}

\newcommand{\colord}{{\color{blue}{d}}}

\def\Eqref#1{Equation~\ref{#1}}
\def\eqref#1{\Eqref{#1}}
\def\1{\bm{1}}

\DeclareMathAlphabet{\mathsfit}{\encodingdefault}{\sfdefault}{m}{sl}
\SetMathAlphabet{\mathsfit}{bold}{\encodingdefault}{\sfdefault}{bx}{n}

\newcommand{\R}{\mathbb{R}}

\usepackage{subcaption}
\usepackage{bbm}
\usepackage{bm}
\usepackage{mathtools}
\usepackage{amsthm, amssymb}

\newtheorem{lemma}{Lemma}%[section]

\newtheorem{remark}{Remark}%[section]

\def\1{\bm{1}}

\newcommand{\X}{\mathcal{X}}

\newcommand{\T}{\mathcal{T}}
\newcommand{\K}{\mathcal{K}}

\newcommand{\chic}{\chi_{c^*}}

\newcommand{\N}{\mathcal N}
\newcommand{\bparam}{\mu}

\newcommand{\h}{y}
\newcommand{\hatthetal}{{\hat \Theta}^{(l)}}
\newcommand{\A}{\mathcal A}

\newcommand{\Id}{\textbf{Id}} 
\newcommand{\sw}{\sigma_\omega}
\newcommand{\sws}{\sigma_\omega^2}
\newcommand{\sbb}{\sigma_b}
\newcommand{\sbs}{\sigma_b^2}
\newcommand{\ql}{q^{(l)}}
\newcommand{\pl}{p^{(l)}}
\newcommand{\pll}{p^{(l+1)}}
\newcommand{\qll}{q^{(l+1)}}
\newcommand{\elab}{\epsilon_{ab}^{(l)}}
\newcommand{\dlab}{\delta_{ab}^{(l)}}
\newcommand{\ellab}{\epsilon_{ab}^{(l+1)}}
\newcommand{\dllab}{\delta_{ab}^{(l+1)}}
\newcommand{\elaba}{\epsilon_{ab,\alpha}^{(l)}}
\newcommand{\dlaba}{\delta_{ab, \alpha}^{(l)}}
\newcommand{\ellaba}{\epsilon_{ab, \alpha}^{(l+1)}}
\newcommand{\dllaba}{\delta_{ab, \alpha}^{(l+1)}}
\newcommand{\cstar}{c^{*}}

\newcommand{\pstar}{p^{*}}
\newcommand{\pabstar}{p_{ab}^{*}} 
\newcommand{\pabl}{p^{(l)}_{ab}}
\newcommand{\qabl}{q^{(l)}_{ab}}
\newcommand{\qabstar}{q^{*}_{ab}}
\newcommand{\pabll}{p^{(l+1)}_{ab}}
\newcommand{\qabll}{q^{(l+1)}_{ab}}
\newcommand{\bigo}{\mathcal {O}}
\newcommand{\lmax}{\lambda_{\rm max}}
\newcommand{\lmaxl}{\lambda_{\rm max}^{(l)}}

\newcommand{\lrest}{\lambda_{\rm bulk}}
\newcommand{\lrestl}{\lambda_{\rm bulk}^{(l)}}
\newcommand{\kappal}{\kappa^{(l)}}
\newcommand{\Thetalft}{\Thetal_{\rm flatten}}
\newcommand{\Thetalpool}{\Thetal_{\rm pool}}

\newcommand{\pp}[1]{\left( #1 \right)}

\newcommand{\expect}[2]{\mathbbm{E}_{#1}\left[ #2 \right]}
\newcommand{\finntk}{\hat\Theta}
\newcommand{\infntk}{\Theta}

\newcommand{\Kl}{\mathcal K^{(l)}}
\newcommand{\Klft}{\mathcal K^{(l)}_{\rm flatten}}
\newcommand{\Klpool}{\mathcal K^{(l)}_{\rm pool}}
\newcommand{\Thetalfc}{\Thetal_{\rm fc}}

\newcommand{\Deltal}{P(\Thetal)}
\newcommand{\Kll}{\mathcal K^{(l+1)}}
\newcommand{\Thetal}{ \Theta^{(l)}}
\newcommand{\Thetall}{\Theta^{(l+1)}}

\newcommand{\pab}{p^*_{ab}}

\newcommand{\q}{\quad}

\newcommand{\dT}{\dot \T}
\newcommand{\qstar}{q^{*}} 
 
\newcommand{\Ytrain}{Y_{\text{train}}} 
\newcommand{\gammal}{\gamma^{(L)}}
\newcommand{\NL}{N^{(L)}}
\newcommand{\WL}{W^{(L+1)}}

%% file: tex_zoo/intro.tex
\section{Introduction}

Machine learning models based on deep neural networks have attained state-of-the-art performance across a dizzying array of tasks including vision~\citep{cubuk2019}, speech recognition~\citep{park2019}, machine translation~\citep{bahdanau2014}, chemical property prediction~\citep{gilmer2017}, diagnosing medical conditions~\citep{raghu2019}, and playing games~\citep{silver1140}. Historically, the rampant success of deep learning models has lacked a sturdy theoretical foundation: architectures, hyperparameters, and learning algorithms are often selected by brute force search~\citep{bergstra2012} and heuristics~\citep{glorot2010understanding}. Recently, significant theoretical progress has been made on several fronts that have shown promise in making neural network design more systematic. In particular, in the infinite width (or channel) limit, the distribution of functions induced by neural networks with random weights and biases has been precisely characterized before, during, and after training.

The study of infinite networks dates back to seminal work by~\citet{neal} who showed that the distribution of functions given by single hidden-layer networks with random weights and biases in the infinite-width limit are Gaussian Processes (GPs). Recently, there has been renewed interest in studying random, infinite, networks starting with concurrent work on ``conjugate kernels''~\citep{daniely2016, daniely2017sgd} and ``mean-field theory''~\citep{poole2016exponential,schoenholz2016}. Among numerous contributions, the pair of papers by Daniely \textit{et al.} argued that the empirical covariance matrix of pre-activations becomes deterministic in the infinite-width limit and called this the conjugate kernel of the network. Meanwhile, from a mean-field perspective, the latter two papers studied the properties of these limiting kernels. In particular, the spectrum of the conjugate kernel of wide, fully-connected, networks approaches a well-defined and data-independent limit when the depth exceeds a certain scale, $\xi$. Networks with $\tanh$-nonlinearities (among other bounded activations) exhibit a phase transition between two limiting spectral distributions of the conjugate kernel as a function of their hyperparameters with $\xi$ diverging at the transition. It was additionally hypothesized that networks were un-trainable when the conjugate kernel was sufficiently close to its limit. 

Since then this analysis has been extended to include a wide range for architectures such as convolutions~\citep{xiao18a}, recurrent networks~\citep{chen2018rnn, gilboa2019}, networks with residual connections~\citep{yang2017}, networks with quantized activations~\citep{blumenfeld2019}, the spectrum of the fisher~\citep{karakida2018universal}, a range of activation functions~\citep{hayou2018selection}, and batch normalization~\citep{yang2019mean}. In each case, it was observed that the spectra of the kernels correlated strongly with whether or not the architectures were trainable. While these papers studied the properties of the conjugate kernels, especially the spectrum in the large-depth limit, a branch of concurrent work took a Bayesian perspective: that many networks converge to Gaussian Processes as their width becomes large~\citep{lee2018deep, matthews2018, novak2018bayesian, garrigaalonso2018deep, yang2019scaling}. In this case, the Conjugate Kernel was referred to as the Neural Network Gaussian Process (NNGP) kernel, which is used to train neural networks in a fully Bayesian fashion. As such, the NNGP kernel characterizes performance of the corresponding NNGP.    

Together this work offered a significant advance to our understanding of wide neural networks; however, this theoretical progress was limited to networks at initialization or after Bayesian posterior estimation and provided no link to gradient descent. Moreover, there was some preliminary evidence that suggested the situation might be more nuanced than the qualitative link between the NNGP spectrum and trainability might suggest. For example,~\citet{philipp2017exploding} showed that deep $\tanh$ FCNs could be trained after the kernel reached its large-depth, data-independent, limit but that these networks did not generalize to unseen data. 

Recently, significant theoretical clarity has been reached regarding the relationship between the GP prior and the distribution following gradient descent. In particular,~\citet{Jacot2018ntk} along with followup work~\citep{lee2019wide, chizat2019lazy} showed that the distribution of functions induced by gradient descent for infinite-width networks is a Gaussian Process with a particular compositional kernel known as the Neural Tangent Kernel (NTK). In addition to characterizing the distribution over functions following gradient descent in the wide network limit, the learning dynamics can be solved analytically throughout optimization.

In this paper, we leverage these developments and revisit the relationship between architecture, hyperparameters, trainability, and generalization in the large-depth limit for a variety of neural networks. In particular, we make the following contributions:
\begin{itemize}
    \item {\bf Trainability.} We compute the large-depth asymptotics of several quantities related to trainability, including the largest/smallest eigenvalue of the NTK, $\lambda_{\text{max}/\text{min}}$, and the condition number $\kappa = \lambda_{\text{max}}/\lambda_{\text{min}}$; see Table~\ref{table evolution ntk}. 

    \item {\bf Generalization.} We characterize the {\it mean predictor} $P(\Theta)$, which is intimately related to the prediction of wide neural networks on the test set following gradient descent training. As such, the mean predictor is intimately related to the model's ability to generalize. In particular, we argue that networks fail to generalize if the mean predictor becomes data-independent.
    % We will quantify this breakdown by contracting the mean predictor with mean centered labels $P(\Theta)Y_{\text{train}}$.

    \item We show that the \emph{ordered} and \emph{chaotic} phases identified in~\citet{poole2016exponential} lead to markedly different limiting spectra of the NTK. In the ordered phase the trainability of neural networks degrades at large depths, but their ability to generalize persists. By contrast, in the chaotic phase we show that trainability improves with depth, but generalization degrades and neural networks behave like hash functions.
    
    A corollary of these differences in the spectra is that, as a function of depth, the optimal learning rates ought to decay exponentially in the chaotic phase, linearly on the order-to-chase trainsition line, and remain roughly a constant in the ordered phase.
    \item We examine the differences in the above quantities for fully-connected networks (FCNs) and convolutional networks (CNNs) with and without pooling and precisely characterize the effect of pooling on the interplay between trainability, generalization, and depth.
\end{itemize}

In each case, we provide empirical evidence to support our theoretical conclusions. Together these results provide a complete, analytically tractable, and dataset-independent theory for learning in very deep and wide networks. Philosophically, we find that trainability and generalization are distinct notions that are, at least in this case, at odds with one another. Indeed, good conditioning of the NTK (which is a necessary condition for training) seems necessarily to lead to poor generalization performance. It will be interesting to see whether these results carry over in shallower and narrower networks. The tractable nature of the wide and deep regime leads us to conclude that these models will be an interesting testbed to investigate various theories of generalization in deep learning.

%%%%%%%%%%%%%%%%%%%%%%%%%%%%%%%%%%%%%%%%%%%%%%%%%%%%%%%%%%%%%%%%%%%%%%%%%%%%%%%%%%%%%%%%%%%%%%%%%%%%%%%%%%%%%%%%%%%%%%%%%%%%%%%%%%%%%%%%%%%%%%%%%%%%%%%%%%%%%%%%%%%%%%%%%%%%%%%%%%%%%%%%%%%%%%%%%%%%%%%%%%%%%%
\begin{table}[t]
\begin{center}
\resizebox{\columnwidth}{!}{%
\begin{tabular}{llll}
        \toprule
                  & \multicolumn{3}{c}{{\color[HTML]{000000} NTK $\Thetal$ of  FC/CNN-F, \quad  \color{blue}{CNN-P}}}\\[0.1cm]  
        \midrule
\multirow{-2}{*}{}  & Ordered     $\chi_1 <1$       & Critical $\chi_1=1$            & Chaotic $\chi_1 > 1$            \\ [0.4cm]
                 $ \lmaxl$                 &   $m\pstar +m \bigo (l\chi_1^l)$        &       $\frac {m{\colord} +2} {3\colord} l\qstar   + m \bigo(1)$         &      $\Theta(\chi_1^l)\color{blue}{/}\colord$  \\ [0.4cm]
                  $\lrestl$                 &  $ \bigo(l\chi_1^l)\color{blue}{/}\colord$          &     $\frac 2 {3 \colord } l\qstar  + \frac{1}{\colord}\bigo(1)$          &            $\Theta(\chi_1^l)\color{blue}{/}\colord$                  \\ [0.4cm]
                  $\kappal$                 &   $\colord m\pstar \Omega(\chi_1^{-l} / l)$         
                  &    $\frac {m{\colord} +2} 2 + \colord  m\bigo(l^{-1})$            & $1 + \bigo({\colord}\chi_1^{-l})$                
                  \\ [0.4cm]
                  $P(\Thetal)Y_{\text{train}}$                 &   $ \bigo(1)$         
                  &    $\colord  \bigo(l^{-1})$            
                  & $\colord \bigo(l (\chi_{c^*}/\chi_1)^l  )$            
                  \\
        \bottomrule
\end{tabular}% 
} 
\end{center}
\captionsetup{singlelinecheck=off}
    \caption{\textbf{Evolution of the NTK spectra and $\Deltal$ as a function of depth $l$.} The NTKs of FCN and CNN without pooling (CNN-F) are essentially the same and the scaling of $\lmaxl$, $\lrestl$, $\kappal$, and $\Delta^{(l)}$ for these networks is written in black. Corrections to these quantities due to the addition of an average pooling layer ({\color{blue} CNN-P}) with window size $\colord$ is written in blue.}
    \label{table evolution ntk}
\end{table}

%%%%%%%%%%%%%%%%%%%%%%%%%%%%%%%%%%%%%%%%%%%%%%%%%%%%%%%%%%%%%%%%%%%%%%%%%%%%%%%%%%%%%%%%%%%%%%%%%%%%%%%%%%%%%%%%%%%%%%%%%%%%%%%%%%%%%%%%%%%%%%%%%%%%%%%%%%%%%%%%%%%%%%%%%%%%%%%%%%%%%%%%%%%%%%%%%%%%%%%%%%%%%%

%% file: tex_zoo/body.tex
\section{Background}\label{sec:background}
We summarize recent developments in the study of wide random networks. We will keep our discussion relatively informal; see e.g. ~\citep{novak2018bayesian} for a more rigorous version of these arguments. To simplify this discussion and as a warm-up for the main text, we will consider the case of FCNs. Consider a fully-connected network of depth $L$ where each layer has a width $N^{(l)}$ and an activation function $\phi:\mathbb R\to\mathbb R$. In the main text we will restrict our discussion to $\phi=\text{erf}$ or $\tanh$ for clarity, however we include results for a range of architectures including $\phi= \text{ReLU}$ with and without skip connections and layer normalization in the supplementary material (see Sec.~\ref{sec:RElu-networks}). We find that the high level picture described here applies to a wide range of architectural components, though important specifics - such as the phase diagram - can vary substantially. For simplicity, we will take the width of the hidden layers to infinity sequentially: $N^{(1)} \to\infty, \dots, N^{(L-1)} \to\infty$. The network is parameterized by weights and biases that we take to be randomly initialized with $W_{ij}^{(l)}, b_i^{(l)}\sim\mathcal N(0, 1)$ along with hyperparameters, $\sigma_w$ and $\sigma_b$ that set the scale of the weights and biases respectively. Letting the $i$\textsuperscript{th} pre-activation in the $l$\textsuperscript{th} layer due to an input $x$ be given by $z^{(l)}_i(x)$, the network is then described by the recursion, for $0\leq l \leq L-1$, 
\begin{equation}\label{eq:fc_def_recap_recursion}
    z^{(l+1)}_i(x) = \frac{\sigma_w}{\sqrt {N^{(l)}}}\sum_{j=1}^{N^{(l)}}W^{(l+1)}_{ij}\phi(z^{(l)}_j(x)) + \sigma_b b^{(l+1)}_i 
\end{equation}
Notice that as $N^{(l)}\to\infty$, the sum ends up being over a large number of random variables and we can invoke the central limit theorem to conclude that the $\{z_i^{(l+1)}\}_{i\in [N^{(l+1)}]}$ are i.i.d. Gaussian with zero mean. Given a dataset of $m$ points, the distribution over pre-activations can therefore be described completely by the covariance matrix, i.e. the NNGP kernel, between neurons in different inputs $\Kl(x,x') = \mathbb E[z^{(l)}_{i}(x)z^{(l)}_{i}(x')].$ Inspecting~\eqref{eq:fc_def_recap_recursion}, we see that $\Kll$ can be computed in terms of $\Kl$ as 
\begin{align}\label{eq:fc_mft_recap_recursion-main}
    \Kll(x, x') 
    % &= \sigma_w^2\mathbb E_{(z,z')\sim\mathcal N(0,\Kl(x, x'))}[\phi(z)\phi(z')] + \sigma_b^2
    % \nonumber
     &\equiv \sigma_w^2 \mathcal T(\Kl)(x, x') + \sigma_b^2 \,
     \\
     \mathcal T(\K) & \equiv \mathbb E_{z\sim\mathcal N(0, \K)}[\phi(z)\phi(z)^T]
\end{align}
% Here $\T$ is operator from the space of positive semi-definite matrices to itself. 

\eqref{eq:fc_mft_recap_recursion-main} describes a dynamical system on positive semi-definite matrices $\mathcal K$. It was shown in~\citet{poole2016exponential} that fixed points, $\mathcal K^*(x,x')$, of these dynamics exist such that $\lim_{l\to\infty}\Kl(x,x')=\mathcal K^*(x,x')$ with $\mathcal K^*(x,x') = q^*[\delta_{x,x'} + c^*(1-\delta_{x,x'})]$ independent of the inputs $x$ and $x'$. The values of $q^*$ and $c^*$ are determined by the hyperparameters, $\sigma_w$ and $\sigma_b$. However ~\eqref{eq:fc_mft_recap_recursion-main} admits multiple fixed points (e.g. $c^* = 0, 1$) and the stability of these fixed points plays a significant role in determining the properties of the network. Generically, there are large regions of the $(\sigma_w,\sigma_b)$ plane in which the fixed-point structure is constant punctuated by curves, called phase transitions, where the structure changes; see Fig~\ref{fig:phase-diagram} for $\tanh$-networks.  

The rate at which $\mathcal K(x, x')$ approaches or departs $\mathcal K^*(x, x')$ can be determined by expanding \eqref{eq:fc_mft_recap_recursion-main} about its fixed point, $\delta\mathcal K(x, x') = \mathcal K(x, x') - \mathcal K^*(x, x')$ to find
% \footnote{More precisely, one needs to consider the Jacobian of $\T$ as an operator from positive semi-definite matrices to positive semi-definite matrices. We refer the readers to Section B of \citet{xiao18a} for more details.}
\begin{equation}
    \delta\Kll(x, x') \approx \sigma_w^2\dot{\mathcal T}(\mathcal K^*(x, x'))\delta\Kl(x, x')
\end{equation}
with $\dot{\mathcal T}(\K) = \mathbb E_{(z_1, z_2)\sim \mathcal N(0, \K)}[\dot\phi(z_1)\dot\phi(z_2)]$ and $\dot \phi$ is the derivative of $\phi$. This expansion naturally exhibits exponential convergence to - or divergence from - the fixed-point as $\delta\mathcal K^{(l)}(x, x')\sim \chi(x, x')^l$ where $\chi(x,x') = \sigma_w^2\dot{\mathcal T}(\mathcal K^*(x, x'))$. Since $\mathcal K^*(x, x')$ does not depend on $x$ or $x'$ it follows that $\chi(x, x')$ will take on a single value, $\chi_{c^*}$, whenever $x\neq x'$. If $\chi_{c^*} < 1$ then this $\mathcal{K^*}$ fixed point is stable, but if $\chi_{c^*} > 1$ then the fixed point is unstable and, as discussed above, the system will converge to a different fixed point. If $\chi_{c^*} = 1$ then the hyperparameters lie at a phase transition and convergence is non-exponential. As was shown in~\citet{poole2016exponential}, there is always a fixed-point at $c^* = 1$ whose stability is determined by $\chi_1$. This is the so-called ordered phase since any pair of inputs will converge to identical outputs. The line defined by $\chi_1=1$ defines the order-to-chaos transition separating the ordered phase from the ``chaotic'' phase (where $c^* > 1$). Note, that $\chi_{c^*}$ can be used to define a depth-scale, $\xi_{c^*} = -1 / \log(\chi_{c^*})$ that describes the number of layers over which $\Kl$ approaches $\mathcal K^*.$

This provides a precise characterization of the NNGP kernel at large depths. As discussed above, recent work~\citep{Jacot2018ntk, lee2019wide, chizat2019lazy} has connected the prior described by the NNGP with the result of gradient descent training using a quantity called the NTK. To construct the NTK, suppose we enumerate all the parameters in the fully-connected network described above by $\theta_\alpha$. The finite width NTK is defined by $\finntk(x, x') = J(x)J(x')^T$ where $J_{i\alpha}(x) = \partial_{\theta_\alpha}z^L_i(x)$ is the Jacobian evaluated at a point $x$. The main result in~\citet{Jacot2018ntk} was to show that in the infinite-width limit, the NTK converges to a deterministic kernel $\infntk$ and remains constant over the course of training. As such, at a time $t$ during gradient descent training with an MSE loss, the expected outputs of an infinitely wide network, $\mu_t(x) = \mathbb E[z^L_i(x)]$, evolve as
\begin{align}\label{eq:fc_ntk_recap_dynamics}
    \mu_t(X_{\text{train}}) &= (\Id - e^{-\eta\infntk_{\text{train, train}}t})Y_{\text{train}}
    \\
    \mu_t(X_{\text{test}}) &= \infntk_{\text{test, train}}\infntk_{\text{train, train}}^{-1}(\Id - e^{-\eta\infntk_{\text{train, train}}t})Y_{\text{train}}
    \label{eq:ntk-predictor} 
% \normalsize 
\end{align}
\normalsize
for train and test points respectively; see Section 2 in \citet{lee2019wide}. Here $\Theta_{\text{test, train}}$ denotes the NTK between the test inputs $X_{\text{test}}$ and training inputs $X_{\text{train}}$ and $\Theta_{\text{train, train}}$ is defined similarly. 
Since $\finntk$ converges to $\infntk$ as the network's width approaches infinity, the gradient flow dynamics of real network also converge to the dynamics described by \eqref{eq:fc_ntk_recap_dynamics} and \eqref{eq:ntk-predictor}
\citep{Jacot2018ntk, lee2019wide, chizat2019lazy, yang2019scaling, arora2019exact,huang2019dynamics}.  
As the training time, $t$, tends to infinity we note that these equations reduce to $\mu(X_{\text{train}}) = Y_{\text{train}}$ and $\mu(X_{\text{test}}) = \Theta_{\text{test, train}}\Theta_{\text{train, train}}^{-1} Y_{\text{train}}$. Consequently we call
\begin{align}\label{eq:linear-predictor} 
    P(\Theta) \equiv \Theta_{\text{test, train}}\Theta_{\text{train, train}}^{-1}
\end{align}
 the ``mean predictor''. We can also compute the mean predictor of the NNGP kernel, $P(\K)$, which analogously can be used to find the mean of the posterior after Bayesian inference. We will discuss the connection between the mean predictor and generalization in the next section.
 
 In addition to showing that the NTK describes networks during gradient descent, \citet{Jacot2018ntk} showed that the NTK could be computed in closed form in terms of $\mathcal T$, $\dot{\mathcal T}$, and the NNGP as,
\begin{equation}\label{eq: NTK-recursive}
 \scalebox{.9}{
$    \Thetall(x, x') = \Kll(x, x') + \sigma_w^2\dot{\T}(\Kl)(x, x')\Thetal(x, x')\,$.}
\end{equation}
where $\Thetal$ is the NTK for the pre-activations at layer-$l$. 
        \section{Metrics for Trainability and Generalization at Large Depth}
        We begin by discussing the interplay between the conditioning of $\Theta_{\text{train, train}}$ and the trainability of wide networks. We can write \eqref{eq:fc_ntk_recap_dynamics} in terms of the spectrum of $\Theta_{\text{train, train}}$. To do this we write the eigendecomposition of $\Theta_{\text{train, train}}$ as $\Theta_{\text{train, train}} = U^TDU$ with $D$ a diagonal matrix of eigenvalues and $U$ a unitary matrix. In this case \eqref{eq:fc_ntk_recap_dynamics} can be written as,
        \begin{equation}\label{eq:fc_ntk_dynamics_eigen}
        \tilde\mu_t(X_{\text{train}})_i = (\text{Id} - e^{-\eta \lambda_i t})\tilde Y_{\text{train},i}   
        \end{equation}
        where $\lambda_i$ are the eigenvalues of $\Theta_{\text{train, train}}$ and $\tilde\mu_t(X_{\text{train}}) = U\mu_t(X_{\text{train}})$, $\tilde Y_{\text{train}} = U Y_{\text{train}}$ are the mean prediction and the labels respectively written in the eigenbasis of $\Theta_{\text{train,train}}$. If we order the eigenvalues such that $\lambda_0 \geq \cdots \geq \lambda_m$ then it has been hypothesized\footnote{For finite width, the optimization problem is non-convex and there are not rigorous bounds on the maximum learning rate.} in e.g.~\citet{lee2019wide} that the maximum feasible learning rate scales as $\eta \sim 2 / \lambda_0$ as we verify empirically in section 4. Plugging this scaling for $\eta$ into \eqref{eq:fc_ntk_dynamics_eigen} we see that the smallest eigenvalue will converge exponentially at a rate given by $1/\kappa$, where $\kappa =  \lambda_0 / \lambda_m$ is the condition number.  It follows that if the condition number of the NTK associated with a neural network diverges then it will become untrainable and so we use $\kappa$ as a metric for trainability. 
        
        We will see that at large depths, the spectrum of $\Theta_{\text{train, train}}$ typically features a single large eigenvalue, $\lambda_{\text{max}}$, and then a gap that is large compared with the rest of the spectrum. We therefore will often refer to a typical eigenvalue in the bulk as $\lrest$ and approximate the condition number as $\kappa = \lambda_{\text{max}} / \lrest$.

        We now turn our attention to generalization. At large depths, we will see that $\Thetal_{\text{test, train}}$ and $\Thetal_{\text{train, train}}$ converge their fixed points independent of the data distribution. Consequently it is often the case that $P(\Theta^*)$ will be data-independent and the network will fail to generalize. In this case, by symmetry, it is necessarily true that $P(\Theta^*)$ will be a constant matrix. Contracting this matrix with a vector of labels $Y_{\text{train}}$ that have been standardized to have zero mean it will follow that $P(\Theta^*)Y_{\text{train}} = 0$ and the network will output zero in expectation on all test points. Clearly, in this setting the network will not be able to generalize. At large, but finite, depths the generalization performance of the network can be quantified by considering the rate at which $P(\Thetal)Y_{\text{train}}$ decays to zero. There are cases, however, where despite the data-independence of $\Theta^*$, $\lim_{l\to\infty} P(\Thetal)Y_{\text{train}}$ remains nonzero and the network can continue to generalize even in the asymptotic limit. In either case, we will show that precisely characterizing $P(\Thetal)Y_{\text{train}}$ allows us to understand exactly where networks can, and cannot, generalize.

        Our goal is therefore to characterize the evolution of the two metrics $\kappal$ and $\Deltal$ in $l$. We follow the methodology outlined in~\citet{schoenholz2016, xiao18a} to explore the spectrum of the NTK as a function of depth. We will use this to make precise predictions relating trainability and generalization to the hyperparameters $(\sigma_w, \sigma_b, l)$. Our main results are summarized in Table \ref{table evolution ntk} which describes the evolution of $\lmaxl$ (the largest eigenvalue of $\Thetal$), $\lrestl$ (the remaining eigenvalues), $\kappal$, and $\Deltal$ as a function of depth for three different network configurations (the ordered phase, the chaotic phase, and the phase transition). We study the dependence on: the size of the training set, $m$; the choices of architecture including fully-connected networks (FCN), convolutional networks with flattening (CNN-F), and convolutions with pooling (CNN-P); and the size, $d$, of the window in the pooling layer (which we always take to be the penultimate layer).
        
        Before discussing the methodology it is useful to first give a qualitative overview of the phenomenology. We find identical phenomenology between FCNs and CNN-F architectures. In the ordered phase, $\Thetal\to \pstar \bm1 \bm1^T$, $\lmaxl\to mp^*$ and $\lrestl=\bigo (l\chi_1^l)$. At large depths since $\chi_1 < 1$ it follows that $\kappal\gtrsim mp^* / (l\chi_1^l)$ and so the condition number diverges exponentially quickly. Thus, in the ordered phase we expect networks not to be trainable (or, specifically, the time they take to learn will grow exponentially in their depth). Here $\Deltal$ converges to a data dependent constant independent of depth; thus, in the ordered phase networks fail to train but can generalize indefinitely. 
        
        By contrast, in the chaotic phase we see that there is no gap between $\lmaxl$ and $\lrestl$ and networks become perfectly conditioned and are trainable everywhere.  However, in this regime we see that the mean predictor scales as $l (\chi_{c^*} / \chi_1)^l$. Since in the chaotic phase $\chi_{c^*} < 1$ and $\chi_1 > 1$ it follows that $\Deltal\to 0$ over a depth $\xi_{*} = -1/\log(\chi_{c^*} / \chi_1)$. Thus, in the chaotic phase, networks fail to generalize at a finite depth but remain trainable indefinitely. Finally, introducing pooling modestly augments the depth over which networks can generalize in the chaotic phase but reduces the depth in the ordered phase. We will explore all of these predictions in detail in section~\ref{sec:experiments}.

%% file: tex_zoo/theory_experiments.tex
%%%%%%%%%%%%%%%%%%%%%%%%%%%%%%%%%%%%%%%%%%%%%%%%%%%%%%%%%%%%%%%%%%%%%%%%%%%%%%%%%%%%%%%%%%%%%%%%%%%%%%%%%%%%%%%%%%%%%%%%%%%%%%%%%%%%%%%%%%%%%%%%%%%%
%%%% condition number plots 

\begin{figure*}[t] 
\begin{center}
         \begin{subfigure}[b]{.33\textwidth}
                \centering
\includegraphics[width=1.\textwidth]{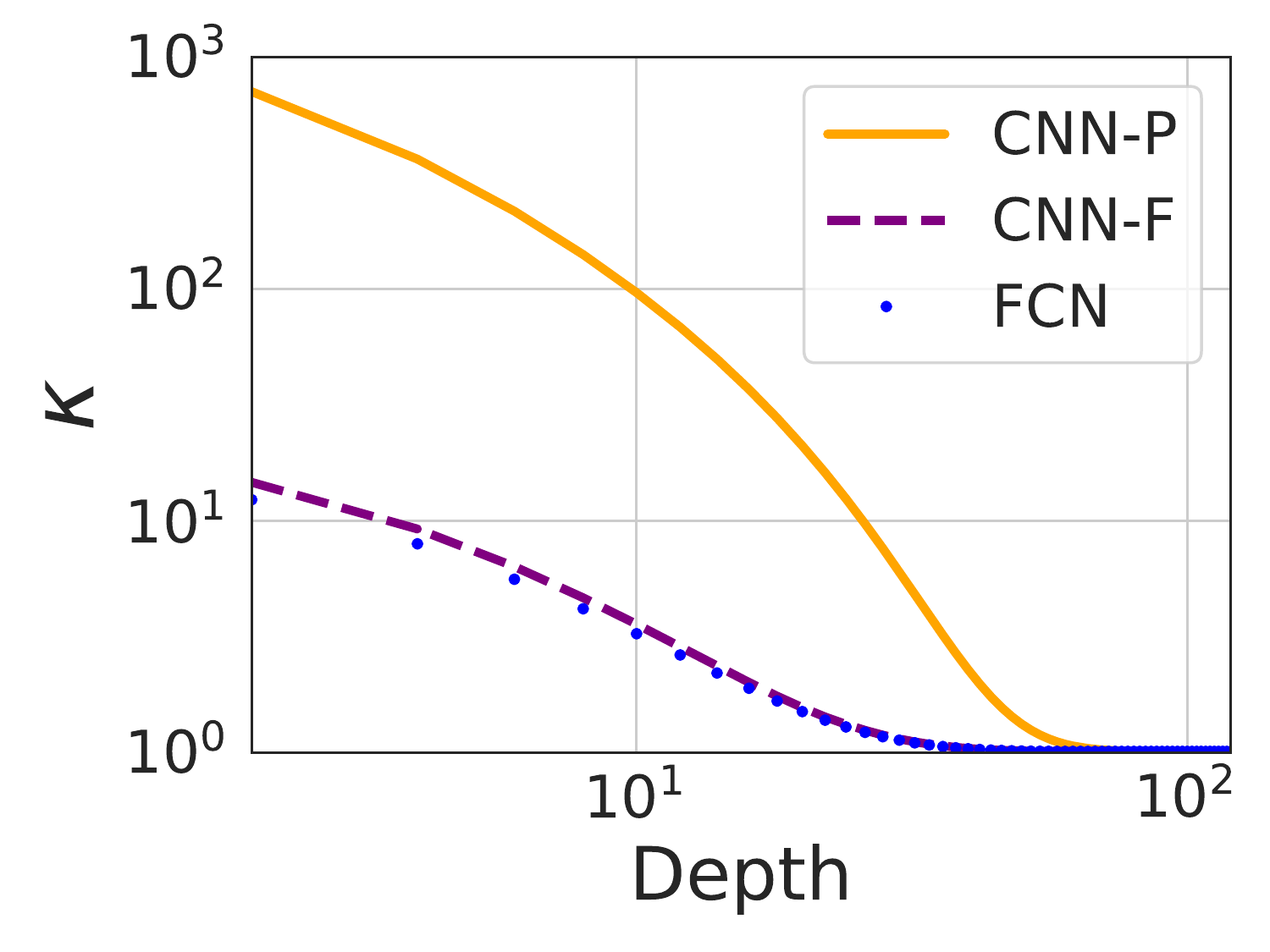}
                \caption{Chaotic}
                \label{fig:kappa-chaotic}
        \end{subfigure}
        \begin{subfigure}[b]{0.33\textwidth}
                \centering
               \includegraphics[width=1\textwidth]{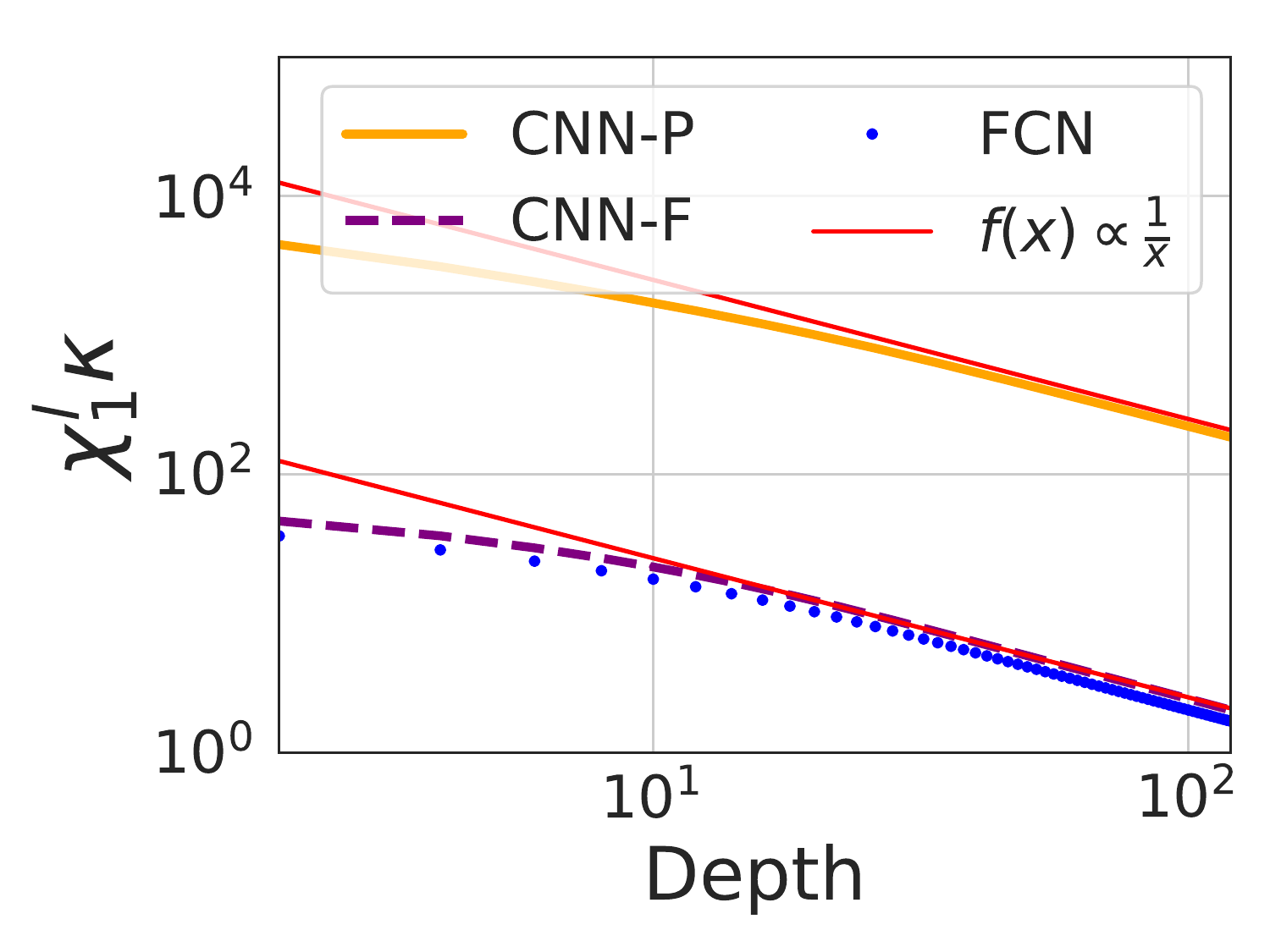}
                \caption{Ordered}
                \label{fig:kappa-order}
        \end{subfigure}%
        \begin{subfigure}[b]{0.33\textwidth}
                \centering
\includegraphics[width=1.\textwidth]{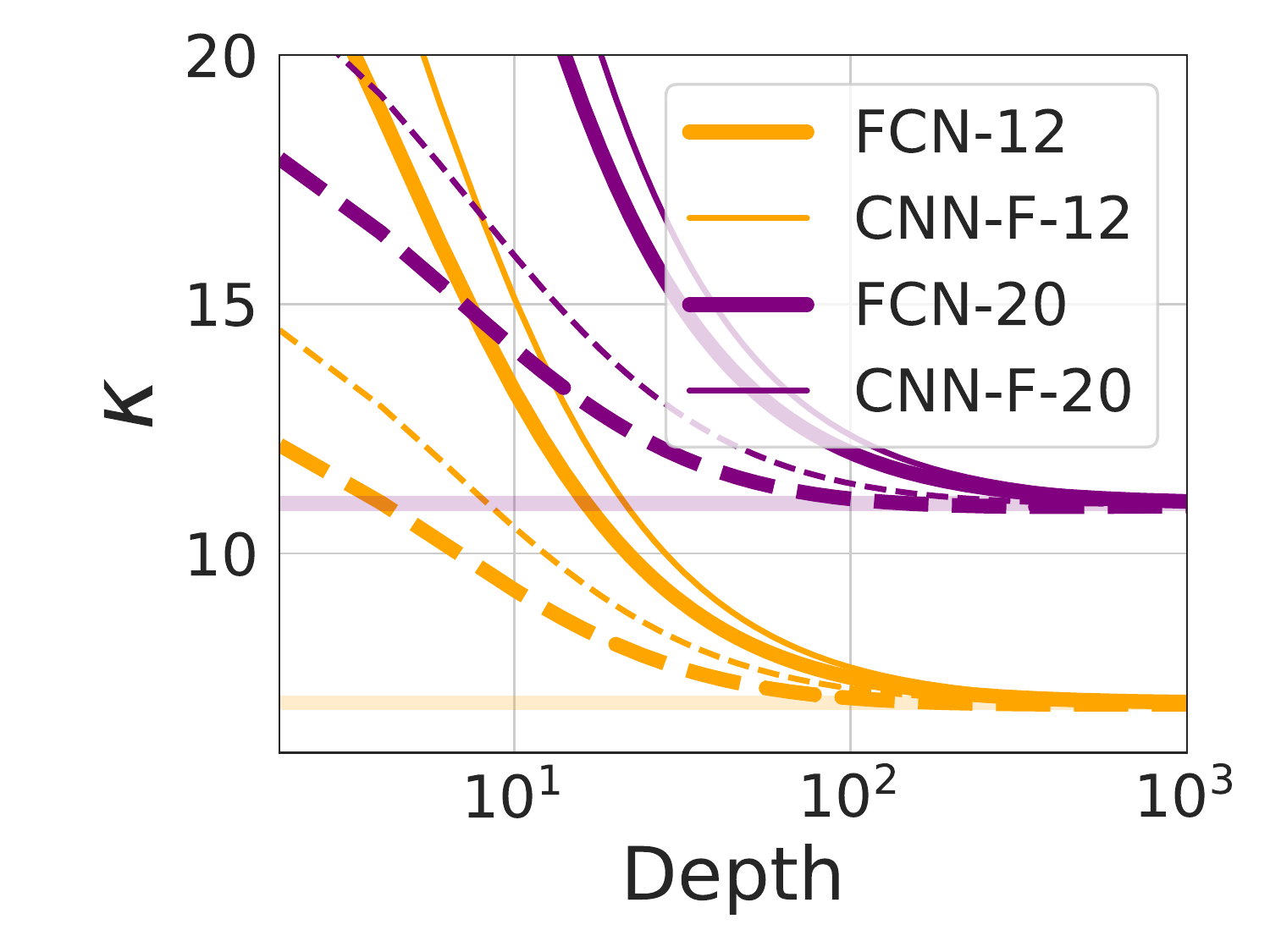}
                \caption{Critical: FCN and CNN-F}
                \label{fig:kappa-critical}
        \end{subfigure}%
        \\ 
 \begin{subfigure}[b]{0.33\textwidth}
                \centering
     \includegraphics[width=1.\textwidth]{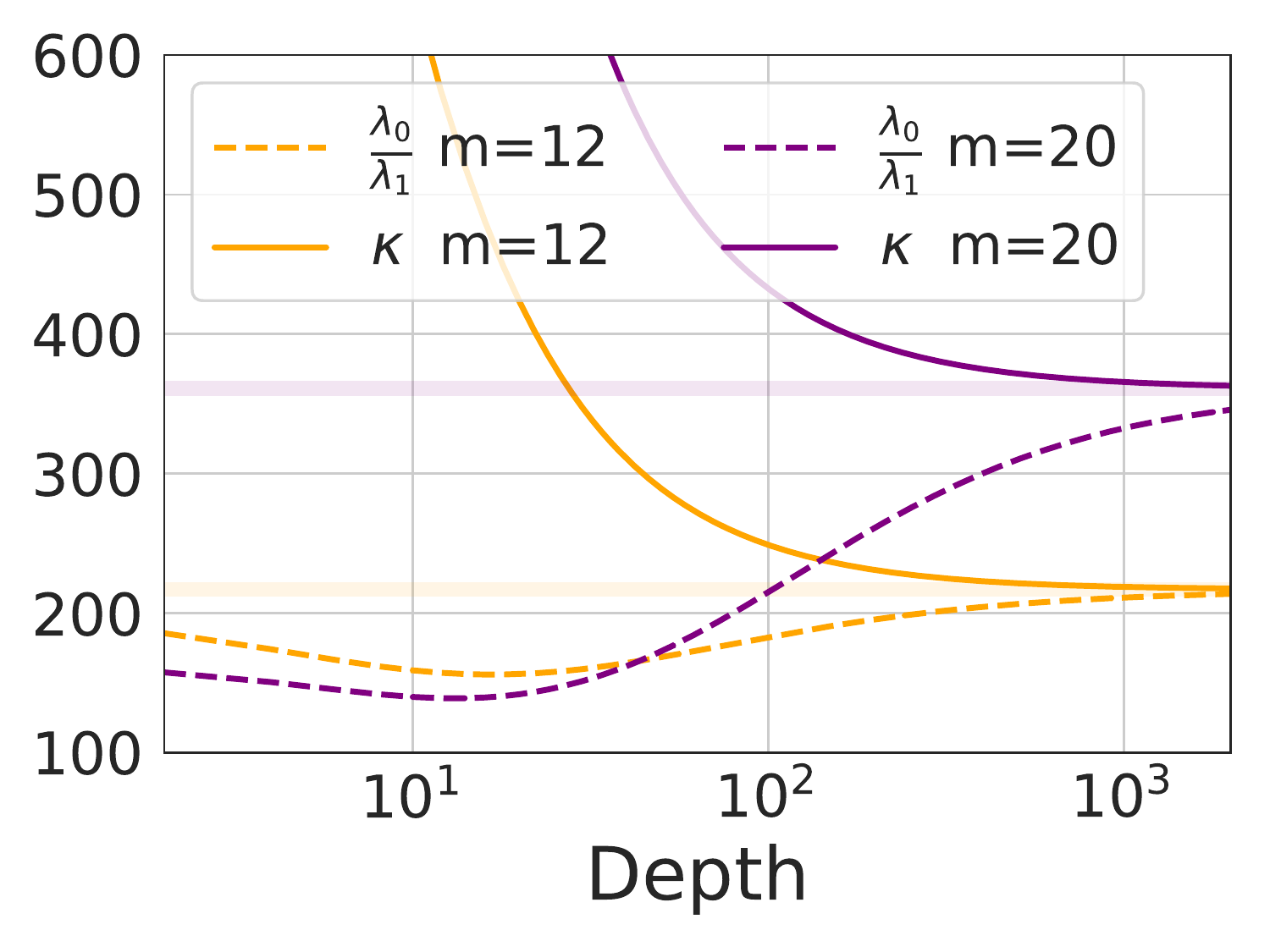}
                \caption{Critical: CNN-P}
                \label{fig:kappa-cnn-p}
        \end{subfigure}%
        \begin{subfigure}[b]{0.33\textwidth}
                \centering
              \includegraphics[width=1.\textwidth]{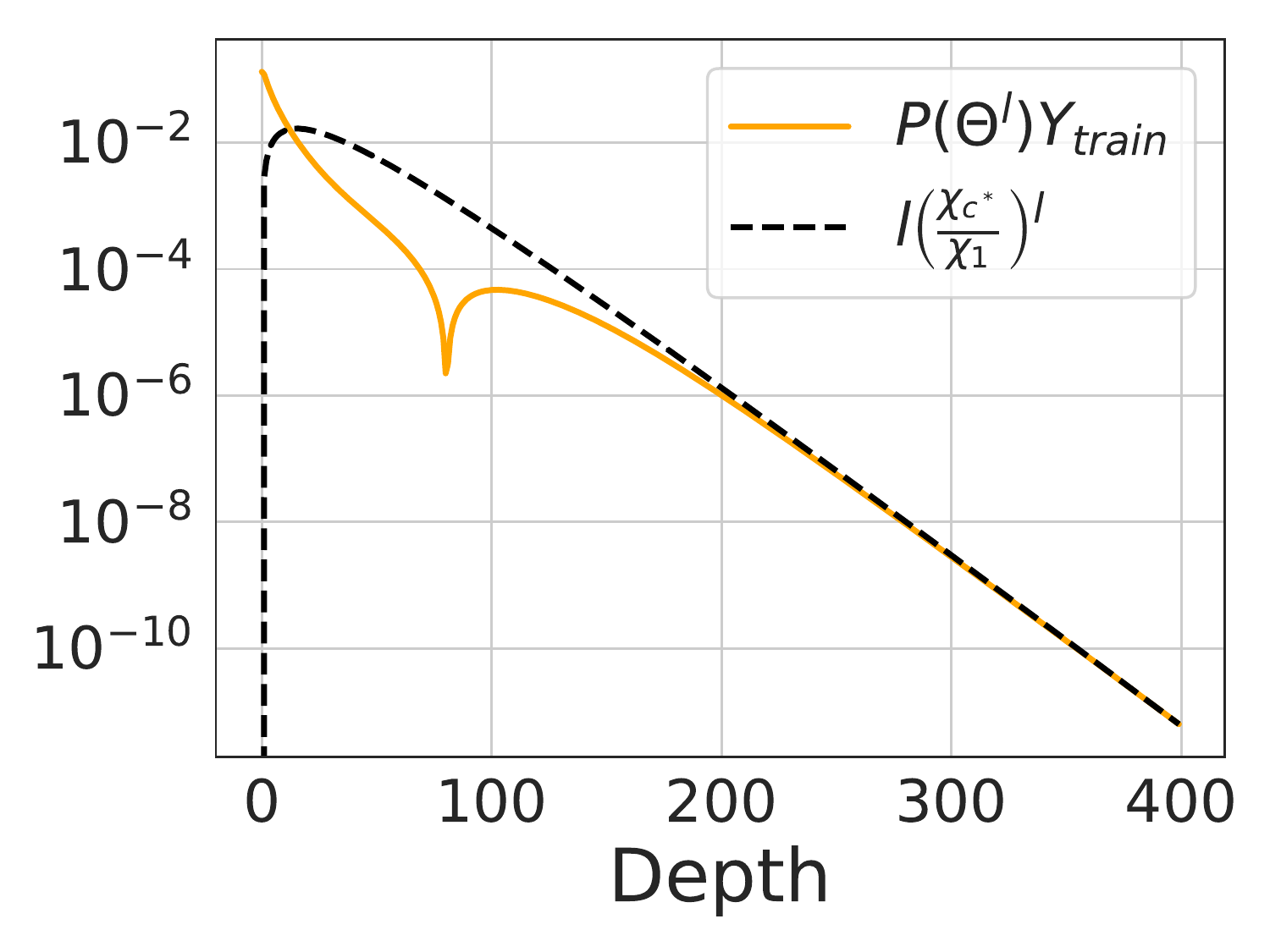}
                \caption{Chaotic: $P(\Thetal)Y_{\text{train}}$}
                \label{fig:mean-pred-chaotic}
        \end{subfigure}%
 \begin{subfigure}[b]{0.33\textwidth}
                \centering
          \includegraphics[width=1.\textwidth]{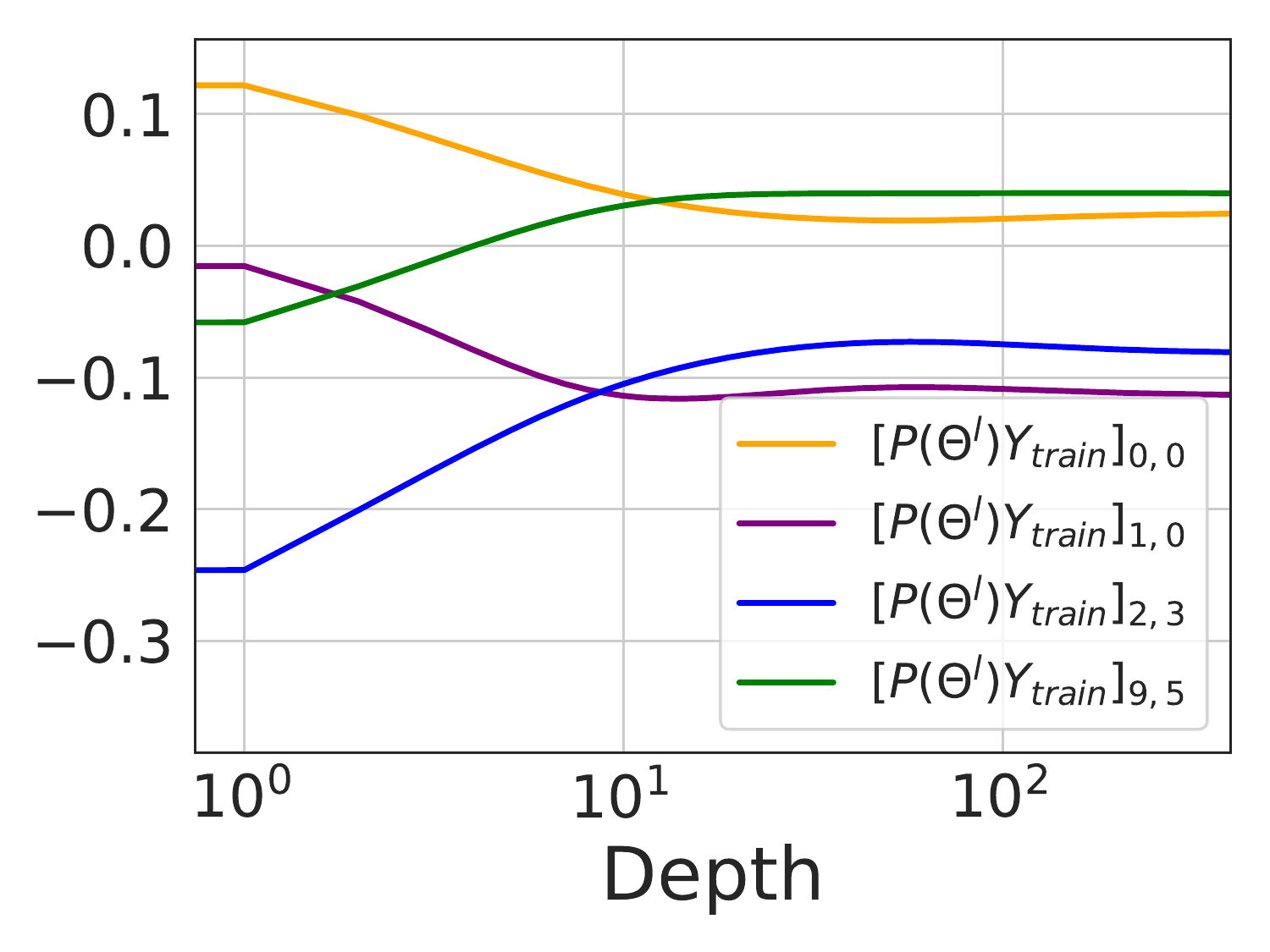}
                \caption{Ordered: $P(\Thetal)Y_{\text{train}}$}
                \label{fig:mean-pred-ordered}
        \end{subfigure}%
\end{center}
\caption{{\bf Condition number and mean predictor of NTKs and their rate of convergence for FCN, CNN-F and CNN-P.}
(a) In the chaotic phase, $\kappal$ converges to 1 for all architectures. 
(b) We plot $\chi_1^l\kappal$, confirming that
 $\kappa$ explodes with rate $1/l\chi_1^l$ in the ordered phase. 
In (c) and (d), the solid lines are $\kappal$ and dashed lines are the ratio between first and second eigenvalues. We see that, on the order-to-chaos transition,  these two numbers converge to $\frac {m+2}{2}$ and $\frac{dm+2}{2}$ (horizontal lines) for FC/CNN-F and CNN-P respectively, where $m=12$ or $20$ is the batch size and $d=36$ is the spatial dimension. 
(e) In the chaotic phase, the mean predictor decays to zero exponentially fast. (f) In the ordered phase the mean predictor converges to a data dependent value. 
}
\label{fig: conditioning-complete}
\end{figure*}
\section{A Toy Example: RBF Kernel}
To provide more intuition about our analysis, we present a toy example using RBF kernels which already shares some core observations for deep neural networks. Consider a Gaussian process along with the RBF kernel given by,
\begin{align}
    K_h(x, x') = \exp\left(-\frac{\|x - x'\|^2_2}{h}\right) 
\end{align}
where $x, x' \in X_{\text{train}}$ along with a bandwidth $h>0$. Note that $K_h(x, x)=1$ for all $h$ and $x$. Considering the following two cases.

If the bandwidth is given by $h=2^l$ and $l\to\infty$, then $K_h(x, x') \approx 1 - 2^{-l}\|x - x'\|^2_2$ which converges to $1$ exponentially fast. Thus, the largest eigenvalue of $K_h$ is $\lambda_{\text{max}}\approx|X_{\text{train}}|$ and the bulk is of order $\lambda_{\text{bulk}}\approx 2^{-l}$. Thus the condition number $\kappa \gtrsim 2^l$ which diverges with $l$. We will see in the \textbf{Ordered Phase} $\Thetal$ behaves qualitatively similar to this setting. 
    
On the other hand, if the bandwidth is given by $h=1/l$ and $l\to \infty$ then the off-diagonals $K_h(x, x') = \exp(-l \|x-x'\|_2^2)\to 0 $. For large $l$, $K_h$ is very close to the identity matrix and the condition number of it is almost 1. In the \textbf{Chaotic Phase}, $\Thetal$ is qualitatively similar to $K_h$.

%%%%%%%%%%%%%%%%%%%%%%%%%%%%%%%%%%%%%%%%%%%%%%%%%%%%%%%%%%%%%%%%%%%%%%%%%%%%%%%%%%%%%%%%%%%%%%%%%%%%%%%%%%%%%%%%%%%%%%%%%%%%%%%%%%%%%%%%%%%%%%%%%%%%
        \section{Large-Depth Asymptotics of the NNGP and NTK}
        \label{sec:theory} 
        We now give a brief derivation of the results in Table~\ref{table evolution ntk}. Details can be found in Sec.\ref{sec:signal-propagation}, \ref{sec:deltal} in the appendix. To simplify notation we will discuss fully-connected networks and then extend the results to CNNs with pooling (CNN-P) and without pooling (CNN-F). 
        
        As in Sec.~\ref{sec:background}, we will be concerned with the fixed points of $\Theta$ as well as the linearization of ~\eqref{eq: NTK-recursive} about its fixed point.
        Recall that the fixed point structure is invariant within a phase so it suffices to consider the ordered phase, the chaotic phase, and the critical line separately. In cases where a stable fixed point exists, we will describe how $\Theta$ converges to the fixed point. We will see that in the chaotic phase and on the critical line, $\Theta$ has no stable fixed point and in that case we will describe its divergence. As above, in each case the fixed points of $\Theta$ have a simple structure with $\Theta^* = p^* ((1- \hat c^{*}) \Id  + \hat c^{*}\1 \1^T)$.  
        
        To simplify the forthcoming analysis, without a loss of generality, we assume the inputs are normalized to have variance $q^*$ \footnote{It has been observed in previous works \citep{poole2016exponential, schoenholz2016} that the diagonals converge much faster than the off-diagonals for $\tanh$- or erf- networks.}. As such, we can treat $\T$ and $\dT$, restricted on $\{\Kl\}_{l}$, as a point-wise functions. To see this note that with this normalization $\K^{(l)}(x, x)=q^*$ for all $l$ and $x$. It follows that both $\T(\Kll)(x, x')$ and $\dT(\Kll)(x,x')$ depend only on $\Kl(x,x')$.  
        
        Since all of the off-diagonal elements approach the same fixed point at the same rate, we use $\qabl \equiv\Kl(x, x')$ and $\pabl\equiv\Thetal(x,x')$ to denote any off diagonal entry of $\Kl$ and $\Thetal$ respectively. We will similarly use $\qabstar$ and $\pabstar$ to denote the limits, $\lim_{l\to\infty}\qabl = \qabstar =c^*\qstar$ and $\lim_{l\to\infty}\pabl = \pabstar = \hat c^* \pstar$. Finally, although the diagonal entries of $\Kl$ are all $q^*$, the diagonal entries of $\Thetal$ can vary and we denote them $\pl$.
        
        In what follows, we split the discussion into three sections according to the values of $\chi_1 \equiv \sws\dot\T(q^*) $ recalling that in \citet{poole2016exponential, schoenholz2016} it was shown that $\chi_1$ controls the fixed point structure. In each section, we analyze the evolution of (1) the entries of $\Thetal$, i.e., $\pl$, $\pabl$, (2) the spectrum $\lmaxl$ and $\lrestl$, (3) the trainability and generalization metrics $\kappal$ and $\Deltal$, and finally (4) discuss the impact on finite width networks. 
        %The deviation for all the results are be found in Sec.~\ref{sec:signal-propagation} and ~\ref{sec:deltal}. 
        
\subsection{The Chaotic Phase $\chi_1 > 1$:}
The chaotic phase is so-named because it has a stable fixed-point $c^* < 1$; as such similar inputs become increasingly uncorrelated as they pass through the network. Our first result is to show that (see Sec.~\ref{sec:signal-prog-chaotic}), 
\small 
\begin{align}
    \begin{cases}
    \qabl = \qabstar + \bigo(\chic^l)
    \\
    \ql = \qstar
    \end{cases}
    \begin{cases}
        \pabl = \pabstar + \bigo(l\chic^l)
        \\
        \pl =  q^* \frac {\chi_1^{l} - 1} {\chi_1 - 1 }
    \end{cases}
\end{align}
\normalsize 
where 
\begin{align}
    \pabstar = \qabstar /(1 -\chic) \,\,\,{\rm and}\,\,\, \chic = \sws \dot \T(\qabstar)
\end{align}
Note that $\chic$ controls the convergence of the $\qabl$ and is always less than 1 in the chaotic phase ~\citep{poole2016exponential, schoenholz2016, xiao18a}. Since $\chi_1 > 1$, $\pl$ diverges with rate $\chi_1^l$ while $\pabl$ remains finite. It follows that $(\pl)^{-1}\Thetal \to \Id$ as $l\to\infty$. Thus, in the chaotic phase, the spectrum of the NTK for very deep networks approaches the diverging constant multiplying the identity. This implies 
\begin{align*}
\lmaxl, \, \lrestl = \pl + O(1) \,\,  {\rm and} \,\, \kappal = 1 + \bigo\left(\frac1 {\pl}\right)
\end{align*}
Figure~\ref{fig:kappa-chaotic} plots the evolution of $\kappal$ in this phase, confirming $\kappal \to 1$ for all three different architectures (FCN, CNN-F and CNN-P).

We now describe the asymptotic behavior of the mean predictor. Since $\Theta_{\text{test, train}}^l$ has no diagonal elements, it follows that it remains finite at large depths and so $P(\Theta^*) \Ytrain = 0$. It follows that in the chaotic phase, the predictions of asymptotically deep neural networks on unseen test points will converge to zero exponentially quickly (see Sec.~\ref{sec:deltal-chaotic}),
\begin{align}\label{eq:chaotic-solution}
\Deltal Y_{\text{train}} \approx
\bigo(l(\chic/\chi_1)^l)\to \bf 0.  
\end{align}
Neglecting the relatively slowly varying polynomial term, this implies that we expect chaotic networks to fail to generalize when their depth is much larger than a scale set by $\xi_* = -1 / \log(\chic/\chi_1)$. We confirm this scaling in Fig~\ref{fig:mean-pred-chaotic}.

We confirm these predictions for finite-width neural network training using SGD as well as gradient-flow on infinite networks in the experimental results; see Fig~\ref{fig:pool vs flatten}.

\subsection{The Ordered Phase $\chi_1 = \sws  \dot \T (q^*)  < 1$:}
The ordered phase is defined by the stability of the $c^* = 1$ fixed point. Here disparate inputs will end up converging to the same output at the end of the network. We show in Sec.~\ref{sec:signal-prog-ordered} that elements of the NNGP kernel and NTK have asymptotic dynamics given by, 
\begin{align}\label{eqn:signal-prop-ordered-main}
    \begin{cases}
    \qabl = \qstar  + \bigo(\chi_1^l)
    \\ 
    \ql = \qstar 
    \end{cases}
    \begin{cases}
    \pabl = \pstar + \bigo(l\chi_1^l) 
    \\
    \pl = \pstar + \bigo(\chi_1^l) 
    \end{cases}
\end{align}
where $\pstar = \qstar / (1 - \chi_1)$. Here all of the entries of $\Thetal$ converge to the same value, $\pstar$, and the limiting kernel has the form  
    $\Theta^* = p^*  \bf{1}_n\bf{1}_m^T$ where $\bf{1_m}$ is the all-ones vector of dimension $m$ (typically $m$ will correspond to the number of datapoints in the training set). The NNGP kernel has the same structure with $p^*\leftrightarrow q^*$.
    Consequently both the NNGP kernel and the NTK are highly singular and feature a single non-zero eigenvalue, $\lambda_{\text{max}} = mp^*$, with eigenvector $\bf{1_m}$.

For large-but-finite depths, $\Thetal$ has (approximately) two eigenspaces: the first eigenspace corresponds to finite-depth corrections to $\lambda_{\text{max}}$, 
\begin{align}
\lmaxl \approx (m-1) \pabl + \pl =  m\pstar + \bigo(l\chi_1^l).
\end{align} 
The second eigenspace comes from lifting the degenerate zero-modes has dimension $(m-1)$ with eigenvalues that scale like 
$
\lrestl = \bigo(\pl -\pabl) = \bigo(l\chi_1^{l}). 
$
It follows that $\kappal \gtrsim (l\chi_1^l)^{-1}$ and so the conditioning number explodes exponentially quickly. We confirm the presence of the $1/l$ correction term in $\kappal$ by plotting $\chi_1^l\kappal$ against $l$ in Figure~\ref{fig:kappa-order}. Neglecting this correction, we expect networks in the ordered phase to become untrainable when their depth exceeds a scale given by $\xi_1 = -1 / \log\chi_1$.

We now turn our discussion to the mean predictor. \eqref{eqn:signal-prop-ordered-main} shows that we can write the finite-depth corrections to the NTK as $\Thetal = p^* {\bf 1 1}^T + \bm{A}^{(l)} l\chi_1^l$. Here ${\bm A}^{(l)}$ is the data-dependent piece that lifts the zero eigenvalues. In the appendix, $\bm {A}^{(l)}$ converges to $\bm A$ as $l\to\infty$; see Lemma \ref{lemma:ordered}. 
In Sec.~\ref{sec:ordered-phase-mean-predictor} we show that despite the singular nature of $\Theta^*$, the mean has a well-defined limit as,
% could be approximately written as 
\begin{equation}
    \lim_{l\to\infty}P(\Thetal)Y_{\text{train}} = (\bm{A}_{\text{test, train}}\bm{A}_{\text{train, train}}^{-1} +\hat {\bm A})Y_{\text{train}},
\end{equation}
where $\hat {\bm A}$ is some correction term. Thus, the mean predictor remains well-behaved and data dependent even in the infinite-depth limit.
% We demonstrate the asymptotic data dependence of $\Deltal$ in Figure~\ref{fig:kappa-order}. 
Thus, we suspect that networks in the ordered phase should be able to generalize whenever they can be trained.  We confirm the asymptotic data-dependence of the mean predictor in Fig~\ref{fig:mean-pred-ordered}.

\subsection{The Critical Line $\chi_1 = \sws  \dot \T (q^*)  = 1$}
On the critical line the $c^* = 1$ fixed point is marginally stable and dynamics become powerlaw. Here, both the diagonal and the off-diagonal elements of $\Thetal$ diverge linearly in the depth with $\frac 1 l\Thetal\to \frac {\qstar} 3 (\bm 1\bm 1^T + 2 \Id)$. The condition number $\kappal$ converges to a finite value and the network is always trainable. However, the mean predictor decreases linearly with depth. In particular we show in Sec.~\ref{sec:signal-prog-critical}, 
\begin{align}
    \begin{cases}
    \qabl = \qstar  + \bigo(\frac 1 l)
    \\ 
    \ql = \qstar 
    \end{cases}
    \begin{cases}
    \pabl = \frac 1 3l \pstar + \bigo(1) 
    \\
    \pl = l \pstar 
    \end{cases}
\end{align}
For large $l$ it follows that $\Thetal$ essentially has two eigenspaces: one has dimension one and the other has dimension $(m-1)$ with 
\small 
\begin{align}
    \lmaxl = \frac {(m+2)q^*l} 3 + \bigo(1),  \,\,
  \lrestl = \frac {2q^* l} 3  +  \bigo(1).
\end{align}
\normalsize
It follows that the condition number $\kappal = \frac {m+2} 2 + m\bigo(l^{-1}) \to \frac {m+2} 2$ as $l\to\infty$. Unlike in the chaotic and ordered phases, here $\kappal$ converges with rate $\bigo(l^{-1})$. Figure~\ref{fig:kappa-critical} confirms the $\kappal\to  \frac {m+2} 2$ for both FCN and CNN-F (the global average pooling in CNN introduces a correction term that we will discuss below). A similar calculation gives $\Deltal = \bigo(l^{-1})$ on the critical line.

In summary, $\kappal$ converges to a finite number and the network ought to be trainable for arbitrary depth but the mean predictor $\Deltal$ decays as a powerlaw. Decay as $l^{-1}$ is much slower than exponential and is slow on the scale of neural networks. This explains why critically initialized networks with thousands of layers could still generalize~\citep{xiao18a}. 

\subsection{The Effect of Convolutions}
The above theory can be extended to CNNs. We will provide an informal description here, with details in Sec.~\ref{sec:conv-ntk}. For an input-images of size $(m, k, k, 3)$ the NTK and NNGP kernels will have shape $(m, k, k, m, k, k)$ and will contain information about the covariance between each pair of pixels in each image. For convenience we will let $d = k^2$. In the large depth setting deviations of both kernels from their fixed point decomposes via Fourier transform in the spatial dimensions as,
%the NTK of CNNs without pooling is essentially the same as the NTK of FCNs.
%Indeed, naively 
\begin{align}
    \delta \Thetal_{\rm CNN} \approx \sum_{q} \rho_q^l \delta \Thetal(q)
\end{align}
where $q$ denotes the Fourier mode with $q = 0$ being the zero-frequency (uniform) mode and $\rho_q$ are eigenvalues of certain convolution operator. Here $\delta\Thetal(q)$ are deviations from the fixed-point for the $q$\textsuperscript{th} mode with $\delta\Thetal(q)\propto \delta\Thetal_{\text{FCN}}$ the fully-connected deviation described above. We show that $\rho_{q=0} = 1$ and $|\rho_{q\neq 0}|<1$ which implies that asymptotically the nonuniform modes become subleading as  $\rho_{q}^l\to 0$. Thus, at large depths different pixels evolve identically as FCNs.

In Sec.~\ref{sec:cnn-pooling} we discuss the differences that arise when one combines a CNN with a flattening layer compared with an average pooling layer at the readout. In the case of flattening, the pixel-pixel correlations are discarded and $\Thetal_{\rm CNN-F} \approx \Thetal_{\rm FCN}$. The plots in the first row of Figure~\ref{fig: conditioning-complete} confirm that the $\kappal$ of $\Thetal_{\rm CNN-F}$ and of $\Thetal_{\rm FCN}$ evolve almost identically in all phases.  Note that this clarifies an empirical observation in \citet{xiao18a} (Figure 3 of \citet{xiao18a}) that test performance of critically initialized CNNs degrades towards that of FCNs as depth increases. This is because (i) in the large width limit, the prediction of neural networks is characterized by the NTK and (ii)
the NTKs of the two models are almost identical for large depth. However, when CNNs are combined with global average pooling a correction to the spectrum of the NTK (NNGP) emerges oweing to pixel-pixel correlations; this alters the dynamics of $\kappal$ and $\Deltal$. In particular, we find that global average pooling increases $\kappal$ by a factor of $d$ in the ordered phase and on the critical line; see Table~\ref{table evolution ntk} for the exact correction as well as Figures~\ref{fig:kappa-cnn-p} for experimental evidence of this correction.   

\subsection{Dropout, Relu and Skip-connection} 
% copy version; add pooling experiments. 
% move to section 6. 

Adding a dropout to the penultimate layer has a similar effect to adding a diagonal regularization term to the NTK, which significantly improves the conditioning of the NTK in the ordered phase. In particular, adding a single dropout layer can cause $\kappal$ to converge to a finite $\kappa^*$ rather than diverges exponentially; see Figure~\ref{fig:kappa-dropout} and Sec.~\ref{sec:dropout}.  
 
For critically initialized Relu networks (aka, He's initialization \cite{HeZR015}), the entries of the NTK also diverges linearly and $\kappal\to \frac{m+3}{3}$ and $\Deltal=\bigo(1/l)$; see Table~\ref{table evolution NTK-NNGP} and Figure~\ref{fig: conditioning-complete-NNGP}. In addition, adding skip-connections makes all entries of the NTK to diverge exponentially, resulting exploding of gradients. However, we find that skip connections do not alter the dynamics of $\kappal$. Finally, layer normalization could help address the issue of exploding of gradients; see Sec. \ref{sec:RElu-networks}.

\section{Experiments}\label{sec:experiments}

{\bf Evolution of $\kappal$ (Figure~\ref{fig: conditioning-complete}).}
We randomly sample inputs with shape $(m, k, k, 3)$ where $m\in \{12, 20\}$ and $k=6$. We compute the exact NTK with activation function {\it Erf} using the {\it Neural Tangents} library~\citep{neuraltangents2019}. We see excellent agreement between the theoretical calculation of $\kappal$ in Sec.~\ref{sec:theory} (summarized in Table~\ref{table evolution ntk}) and the experimental results Figure~\ref{fig: conditioning-complete}.

\begin{figure*}[h]
\begin{center} 
\includegraphics[width=.93\textwidth]{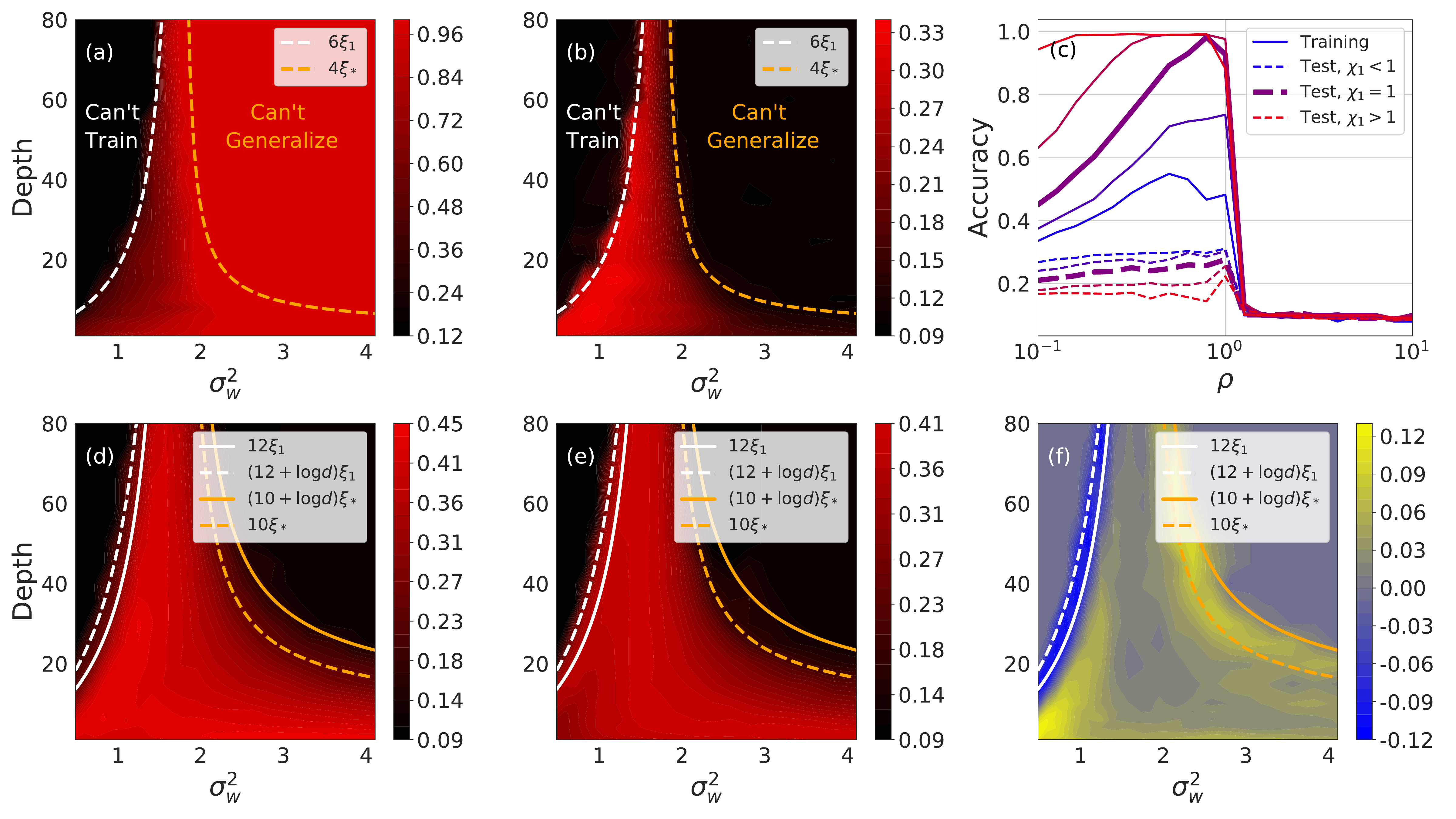}
\end{center}
\caption{{\bf Trainability and generalization are captured by $\kappal$ and $\Deltal$.}
(a,b) The training and test accuracy of CNN-F trained with SGD. The network is untrainable above the green line because $\kappal$ is too large and is ungeneralizable above the orange line because $\Deltal$ is too small. (c) The accuracy vs learning rate for FCNs trained with SGD sweeping over the weight variance. (d,e) The test accuracy of CNN-P and CNN-F using kernel regression. (f) The difference in accuracy between CNN-P and CNN-F networks.}\label{fig:pool vs flatten}
\end{figure*}

{\bf Maximum Learning Rates (Figure~\ref{fig:pool vs flatten} (c)).} 
In practice, given a set of hyper-parameters of a network, knowing the range of feasible learning rates is extremely valuable. As discussed above, in the infinite width setting, \eqref{eq:fc_ntk_recap_dynamics} implies the maximal convergent learning rate is given by $\eta_{\rm theory} \equiv 2/ {\lambda^{(l)}_{\rm max}}$. From our theoretical results above, varying the hyperparameters of our network allows us to vary $\lambda^{(l)}_{\rm max}$ over a wide range and test this hypothesis. This is shown for depth 10 networks varying $\sigma_w^2$ with $\eta = \rho\eta_{\rm theory}$. We see that networks become untrainable when $\rho$ exceeds 2 as predicted. 

{\bf Trainability vs Generalization (Figure~\ref{fig:pool vs flatten} (a,b)).}
We conduct an experiment training finite-width CNN-F networks with 1k training samples from CIFAR-10 with $20\times 20$ different $(\sws, l)$ configurations. We train each network using SGD with batch size $b=256$ and learning rate $\eta = 0.1 \eta_{\rm theory}$. We see in Figure~\ref{fig:pool vs flatten} (a) that deep in the chaotic phase we see that all configurations reach perfect training accuracy, but the network completely fails to generalize in the sense test accuracy is around $10\%$. 
As expected, in the ordered phase we see that although the training accuracy degrades generalization improves. As expected we see that the depth-scales $\xi_1$ and $\xi_*$ control trainability in the ordered phase and generalization in the chaotic phase respectively. We also conduct extra experiments for FCN with more training points (16k); see Figure~\ref{fig:train-test-acc-fcn}. 

{\bf CNN-P v.s. CNN-F: spatial correction (Figure~\ref{fig:pool vs flatten} (d-f)).}
We compute the test accuracy using the analytic equations for gradient flow, \eqref{eq:ntk-predictor}, which corresponds to the test accuracy of ensemble of gradient descent trained neural networks taking the width to infinity.
As above, we use $1k$ training points and consider a $20\times 20$ grid of configurations for $(\sws, l)$. We plot the test performance of CNN-P and CNN-F and the performance difference in Fig \ref{fig:pool vs flatten} (d-f). As expected, we see that the performance of both CNN-P and CNN-F are captured by $\xi_1 = - 1/\log(\chi_1)$ in the ordered phase and by $\xi_* = -1/ (\log \xi_c - \log\xi_1)$ in the chaotic phase. We see that the test performance difference between CNN-P and CNN-F exhibits a region in the ordered phase (a blue strip) where CNN-F outperforms CNN-P by a large margin. This performance difference is due to the correction term $d$ as predicted by the $\Deltal$-row of Table~\ref{table evolution ntk}. We also conduct extra experiments densely varying $\sbs$; see Sec.~\ref{sec:varying-sb}. Together these results provide an extremely stringent test of 
our theory.

%% file: tex_zoo/appendix.tex
\newpage 
\appendix
\section{Related Work}\label{sec:related work}
Recent work \citet{Jacot2018ntk, du2018gradient1, allen2018convergence-fc, du2018gradient, zou2018stochastic} proved global convergence of over-parameterized deep networks by showing that the NTK essentailly remains a constant over the course of training. However, in a different scaling limit the NTK changes over the course of training and global convergence is much more difficult to obtain and is known for neural networks with one hidden layer \citet{mei2018mean,chizat2018global,sirignano2018mean,rotskoff2018neural}. Therefore, understanding the training and generalization properties in this scaling limit remains a very challenging open question.

Two excellent concurrent works ~\citep{hayou2019meanfield, jacot2019freeze} also study the dynamics of $\Thetal(x, x')$ for FCNs (and deconvolutions in \cite{jacot2019freeze}) as a function of depth and variances of the weights and biases. \cite{hayou2019meanfield} investigates role of activation functions (smooth v.s. non-smooth) and skip-connection. \cite{jacot2019freeze} demonstrate that batch normalization helps remove the ``ordered phase'' (as in~\cite{yang2019mean}) and a layer-dependent learning rate allows every layer in a network to contribute to learning. As opposed to these contributions, here we focus our effort on understanding trainability and generalization in this context. We also provide a theory for a wider range of architectures than these other efforts.

\section{Signal propagation of NNGP and NTK} \label{sec:signal-propagation}
In this section, we assume that the activation function $\phi$ has a continuous third derivative. Recall that the recursive formulas for NNGP $\Kl$ and the NTK $\Thetal$ are given by  
\begin{align}\label{eq:fc_mft_recap_recursion}
    \Kll(x, x') &= \sigma_w^2 \mathcal T(\Kl)(x, x') + \sigma_b^2 \,. 
    \\ 
    \Thetall(x, x') &= \Kll(x, x') + \sigma_w^2\dot{\T}(\Kl)(x, x')\Thetal(x, x')\,
\end{align}
where 
\begin{align}
            \begin{cases}
                \T(\K)(x, x') = \mathbb E \phi(u)\phi(v),
                \\
                \dot\T(\K)(x, x') = \mathbb E \dot \phi(u)\dot\phi(v),
            \end{cases} 
            \quad (u, v)^T \sim 
            \N\left(0, \,\,\begin{bmatrix} q^*&\K(x, x') \\\K(x, x') &q^* 
            \end{bmatrix}
            \right) \label{eq: t_reduce}
% \label{sec:signal prop}
\end{align}
Note that we have normalized each input to have variance $\qstar$ and the diagonals of $\Kl$ are equal to $\qstar$ for all $\l$.     
The off-diagonal terms of $\Kl$ and $\Thetal$ are denoted by $\qabl$ and $\pabl$, resp. and the diagonal terms are $\ql$ and $\pl$, resp. The above equations can be simplified to  
\begin{align}
        &\qabll  = \sigma_w^2 \T(\qabl) + \sigma_b^2\,\quad  &&\pabll  = \qabll + \sigma_w^2 \dot \T(\qabl)\,\pabl
        \label{eq:recursive-off-new}
        \\
        &\qll = \qstar\,\quad &&\pll  = \qstar + \sigma_w^2 \dot \T(\qstar)\,\pl
                \label{eq:recursive-diag-new}
        \end{align}
In what follows, we compute the evolution of $\qabl$, $\pabl$, $\pl$ and the spectrum and condition numbers of $\Kl$ and $\Thetal$. We will use $\lmax(\Thetal)/\lmax(\Kl)$, $\lrest(\Thetal)/\lrest(\Kl)$ and $\kappa(\Thetal)/\kappa(\Kl)$ to denote the maximum eigenvalues, the bulk eigenvalues and the condition number of $\Thetal/\Kl$, resp.

\begin{table}[t]
\begin{center}
\resizebox{0.9\columnwidth}{!}{%
\begin{tabular}{llll}
        \toprule
                  & \multicolumn{3}{c}{{\color[HTML]{000000} {\bf NTK} $\Thetal$ of  FC/CNN-F, \quad  \color{blue}{CNN-P}}}\\[0.1cm]  
        \midrule
\multirow{-2}{*}{}  & Ordered     $\chi_1 <1$       & Critical $\chi_1=1$            & Chaotic $\chi_1 > 1$            \\ [0.4cm]
                 $ \lmaxl$                 &   $m\pstar +m \bigo (l\chi_1^l)$        &       $\frac {m{\colord} +2} {3\colord} l\qstar   + m \bigo(1)$         &      $\Theta(\chi_1^l)\color{blue}{/}\colord$  \\ [0.4cm]
                  $\lrestl$                 &  $ \bigo(l\chi_1^l)\color{blue}{/}\colord$          &     $\frac 2 {3 \colord } l\qstar  + \frac{1}{\colord}\bigo(1)$          &            $\Theta(\chi_1^l)\color{blue}{/}\colord$                  \\ [0.4cm]
                  $\kappal$                 &   $\colord m\pstar \Omega(\chi_1^{-l} / l)$         
                  &    $\frac {m{\colord} +2} 2 + \colord  m\bigo(l^{-1})$            & $1 + \bigo({\colord}\chi_1^{-l})$                
                  \\ [0.4cm]
                  $P(\Thetal)Y_{\text{Train}}$                 &   $ \bigo(1)$         
                  &    $\colord  \bigo(l^{-1})$            
                  & $\colord \bigo(l (\chi_{c^*}/\chi_1)^l  )$   
                  \\
        \bottomrule
% .
        \toprule
                  & \multicolumn{3}{c}{{\color[HTML]{000000} {\bf NNGP} $\Kl$ of  FC/CNN-F, \quad  \color{blue}{CNN-P}}}\\[0.1cm]  
        \midrule
\multirow{-2}{*}{}  & Ordered     $\chi_1 <1$       & Critical $\chi_1=1$            & Chaotic $\chi_1 > 1$            \\ [0.4cm]
                 $ \lmaxl$                 &   $m\qstar +m \bigo (\chi_1^l)$        &       $m\qstar + \bigo(l^{-1})$         &      $ ((1-c^{*}){\color{blue}{/}}\colord + m c^{*})\qstar+\bigo(\chic^l)$  \\ [0.4cm]
                  $\lrestl$                 &  $ \bigo(\chi_1^l)\color{blue}{/}\colord$          &     $\bigo(l^{-1})\color{blue}{/}\colord$     &            $(1 - c^*)\qstar{\color{blue}{/}}\colord +\bigo(\chic^l)$                  \\ [0.4cm]
                  $\kappal$                 &   $\colord m\qstar \Omega(\chi_1^{-l})$         
                  &    $\colord m\Omega(l)$            & $1 + \colord m \frac{\cstar}{1-\cstar} + d\bigo(\chic^l)$                
                  \\ [0.4cm]
                  $P(\Kl)Y_{\text{Train}}$                 &   $ \bigo(1)$         
                  &    $\bigo(l^{-1})$            
                  & $\colord \bigo(\chic^l  )$            
                  \\
        \bottomrule
\end{tabular}% 
} 
\end{center}
\captionsetup{singlelinecheck=off}
    \caption{\textbf{Evolution of the NTK/NNGP spectrum and $P(\Thetal)Y_{\text{train}}/P(\Kl)Y_{\text{train}}$ as a function of depth $l$.} The NTKs of FCN and CNN without pooling (CNN-F) are essentially the same and the scaling of $\lmaxl$, $\lrestl$, $\kappal$, and $\Delta^{(l)}$ for these networks is written in black. Corrections to these quantities due to the addition of an average pooling layer ({\color{blue} CNN-P}) with window size $\colord$ is written in blue.}
    \label{table evolution NTK-NNGP}
\end{table}
\subsection{Chaotic Phase}\label{sec:signal-prog-chaotic}
\subsubsection{Correction of the off-diagonal/diagonal}
The diagonal terms are relatively simple to compute. \eqref{eq:recursive-diag-new} gives 
\begin{align}
    \pll = \qstar + \chi_1 \pl
\end{align}
i.e. 
\begin{align}
    \pl  = \frac {1-\chi_1^{(l)}}{1-\chi_1} \qstar \,  
\end{align}
In the chaotic phase, $\chi_1>1$ and $\pl\approx \chi_1^{l-1}\qstar$, i.e. diverges exponentially quickly.    

Now we compute the off-diagonal terms. Since $\chic=\sws \dot \T(\qabstar) <1$ in the chaotic, $\pabstar$ exists and is finite. Indeed,  
letting $l\to\infty$ in equation \ref{eq:recursive-off-new}, we have 
\begin{align}
     &\qabstar = \sigma_w^2 \T(\qabstar) + \sigma_b^2\,
     \\&\pabstar  = \qabstar + \sigma_w^2 \dot \T(\qabstar)\,\pabstar 
\end{align}
which gives 
\begin{align}
    \pabstar = \frac {\qabstar} {1 - \chic} 
\end{align}
To compute the finite depth correction, let  
\begin{align}
    \elab &= \qabl - \qabstar
    \\
    \dlab &= \pabl  - \pabstar
 \end{align}
Applying Taylor's expansion to the first equation of \ref{eq:recursive-off-new} gives 
\begin{align}
    \qabstar  + \ellab &= \sws \T(\qabstar + \elab) + \sigma_b^2
    \\
    &= \sws \T(\qabstar ) + \sigma_b^2 +\sws \dot \T(\qabstar)\elab +  \bigo((\elab)^2)
    \\
    &= \qabstar +\sws \dot \T(\qabstar)\elab +  \bigo((\elab)^2) 
\end{align}
That is 
\begin{align}
    \ellab = \chic \elab + \bigo((\elab)^2) 
\end{align}
Thus $\qabl$ converges to $\qabstar$ exponentially quickly with  
\begin{align}
    \ellab &\approx\chic\elab \approx \chic^{l+1}\epsilon_{ab}^{(0)} 
\end{align}
Similarly, applying Taylor's expansion to the second equation of \ref{eq:recursive-off-new} gives    
\begin{align}\label{eq:delta-ab-epsilon}
    \dllab = (1 + \frac{\chi_{c^{*}, 2}}{\chic}\pabstar) \ellab + \chic\dlab 
    + \bigo((\elab)^2)
\end{align}
where $\chi_{c^*, 2} = \sws \ddot\T(\qabstar)$.
This implies  
\begin{align}\label{eq:off-diagonal-correction}
\elab &\approx\chic^l\, \epsilon_{ab}^{(0)}\\ 
\delta^{(l)}_{ab} &\approx  \chic^l\left[\delta^{(0)}_{ab} + l\left(1 + \frac{\chi_{c^{*}, 2}}{\chic}\pab\right)\epsilon^{(0)}_{ab}\right].   
\end{align}
Note that $\dlab$ contains a polynomial correction term and decays like $l\chic^l$.
\begin{lemma}\label{lamma:chaotic-convergence}
There exist a finite number $\zeta_{ab}$ such that 
\begin{align}\label{eq:convergence-2-finite}
    | \chic^{-l} \elab -\zeta_{ab}| \lesssim \chic^{l}\quad {\text and } \quad 
    |\chic^{-l}\dllab - l(1+ \frac{\chi_{c^{*}, 2}}{\chic}\pabstar)\zeta_{ab}|\lesssim \chic^l. 
\end{align}
\end{lemma}
We want to emphasize that the limits are data-dependent, which was verified in Fig. \ref{fig:mean-pred-chaotic} and \ref{fig:mean-pred-ordered} empirically. 
\begin{proof}
Let $\zeta^{(l)}_{ab} = \chic^{-l} \elab $.  
We will show $\zeta^{(l)}_{ab}$ is a Cauchy sequence. 
For any $k>l$ 
\begin{align} \label{eq:epsilon-ab-exponential}
    |\chic^{-l} \elab - \chic^{-k} \epsilon_{ab}^{(k)}| 
    \leq \sum_{j = l}^\infty  |\chic^{-j} \epsilon^{(j)} - \chic^{-j-1} \epsilon_{ab}^{(j+1)}| =\bigo(\sum_{j=l}^\infty \chic^{-(j+1)}(\epsilon_{ab}^{(j)})^2)
    \lesssim \epsilon_{ab}^{(0)} \sum_{j = l}^\infty   \chic^{j-1}\lesssim \chic^l
\end{align}
Thus $\zeta_{ab}\equiv\lim_{l\to\infty} \chic^{-l} \elab $ exists and
\begin{align}
    |\zeta^{(l)}_{ab} - \zeta_{ab}| 
    \lesssim \chic^l. 
\end{align}
\eqref{eq:delta-ab-epsilon} gives
\begin{align}
   \chic^{-(l+1)}\dllab = (1 + \frac{\chi_{c^{*}, 2}}{\chic}\pabstar) (\chic^{-(l+1)})\ellab + \chic^{-l}\dlab 
    + \chic^{-(l+1)}\bigo((\elab)^2) 
\end{align}
Let $\eta^{(l)}_{ab} = \chic^{-l}\dlab  - l(1 + \frac{\chi_{c^{*}, 2}}{\chic}\pabstar)\zeta_{ab}$.  
Coupled the above equation with \eqref{eq:epsilon-ab-exponential}, we have 
\begin{align}
    |\eta^{(l+1)}_{ab} - \eta^{(l)}_{ab}| \lesssim \chic^l
\end{align}
Summing over all $l$ implies 
\begin{align}
    |\chic^{-l}\dlab  - l(1 + \frac{\chi_{c^{*}, 2}}{\chic}\pabstar)\zeta_{ab}| \lesssim \chic^l. 
\end{align}
\end{proof}

\subsubsection{The Spectrum of the NNGP and NTK}
We consider the spectrum of $\K$ and $\Theta$ in this phase.  
For $\Kl$, we have $\qabstar = \cstar \qstar$ (with $\cstar<1$), $\ql = \qstar$ and $\qabl = \qabstar + \bigo(\chic^l)$. Thus 
\begin{align}
\Kl = \K^{*} + \mathcal E^{(l)}
\end{align}
where 
\begin{align}
        \K^{*} &= \qstar(\cstar \bm 1 \bm 1^T + (1 - \cstar \Id))
        \\
         \mathcal E_{ij}^{(l)} &= \bigo(\chic^l) 
\end{align}
The NNGP $\K^{*}$ has two different eigenvalues: $\qstar(1 + (m-1)\cstar)$ of order 1 
and $\qstar(1 - \cstar)$ of order $(m-1)$, where $m$ is the size of the dataset. 
For large $l$, since the spectral norm of $\mathcal E^l$ is $\bigo(\chic^l)$, the spectrum and condition number of $\Kl$ are 
\begin{align}
\lmax(\Kl) &= \qstar(1 + (m-1)\cstar) + \bigo(\chic^l)
\\
\lrest(\Kl) &= \qstar(1 - \cstar) + \bigo(\chic^l)
\\
\kappa(\Kl) &= \frac {(1 + (m-1)\cstar)}{1 - \cstar} + \bigo(\chic^l).    
\end{align}

For $\Thetal$, we have $\pabl = \pabstar + \bigo(l\chic^l)\to \pabstar<\infty$ and $\pl = \frac{1-\chi_1^l}{1-\chi_1}\qstar\to\infty$, i.e. 
\begin{align}
    (\pl)^{-1}\Thetal = \Id + \bigo((\pl)^{-1}) 
\end{align}
Thus $\Thetal$ is essentially a diverging constant multiplying the identity and  
\begin{align}
    \lmax(\Thetal) &= \pl + \bigo(1)
    \\
    \lrest(\Thetal) &= \pl + \bigo(1)
    \\
    \kappa(\Thetal) &= 1  +  \bigo((\pl)^{-1})
\end{align}

\subsection{Ordered Phase}
\label{sec:signal-prog-ordered}
\subsubsection{The correction of the diagonal/off-diagonal}
In the ordered phase, $\qabl\to \qstar$, $\ql =\qstar$, $\pl \to\pstar$ and $\pabl \to \pstar$.  Indeed, letting $\l\to\infty$ in the equations \ref{eq:recursive-diag-new} and \ref{eq:recursive-off-new}, 
\begin{align}
    \qstar &= \sws  \T(\qstar)  + \sbs \\
    \pstar &=  \frac{1}{1 - \chi_1} \qstar 
\end{align}
The correction of the diagonal terms are
$\pl = \pstar - \frac{\chi_1^l - \chi_1}{1-\chi_1}\qstar$.  
Same calculation as in the chaotic phase implies 
\begin{align}\label{eq:off-diagonal-correction}
\elab &\approx\chi_1^l\, \epsilon_{ab}^{(0)}\\ 
\delta^{(l)}_{ab} &\approx  \chi_1^l\left[\delta^{(0)}_{ab} + l\left(1 + \frac{\chi_{1, 2}}{\chi_1}\pab\right)\epsilon^{(0)}_{ab}\right].   
\end{align}
where $\chi_{1, 2} = \sws \ddot\T(\qstar)$.  
Note that $\dlab$ contains also a polynomial correction term and decays like $l\chi_1^l$.

Similar, in the ordered phase we have the following.   
\begin{lemma}\label{lemma:ordered}
There exists $\zeta_{ab}$ such that 
\begin{align}
    | \chi_1^{-l}\elab -\zeta_{ab}| \lesssim \chi_1^l 
    \quad {\text and} \quad 
    |\chi_1^{-l}l^{-1}\dlab - (1+\frac{\chi_{1, 2}}{\chi_1}\pabstar) \zeta_{ab}| 
    \lesssim \chi_1^l
\end{align}
Therefore the following limits exist 
\begin{align}
    \lim_{l\to\infty} \chi_1^{-l}(\Kl - \K^*)
    \quad {\text{and}} \quad 
        \lim_{l\to\infty} \chi_1^{-l}l^{-1}(\Thetal - \Theta^*)
\end{align}
\end{lemma}
Since the proof is almost identical to Lemma \ref{lamma:chaotic-convergence}, we omit the details. 
\subsubsection{The Spectrum of the NNGP and NTK}
For $\Kl$, we have $\qabstar = \qstar$, $\qabl= \qstar  + \bigo(\chi_1^l)$ and $\ql =\qstar$. Thus
\begin{align}
    \Kl = \qstar \bm1 \bm1^T +\bigo(\chi_1^l)
\end{align}
which implies 
\begin{align}
    \lmax(\Kl) &= m\qstar + \bigo(\chi_1^l)
    \\
    \lrest(\Kl) &= \bigo(\chi_1^l)
    \\
    \kappa(\Kl) &\gtrsim \chi_1^{-l} 
\end{align}

For $\Thetal$, $\pabl = \pstar + \bigo(l\chi_1^l)$ and $\pl = \pstar - \frac{\chi_1^l - \chi_1}{1-\chi_1}\qstar = \pstar + \bigo(\chi_1^l)$. Thus 
\begin{align}
    \Thetal  = \pstar \bm1 \bm1 ^T + \bigo(l\chi_1^l)
\end{align}
which implies 
\begin{align}
    \lmax(\Thetal) &= m \pstar + \bigo(l\chi_1^l)
    \\
    \lrest(\Thetal) &= \bigo(l\chi_1^l)
    \\
    \kappa(\Thetal) &\gtrsim (l\chi_1^l)^{-1}
\end{align}

\subsection{The critical line.}
\label{sec:signal-prog-critical}

\begin{figure*}[h]
\begin{center}
         \begin{subfigure}[b]{.33\textwidth}
                \centering
\includegraphics[height=4.cm]{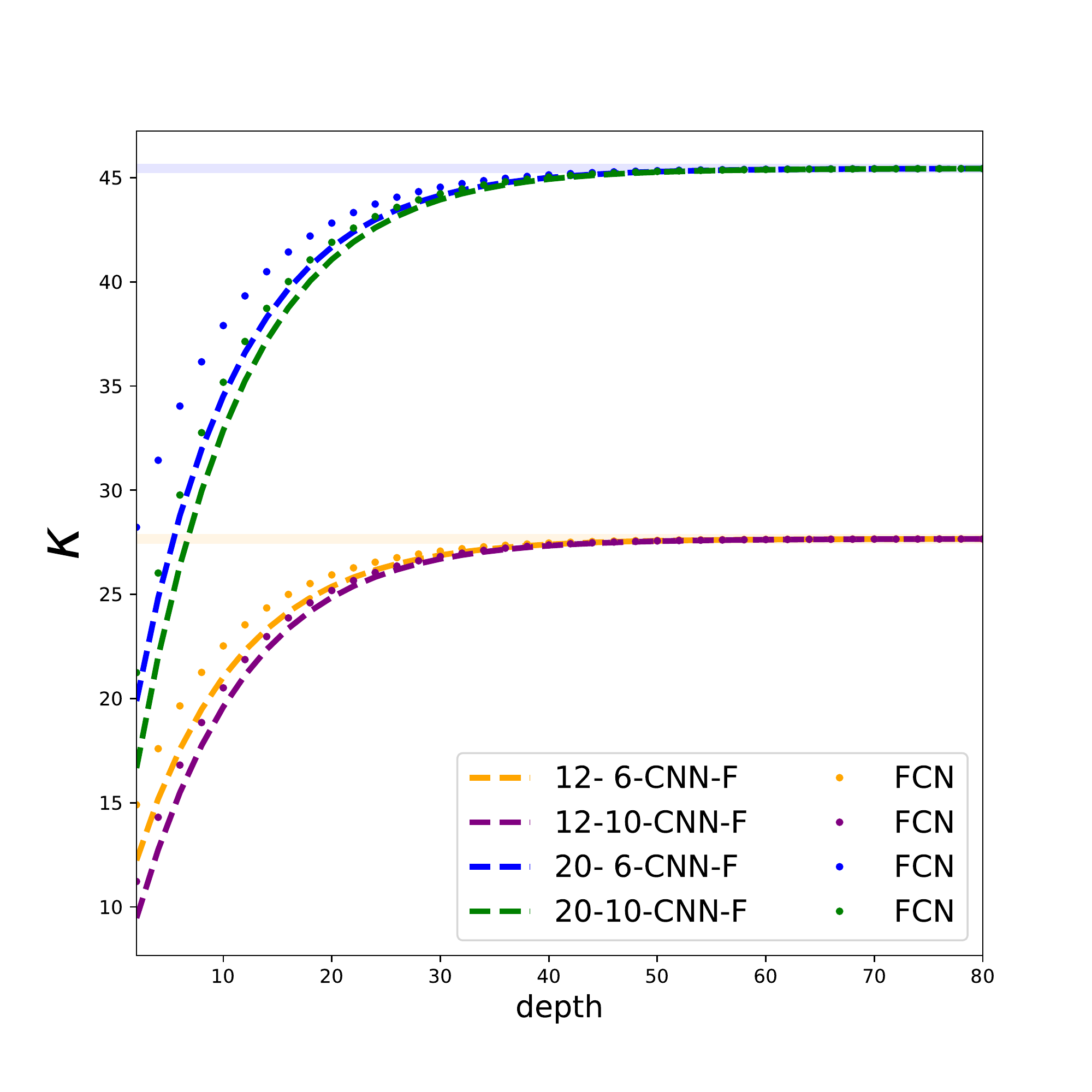}
                \caption{NNGP Chaotic}
                \label{fig:nngp-kappa-chaotic-fcn}
        \end{subfigure}
        \begin{subfigure}[b]{0.33\textwidth}
                \centering
               \includegraphics[height=4.cm]{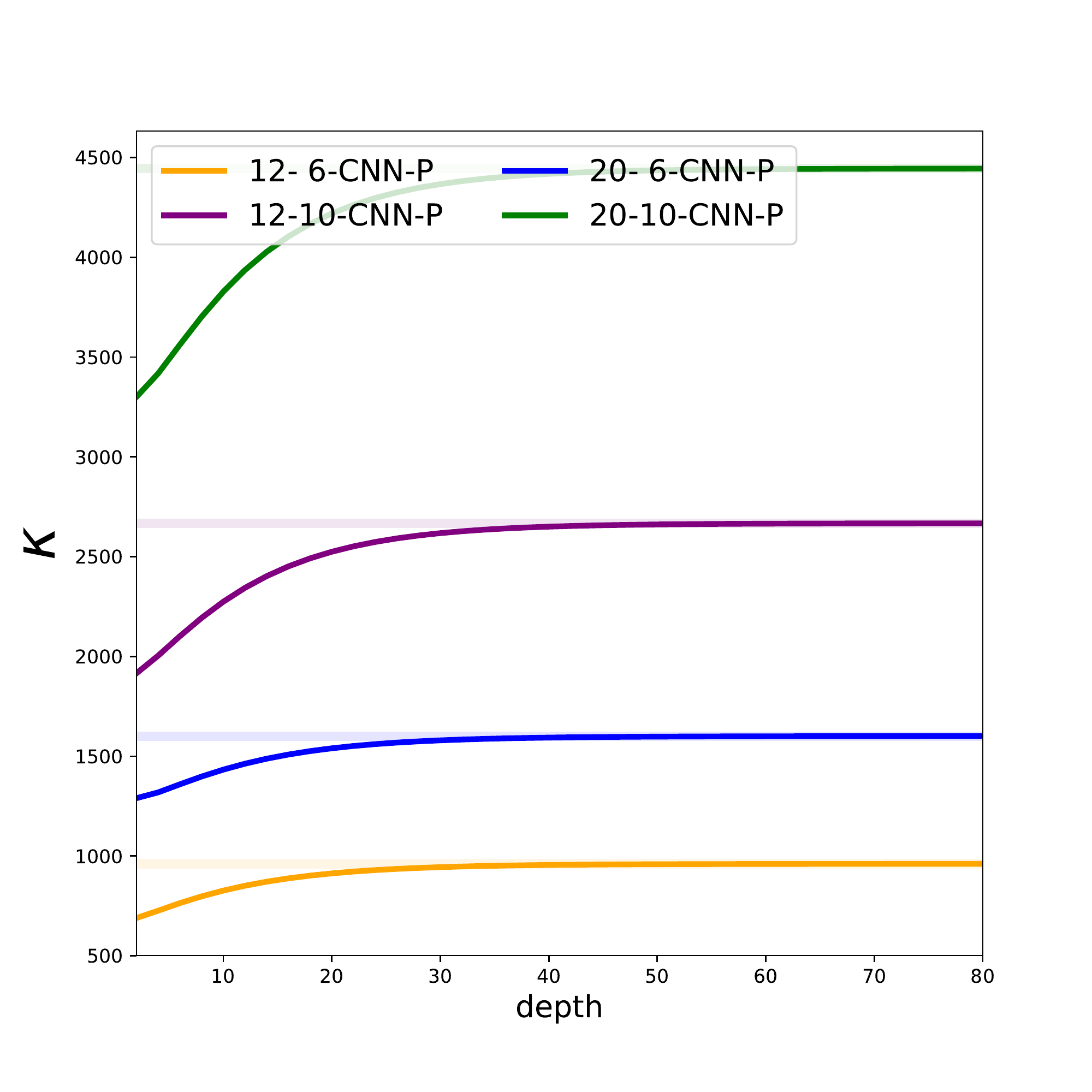}
                \caption{NNGP Chaotic}
                \label{fig:nngp-kappa-chaotic-cnn-p}
        \end{subfigure}%
        \begin{subfigure}[b]{0.33\textwidth}
                \centering
               \includegraphics[height=4.cm]{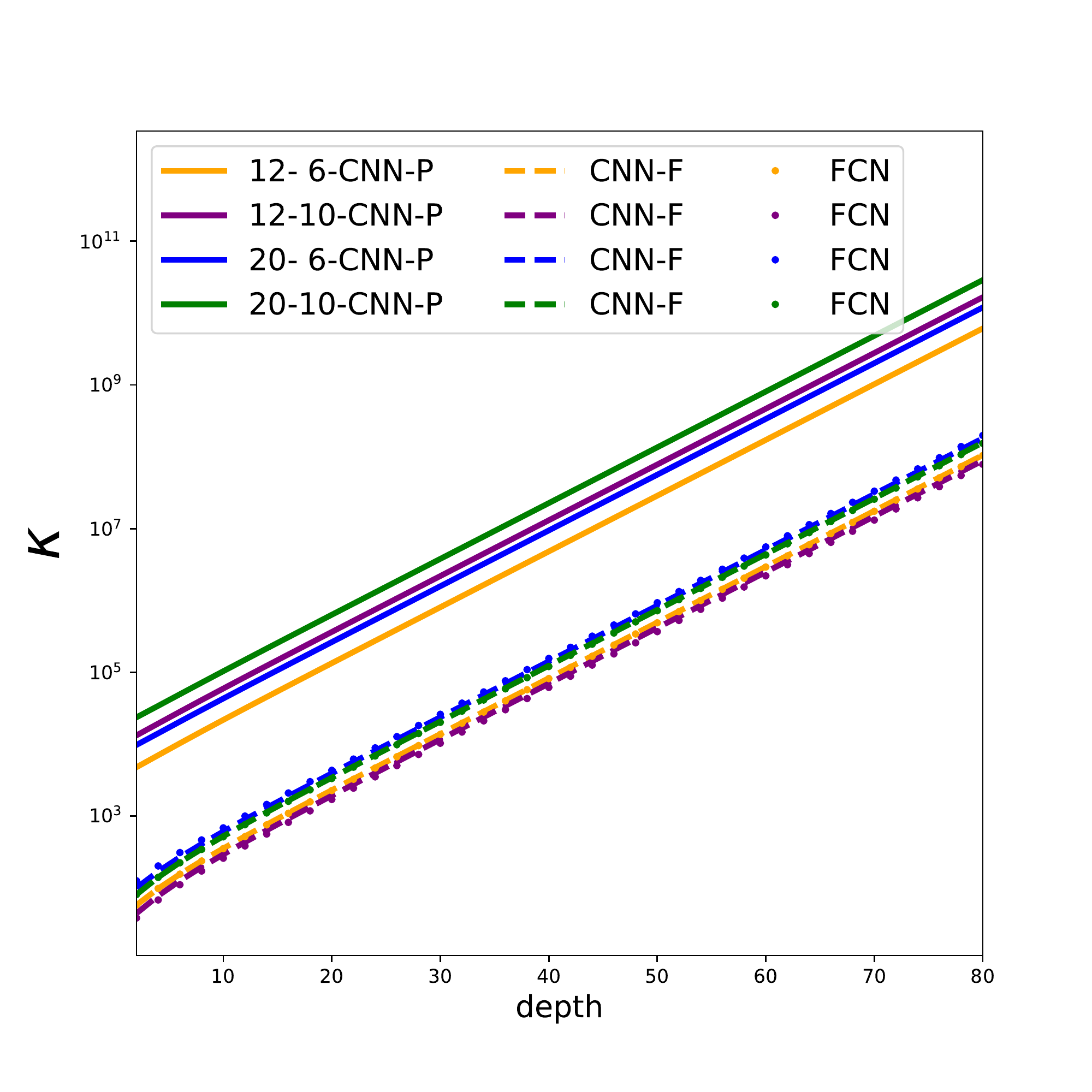}
                \caption{NNGP Ordered}
                \label{fig:nngp-kappa-ordered}
        \end{subfigure}%
        \\ 
 \begin{subfigure}[b]{0.33\textwidth}
                \centering
     \includegraphics[height=4.cm]{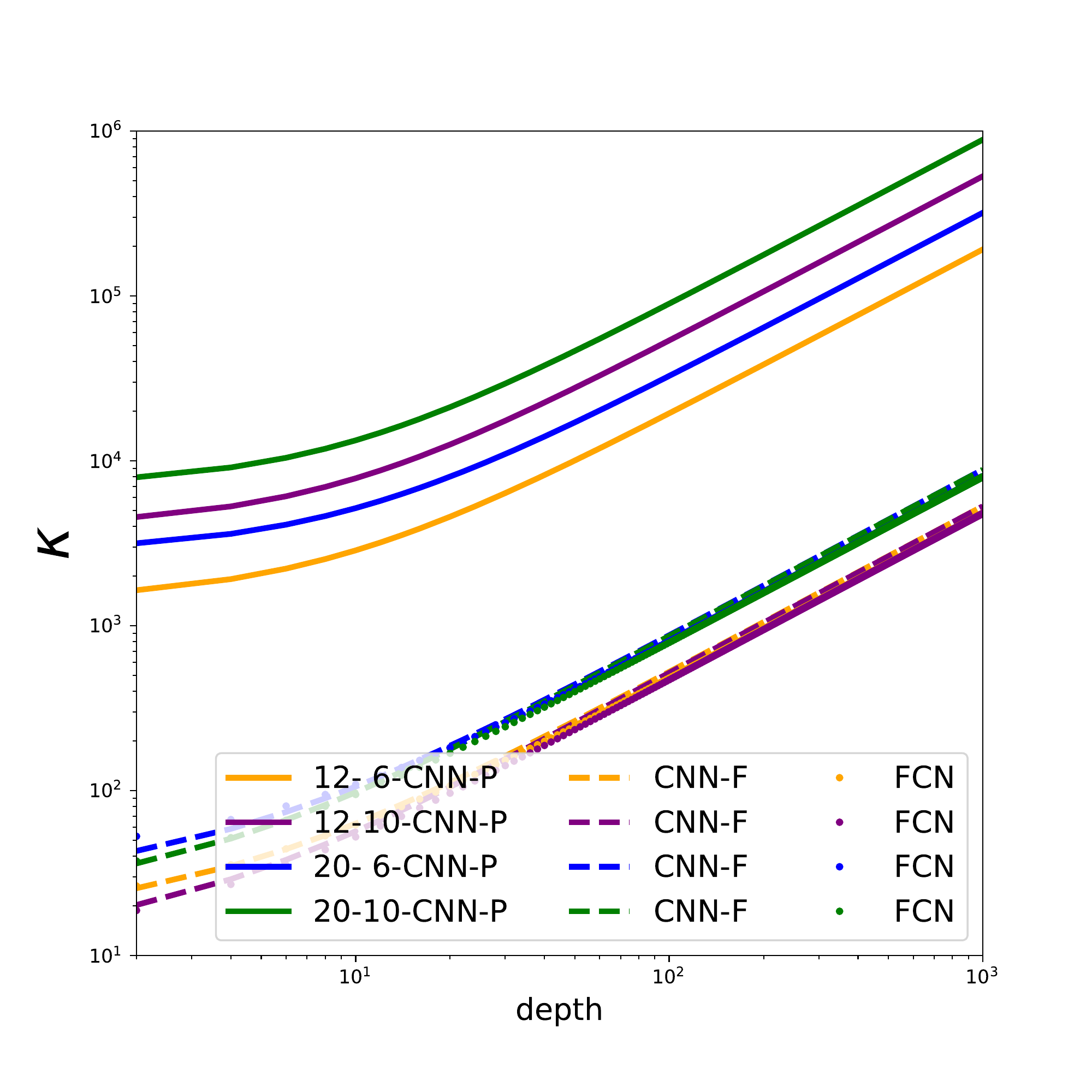}
                \caption{NNGP Critical}
                \label{fig:nngp-critical}
        \end{subfigure}%
        \begin{subfigure}[b]{0.33\textwidth}
                \centering
              \includegraphics[height=4.cm]{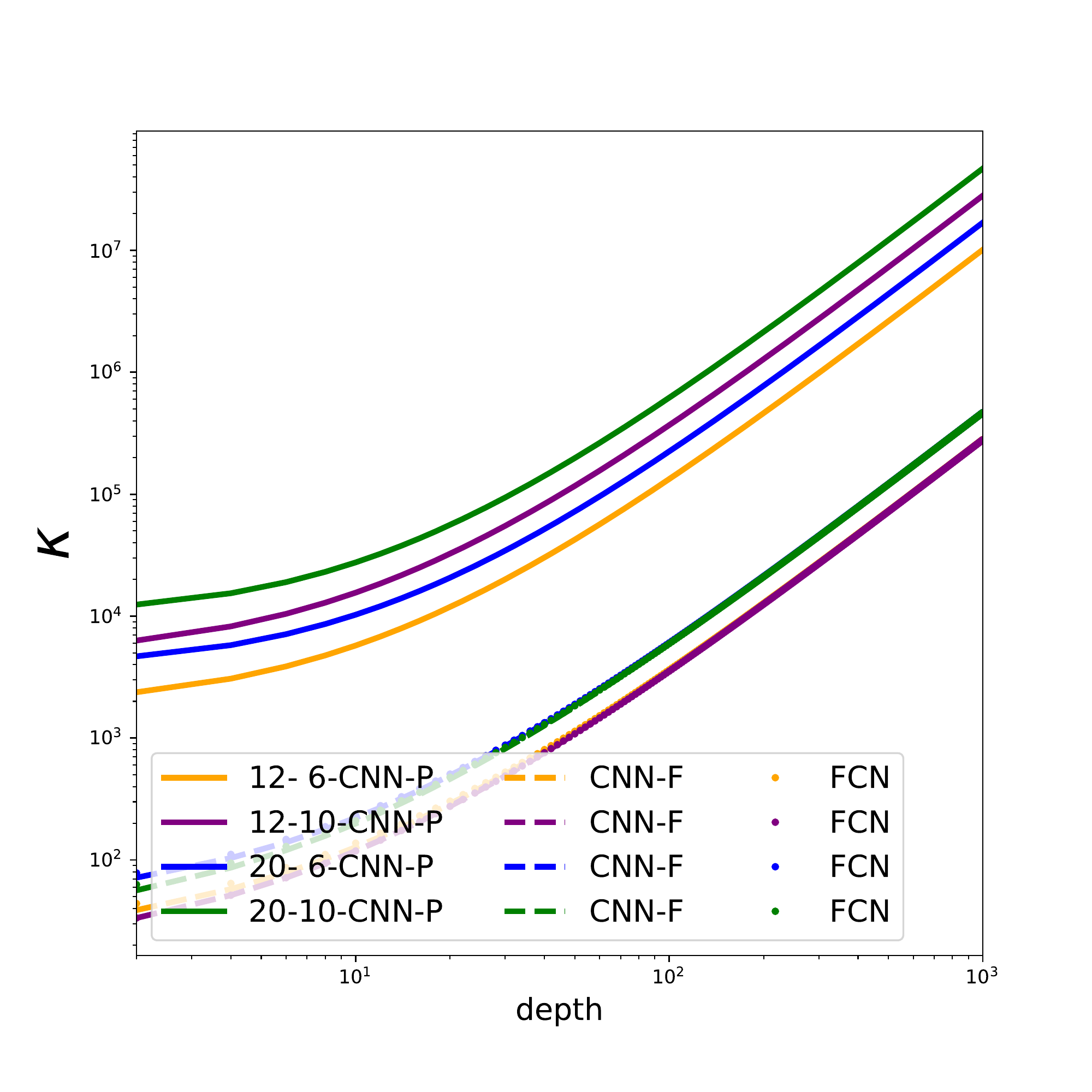}
                \caption{NNGP RELU}
                \label{fig:nngp-relu}
        \end{subfigure}%
 \begin{subfigure}[b]{0.33\textwidth}
                \centering
          \includegraphics[height=4.cm]{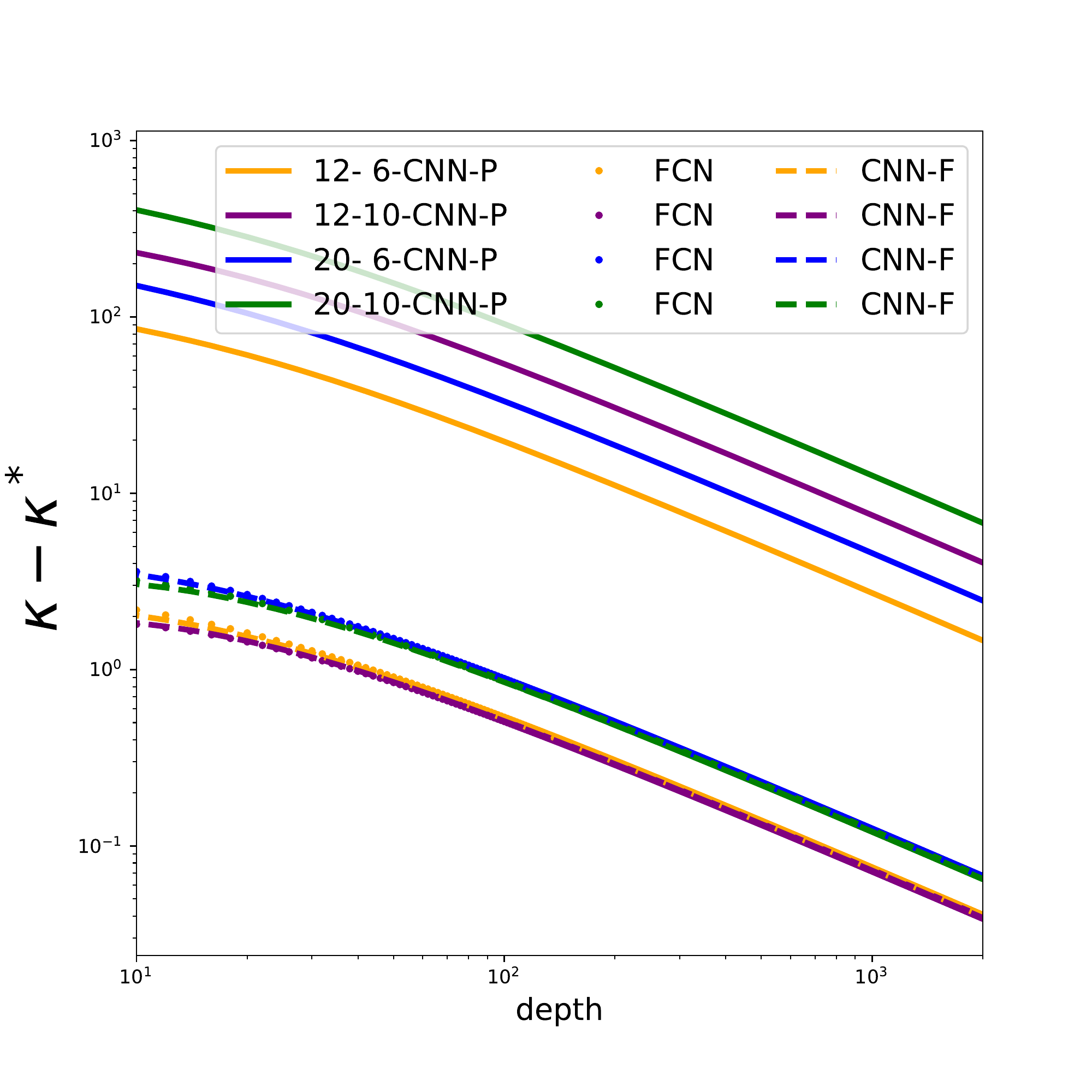}
                \caption{NTK RELU}
                \label{fig:relu-ntk}
        \end{subfigure}%
\end{center}
\caption{{\bf Condition numbers of NNGP and their rate of convergence.} 
In the chaotic phase, $\kappa(\Kl)$ converges to a constant (see Table~\ref{table evolution NTK-NNGP}) for FCN, CNN-F (a) and CNN-P (b). 
However, it diverges exponentially in the ordered phase (c) and linearly on the critical line (d). For critical RELU network, $\kappa(\Kl)$ diverges quadratically (e) while $\kappa(\Thetal)$ converges to a fixed number with rate $(l^{-1})$ (see \eqref{eq:relu-condition-number-nngp-ntk}) and we plot the value of $(\kappa(\Thetal) - \kappa(\Theta^*))$ of the NTK in (f). }
\label{fig: conditioning-complete-NNGP}
\end{figure*}

\subsubsection{Correction of the diagonals/off-diagonals.}
We have $\chi_1=1$ on the critical line. \eqref{eq:recursive-diag-new} implies $\pl = l\qstar$, i.e. the diagonal terms diverge linearly.  
To capture the linear divergence of $\pabl$, define    
\begin{align}
    \elab &= \qabl - \qabstar
    \\
    \dlab &= \pabl - l\qstar 
\end{align}
We need to expand the first equation of \ref{eq:recursive-off-new} to the second order
\begin{align}
    \label{eq:second-ordered-correctoin}
    \ellab = \elab +\frac 1  2 \chi_{1, 2} \left( \elab\right)^2 + \bigo((\elab)^3)   
\end{align}
Here we assume $\T$ has a continuous third derivative (which is sufficient to assume the activation $\phi$ to have a continuous third derivative.) 
The above equation implies  
\begin{align}\label{eq:elab-critical}
\elab = - \frac 2 {\chi_{1, 2}} \frac 1 l + o(\frac 1 l).  
\end{align}
Then
\begin{align}
    \dllab  &= q^{(l+1)}_{ab} - q^*  + \sws \dot \T(q^* + \elab)\pabl - l q^* 
% \\  & \approx \ellab + (\chi_1 + \chi_2 \elab +  \frac 1 2 \chi_3 (\elab)^2) (lq^* + \dlab) - lq* 
% \\ &  \approx \ellab + (1 + \chi_2\elab ) \dlab + lq^*\chi_2 \elab +  \frac 1 2 \chi_3 (\elab)^2 lq^*  
\\  & = \ellab + (\chi_1 + \chi_{1, 2} \elab +  \bigo (\elab)^2) (lq^* + \dlab) - lq* 
\\ &  = \ellab + (1 + \chi_{1, 2}\elab ) \dlab + lq^*\chi_{1, 2} \elab +  \bigo ((\elab)^2) lq^*  
\end{align}
Plugging \eqref{eq:elab-critical} into the above equation gives 
\begin{align}
     \dlab = -\frac 2 3 l q^* + \bigo(1)  \, .  \label{eq:cricital-correcton-appendix}
\end{align}  
\subsubsection{The Spectrum of NNGP and NTK}
For $\Kl$, $\qabl = \qstar + \bigo(l^{-1})$ and $\ql =\qstar$. Thus  
\begin{align}
    \lmax(\Kl) &=  m\qstar + \bigo(1/l) \\ 
    \lrest(\Kl) &= \bigo(1/l) \\
    \kappa(\Kl) &\gtrsim l   
\end{align}
For $\Thetal$, $\pabl = \frac 1 3 \qstar l + \bigo(1)$ and $\pl = l\qstar$. Thus 
\begin{align}
    \lmax(\Thetal) &=  \frac {m+2}{3} l\qstar + \bigo(1) \\ 
    \lrest(\Thetal) &= \frac 2 3 l \qstar + \bigo(1) \\
    \kappa(\Thetal) & = \frac {m+2} {2} + \bigo(1/l) 
\end{align}

\section{NNGP and NTK of Relu networks.}
\label{sec:RElu-networks}
\subsection{Critical Relu.}
We only consider the critical initialization (i.e. He's initialization \citep{HeZR015}) $\sws =2$ and $\sbs=0$, which preserves the norm of an input from layer to layer. We also normalize the inputs to have unit variance, i.e. $\qstar =  q^{(l)}= q^{(0)}=1$.   
Recall that
\begin{align}
    \Kll & = 2 \T (\Kl) 
    \\
    \Thetall &=  \Kll +  2\dot \T (\Kl) \odot \Thetal
\end{align}
This implies 
\begin{align}
    \pll = \ql +  2\dot \T (\ql) \pl = 1 + 2\dot \T(1) \pl = 1 + \pl   
\end{align}
which gives $\pl =l$.  
Using the equations in Appendix C of~\citep{lee2019wide} gives 
\begin{align}
    2\T(1 - \epsilon) = 1  - \epsilon + \frac{2\sqrt 2}{3\pi} \epsilon^{3/2} + \bigo(\epsilon^{5/2})
\end{align}
and taking the derivative w.r.t. $\epsilon$ 
\begin{align}
    2 \dT(1 - \epsilon) = 1 - \frac{\sqrt 2}{\pi} \epsilon^{1/2} + \bigo(\epsilon^{3/2}) \quad {\textrm {as}} \q \epsilon \to 0^+. 
\end{align}
Thus  
\begin{align}
    1 - \ellab = 1 -\elab  +  \frac{2\sqrt 2}{3\pi} (\elab)^{3/2} + \bigo((\elab)^{5/2})
\end{align}
This is enough to conclude (similar to the above calculation) 
\begin{align}
    \elab &= (\frac {3\pi}{\sqrt 2})^{2} l^{-2} + o(l^{-2})
\end{align}
and 
\begin{align}
    \pabl - \pl &= -\frac 3 4 l  + \bigo(1). 
\end{align}
Recall that the diagonals of $\Kl$ and $\Thetal$ are $\ql = 1$ and $\pl = l$, resp. Therefore the spectrum and the condition numbers of $\Kl$ and $\Thetal$ for large $l$ are
\begin{align}
\label{eq:relu-condition-number-nngp-ntk}
\begin{cases}
    \lmax(\Thetal)  &= \frac {m+3} 4 l  + \bigo(1)
    \\
    \lrest(\Thetal) &= \frac 3 4 l+ \bigo(1)
    \\
    \kappa(\Thetal) &= \frac {m +3 }3 + \bigo(1/l)  
\end{cases}\quad \quad \quad 
\hfill
\begin{cases}
    \lmax(\Kl)  &=  m  + \bigo(l^{-2})
    \\
    \lrest(\Kl) &= \bigo(l^{-2})
    \\
    \kappa(\Kl) &\gtrsim \bigo(l^2)  
\end{cases}
\end{align}
\subsection{Residual Relu}
\label{sec:residual-relu}
We consider the following ``continuum'' residual network
\begin{align}
    x^{(t+dt)} = x^{(t)} + (dt)^{1/2} (W\phi(x^{(t)}) + b) 
\end{align}
where $t$ denotes the `depth' and $dt>0$ is sufficiently small and $W$ and $b$ are the weights and biases. We also set $\sws=2$ (i.e. $\mathbb E [WW^T] = 2 \Id$) and $\sbs=0$ (i.e. $b=0$). The NNGP and NTK have the following form 
\begin{align}
    \K^{(t+dt)} &=  \K^{(t)} + 2 dt \T(\K^{(t)}) 
    \\
    \Theta^{(t+dt)} &=  \Theta^{(t)} + 2dt \T(\K^{(t)}) + 2 dt  \dT(\K^{(t)}) \odot \Theta^{(t)}
\end{align}
Taking the limit $dt\to 0$ gives 
\begin{align}
    \dot\K^{(t)} &= 2  \T(\K^{(t)}) 
    \\
    \dot\Theta^{(t)} &=  2 \T(\K^{(t)}) + 2\dT(\K^{(t)})\odot \Theta^{(t)} \label{eq:pabdot }
\end{align}
Using the fact that $q^{(0)} = 1$ (i.e. the inputs have unit variance), we can compute the diagonal terms $q^{(t)} = e^t$ and $p^{(t)} = t e^t$. Letting  $q_{ab}^{(t)} = e^tc_{ab}^{(t)}$ and applying the above fractional Taylor expansion to $\T$ and $\dot \T$, we have
\begin{align}
    \dot c_{ab}^{(t)} = -  \frac{2\sqrt 2}{3\pi} (1 - c_{ab}^{(t)})^{\frac 3 2 } + O((1 - c_{ab}^{(t)})^{\frac 5 2 })
\end{align}
Ignoring the higher order term and set $y(t) = (1 - c_{ab}^{(t)})$, we have 
\begin{align}
    \dot y = \frac{2\sqrt 2}{3\pi} y^{\frac 3 2 } \, .
\end{align}
Solving this gives $y(t) = \frac{9\pi^2}{2} t^{-2}$ (note that  $y(\infty)=0$), which implies
\begin{align}
    q^{(t)}_{ab} &= (1 - \frac{9\pi^2}{2} t^{-2} + o(t^{-2})) e^t \, . 
\end{align}
 Applying this estimate to \eqref{eq:pabdot } gives      
\begin{align}
    p^{(t)}_{ab} &= (\frac 1 4 t + \bigo(1))e^t \,.
\end{align}
Thus the limiting condition number of the NTK is $m/3+1$. This is the same as the above non-residual Relu case although the entries of $\K^{(t)}$ and $\Theta^{(t)}$  blow up exponentially with $t$.

\subsection{Residual Relu + Layer Norm}
\label{sec:resilual-relu-layer-norm}
As we saw above, all the entries of $\Kl$ and $\Thetal$ of a residual Relu network blow up exponentially, so do its gradients. 
In what follows, we show that normalization could help to avoid this issue. 
We consider the following ``continuum'' residual network with ``layer norm" 
\begin{align}
    x^{(t+dt)} =  \frac 1  {\sqrt {1 + dt}}
    \left (x^{(t)} + (dt)^{1/2} W\phi(x^{(t)})\right) 
\end{align}
We also set $\sws=2$ (i.e. $\mathbb E [WW^T] = 2 \Id$).
The normalization term $\frac 1  {\sqrt {1 + dt}}$ makes sure $x^{(t+dt)}$ has unit norm and removes the exponentially factor $e^t$ in both NNGP and NTK. To ses this, note that 
\begin{align}
    \K^{(t+dt)} &=  \frac 1  {1 + dt} \left(\K^{(t)} + 2 dt \T(\K^{(t)}) \right ) 
    \\
    \Theta^{(t+dt)} &=  \frac 1  {1 + dt} \left( \Theta^{(t)} + dt\K^{(t)} + 2 dt  \dT(\K^{(t)}) \Theta^{(t)}\right) 
\end{align}
Taking the limit $dt\to 0$ gives 
\begin{align}
    \dot\K^{(t)} &= - \K^{(t)} + 2  \T(\K^{(t)}) 
    \\
    \dot\Theta^{(t)} &= 2\T (\K^{(t)}) + 2\dT(\K^{(t)})\odot \Theta^{(t)}
\end{align}
Using the fact that $q^{(0)} = 1$ (i.e. the inputs have unit variance)
and the mapping $2\T$ is norm preserving, we see that $q^{(t)} =1$ because  
\begin{align}
    \dot q^{(t)} &= - q^{(t)} + 2  \T(q^{(t)}) =  0 . 
\end{align}
This implies $p^{(t)}=t$ (note that $\dot p^{(t)}= q^{(t)}=1$ and we assume the initial value $p^{(0)}=0$.)
The off-diagonal terms can be computed similarly and 
\begin{align}
    q^{(t)}_{ab} &= 1 - \frac{9\pi^2}{2} t^{-2} + o(t^{-2})
    \\
    p^{(t)}_{ab} &= \frac 1 4 t + \bigo(1) \,.
\end{align}
Thus the condition number of the NTK is $m/3+1$. This is the same as the non-residual Relu case discussed above.  
\section{Asymptotic of $\Deltal$}\label{sec:deltal}
To keep the notation simple, we denote
$X_d = X_{\text{train}}$, $Y_d =Y_{\text{train}} $, $\Theta_{td} = \Theta_{\text{test, train}} $, $\Theta_{dd} = \Theta_{\text{train, train}} $. Recall that  
\begin{align}
    \Deltal Y_d
    = \left(\Thetal_{td} \left(\Thetal_{dd}\right)^{-1}
    \right)Y_d 
\end{align}
We split our calculation into three parts.  
\subsection{Chaotic phase}\label{sec:deltal-chaotic}
In this case the diagonal $\pl$ diverges exponentially and the off-diagonals $\pabl$ converges to a bounded constant $\pabstar$.
We further assume the input labels are centered in the sense $Y_d$ contains the same number of positive (+1) and negative (-1) labels\footnote{When the number of classes is greater than two, we require $Y_d$ to have mean zero along the batch dimension for each class.}.
We expand $\Thetal$ about its ``fixed point" 
\begin{align}
    \Deltal Y_d &= \Thetal_{td} \left(\Thetal_{dd}\right)^{-1} Y_d\\
    &= \left(\Theta^*_{td} + \bigo(\dlab) \right)
    \left(\pl \Id +  \pabstar (\bm1\bm1^T - \Id) + \bigo(\dlab) \right)^{-1}Y_d
\\
&= (\pl)^{-1}\left(\Theta^*_{td} + \bigo(\dlab) \right)
    \left(\Id - \frac {\pabstar}{\pl }(\bm1\bm1^T - \Id) + \bigo(\dlab/\pl ) \right)Y_d
    \\
&= (\pl)^{-1}\left(\Theta^*_{td} + \bigo(\dlab) \right)
    \left(\Id - \frac {\pabstar}{\pl }(\bm1\bm1^T - \Id) + \bigo(\dlab/\pl ) \right)Y_d
    \\
    &= (\pl)^{-1}\left(\bigo(\dlab) + \bigo(\dlab/\pl ) \right)Y_d
\end{align}
In the last equation, we have used the fact $\bm1\bm1^TY_d={\bf 0}$ and $\Theta_{td}^* Y_d ={\bf 0}$ since $Y_d$ is balanced. Therefore \begin{align}
    \Deltal Y_d = \bigo((\pl)^{-1}\dlab) = \bigo(l(\chic/\chi_1)^l) \,. 
\end{align}
\begin{remark}
Without centering the labels $Y_d$ and normalizing each input in $X_d$ to have the same variance, we will get a $\chi_1^{l}$ decay for $\Deltal Y_d$ instead of $l(\chic/\chi_1)^l$.  
\end{remark}
\subsection{Critical line}
Note that in this phase, both the diagonals and the off-diagonals diverge linearly. In this case
\begin{align}
    \lim_{l\to\infty} \frac{1}{lq^*} \Thetal_{td} = \frac 1 3 \bm1_t\bm1_d^{T}
    \quad \lim_{l\to\infty} \frac{1}{lq^*} \Thetal_{dd}= B\equiv  \frac 2 3 \Id + \frac 1 3 \bm1_d\bm1_d^{T}
\end{align}
Here we use $\bm1_d$ to denote the all `1' (column) vector with length equal to the number of training points in $X_d$ and $\bm1_t$ is defined similarly. Note that the constant matrix $B$ is invertible. By \eqref{eq:cricital-correcton-appendix} 
\begin{align}
    P(\Thetal) &= \frac 1 3 \left(\frac 3 {l\qstar}\Thetal_{td}\right) \left(\frac 1 {l\qstar}\Thetal_{dd}\right)^{-1}
    \\
    &= \frac 1 3\left(\bm1_t\bm1_d^T + \bigo(1/l\qstar) \right)
    \left(B 
    + \bigo(1/l\qstar) \right)^{-1}
    \\
    &= 
    \frac 1 3\left(\bm1_t\bm1_d^T + \bigo(1/l\qstar) \right)
    \left(B^{-1} 
    + \bigo(1/l\qstar) \right)
    \\
    &=
    \frac 1 3 \bm1_t\bm1_d^T B^{-1} 
    + \bigo(1/l\qstar)
\end{align}
The term $\bm1_t\bm1_d^T B^{-1} $ is independent of the inputs and  
$\bm1_t\bm1_d^T B^{-1} Y_{d} = 0 $ when $Y_d$ is centered. 
Thus 
\begin{align}
    \Deltal Y_d = \bigo(1/l\qstar)
\end{align}

\subsection{Ordered Phase}\label{sec:ordered-phase-mean-predictor}
In the ordered phase, we have that $\Thetal_{dd} = p^*\bm 1_d\bm 1^T_d + l \chi_1^l \bm A^{(l)}_{dd}$ where $\bm A^{(l)}_{dd}$, a symmetric matrix, represents the data-dependent piece of $\Thetal_{dd}$.
By Lemma \ref{lemma:ordered}, $\bm A^{(l)}_{dd}\to
\bm A_{dd}$ as $l\to\infty$. To simply the notation, in the calculation below we will replace $\bm A^{(l)}_{dd}$ by $\bm A_{dd}$.  
We also assume $\bm A_{dd}$ is invertible. 
To compute the mean predictor, $P(\Thetal)$, asymptotically we begin by computing $(\Thetal_{dd})^{-1}$ via the Woodbury identity,
\begin{align}
    (\Thetal_{dd})^{-1} &= \left(p^*\bm 1_d\bm 1^T_d + l\chi_1^l \bm A_{dd}\right)^{-1}\\
                    &=l^{-1}\chi_1^{-l}\left[\bm A_{dd}^{-1} - \bm A_{dd}^{-1}\bm 1_d \left(\frac1{p^*} + \frac{\bm 1_m^T\bm A_{dd}^{-1}\bm 1_d}{l \chi_1^l}\right)^{-1}\bm 1_d^T \bm A_{dd}^{-1}l^{-1}\chi_1^{-1}\right]\\  
                    &=l^{-1}\chi_1^{-l}\left[\bm A_{dd}^{-1} - \hat p \bm A_{dd}^{-1}\bm 1_d\bm 1_d^T \bm A_{dd}^{-1} \right]
                    \\
                    &= l^{-1}\chi_1^{-l}\left[\bm A_{dd}^{-1} - \hat p \bm a \bm a^T \right]
\end{align}
where we have set 
\begin{align}
     \bm a = \bm A_{dd}^{-1}\bm 1_d \quad {\rm and} \quad 
     \hat p = \frac{p^*}{l\chi_1^l + p^*\bm 1_d^T\bm A_{dd}^{-1}\bm 1_d} = \frac{p^*}{l\chi_1^l + p^*\bm 1_d^T\bm a}
\end{align}
and $\bm a_i = \frac 1m\sum_j \bm A_{ij}^{-1}$. Noting that $\Thetal_{td} = p^*\bm 1_t\bm 1_d^T + l \chi_1^l\bm A_{td}$ we can compute the mean predictor,
\begin{align}
    P(\Thetal) &= \Theta_{td}^l(\Theta_{dd}^l)^{-1} = (p^*\bm 1_t\bm 1_d^T + l\chi_1^l\bm A_{td})l^{-1}\chi_1^{-l}\left[\bm A_{dd}^{-1} - \hat p\bm a \bm a ^T\right]
    \\
    & = \bm A_{td} \bm A_{dd}^{-1}  - \hat p \bm A_{td}\bm a \bm a ^T +  
    l^{-1}\chi_1^{-l}p^* (\bm 1_t\bm 1_d^T \bm A_{dd}^{-1}  -  \hat p\bm 1_t\bm 1_d^T \bm a \bm a ^T) \\ 
    & = \bm A_{td} \bm A_{dd}^{-1}  - \hat p \bm A_{td}\bm a \bm a ^T +  
    l^{-1}\chi_1^{-l}p^* (1 - \hat p\bm 1_d^T \bm a )\bm 1_t\bm a^T   \\ 
    & = \bm A_{td} \bm A_{dd}^{-1}  - \hat p \bm A_{td}\bm a \bm a ^T +  
    \hat p\bm 1_t\bm a^T
\end{align}
Note that there is no divergence in $P(\Thetal)$ as $l\to\infty$ and the limit is well-defined.
The term $\hat p\bm 1_t\bm a^T$ is independent from the input data. 
\begin{equation}
    \lim_{l\to\infty}\Deltal Y_{\text{train}} = (\bm A_{td}\bm A_{dd}^{-1}- \hat p \bm A_{td}\bm a \bm a ^T +  
    \hat p\bm 1_t\bm a^T)Y_{\text{train}}
    \equiv (\bm A_{td}\bm A_{dd}^{-1} + \hat {\bm A}) Y_{\text{train}}
\end{equation}
We therefore see that even in the infinite-depth limit the mean predictor retains its data-dependence and we expect these networks to be able generalize indefinitely.

\section{Dropout}\label{sec:dropout}
\begin{figure}[h]
    \centering
    \includegraphics[width=0.5\textwidth]{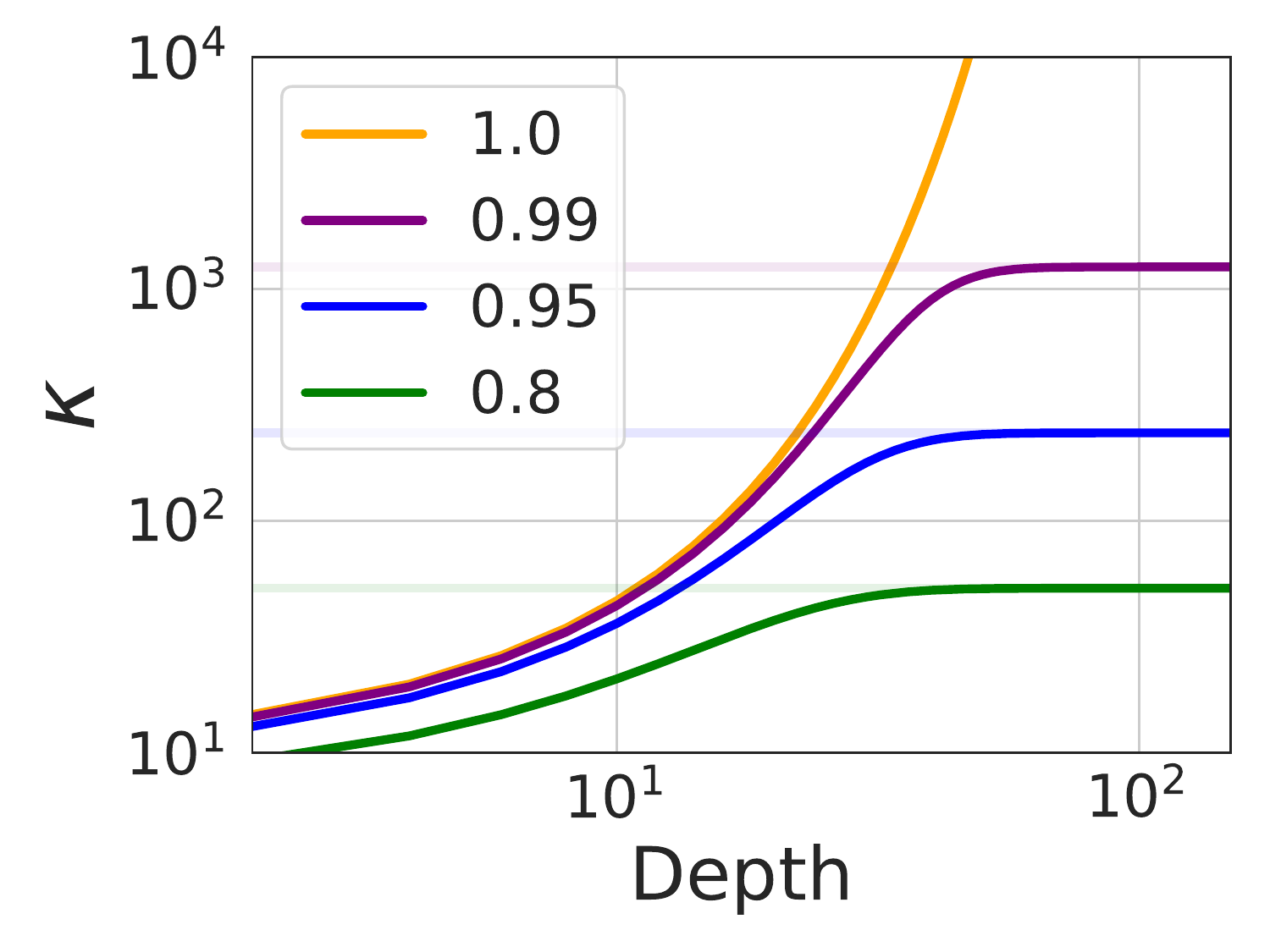}
    \caption{{\bf Dropout improves conditioning of the NTK.} In the ordered phase, the condition number $\kappal$ explodes exponentially (yellow) as $l\to\infty$. However, a dropout layer could significantly improves the conditioning, making $\kappal$ converge to a finite constant (horizontal lines) \eqref{eq:cond-dropout}.} 
    \label{fig:kappa-dropout}
\end{figure}

In this section, we investigate the effect of adding a dropout layer to the penultimate layer. Let $0<\rho\leq 1$ and $\gammal_j(x)$ be iid random variables
\begin{align}
    \gammal_j(x) = 
    \begin{cases} 1, \quad \text{with probability} \quad  \rho
    \\
    0,   \quad \text{with probability} \quad 1 - \rho . 
    \end{cases}
\end{align}
For $0\leq l\leq L-1$, 
\begin{equation}\label{eq:fc_def_recap_recursion_app}
    z^{(l+1)}_i(x) = \frac{\sigma_w}{\sqrt {N^{(l)}}}\sum_jW^{(l+1)}_{ij}\phi(z^{(l)}_j(x)) + \sigma_b b^{(l+1)}_i 
\end{equation}
and for the output layer, 
\begin{equation}\label{eq:fc_def_recap_recursion_app_2}
    z^{(L+1)}_i(x) = \frac{\sigma_w}{\rho\sqrt {\NL}}\sum_{j=1}^{\NL}\WL_{ij}\phi(z^{(L)}_j(x))\gammal_j(x) + \sigma_b b^{(L+1)}_i \hspace{3pc}
\end{equation}
where $W^{(l)}_{ij}$ and $ b^{(l)}_i$ are iid Gaussians $\N(0, 1)$. 
Since no dropout is applied in the first $L$ layers, the NNGP kernel $\Kl$ and $\Thetal$ can be computed using \eqref{eq:fc_mft_recap_recursion} and \eqref{eq: NTK-recursive}. Let  $\K_\rho^{(L+1)}$ and $\Theta_\rho^{(L+1)}$ denote the NNGP and NTK of the $(L+1)$-th layer. Note that when $\rho=1$, $\K_1^{(L+1)}= \K^{(L+1)}$ and $\Theta_{1}^{(L+1)} = \Theta^{(L+1)}$\,. We will compute the correction induced by $\rho<1$. 
The fact 
\begin{align}
\mathbb E [\gammal_j(x)\gammal_i(x') ] =  
\begin{cases} \rho^2, \quad \text{if} \quad  (j, x) \neq (i, x')
    \\
    \rho,   \quad \text{if} \quad (j, x) = (i, x') 
    \end{cases}
\end{align}
implies that the NNGP kernel $\K^{(L+1)}_{\rho}$ \citep{schoenholz2016} is 
\begin{align}
\K^{(L+1)}_{\rho}(x, x') \equiv 
    \mathbb E [z^{(L+1)}_i(x)z^{(L+1)}_i(x') ] =  
\begin{cases} \sigma_w^2 \mathcal T(\K^{(L)}(x, x')) + \sigma_b^2, \quad \text{if} \quad  x \neq  x'
    \\
    \\
    \frac 1 \rho\sigma_w^2 \mathcal T(\K^{(L)}(x, x)) + \sigma_b^2  \quad \text{if} \quad  x = x' \,.
    \end{cases}
\end{align}
Now we compute the NTK $\Theta^{(L+1)}_\rho$, which is a sum of two terms 
\begin{align}\label{eq:dropout-}
\Theta^{(L+1)}_\rho(x, x') =  
    \mathbb E \left[ \frac{\partial z^{(L+1)}_i(x)}{\partial \theta^{(L+1)} } \left(\frac {\partial z^{(L+1)}_i(x')} {\partial \theta^{(L+1)}}\right)^T\right ] 
    +
     \mathbb E \left[ \frac{\partial z^{(L+1)}_i(x)}{\partial \theta^{(\leq L)} } \left(\frac {\partial z^{ (L+ 1 )}_i(x')} {\partial \theta^{(\leq L)}}\right)^T\right ] . 
\end{align}
Here $\theta^{(L+1)}$ denote the parameters in the $(L+1)$ layer, namely, $W_{ij}^{(L+1)}$ and $b^{(L+1)}_i $ and $\theta^{(\leq L)}$ the remaining parameters. Note that the first term in \eqref{eq:dropout-} is equal to $\K^{(L+1)}_\rho(x, x')$. Using the chain rule, the second term is equal to 
\begin{align}
   &\frac{\sws}{\rho^2 {N^{(L)}} } \mathbb E \left [ \sum_{j, k=1}^{{N^{(L)}}}W^{(L+1)}_{ij}W^{(L+1)}_{ik}\dot \phi(z^{(L)}_j(x))\gammal_j(x) \dot \phi(z^{(L)}_k(x'))\gammal_j(x')
    \frac {\partial z^{(L)}_j(x)}{\partial \theta^{(\leq L)}}
    \left(\frac {\partial z^{(L)}_k(x')}{\partial \theta^{(\leq L)}}\right)^T
    \right ]
    \\
    =
    &\frac{\sws}{\rho^2 {N^{(L)}} } \mathbb E \left [ \sum_{j}^{{N^{(L)}}}\dot \phi(z^{(L)}_j(x))\gammal_j(x) \dot \phi(z^{(L)}_j(x'))\gammal_j(x')
    \frac {\partial z^{(L)}_j(x)}{\partial \theta^{(\leq L)}}
    \left(\frac {\partial z^{(L)}_j(x')}{\partial \theta^{(\leq L)}}\right)^T
    \right ]
    \\
    = 
    &\frac{\sws}{\rho^2 } 
    \mathbb E \left [\gammal_j(x) \gammal_j(x')
    \right ] 
    \mathbb E [\dot \phi(z^{(L)}_j(x))\dot \phi(z^{(L)}_j(x'))]
    \mathbb E \left[ \frac {\partial z^{(L)}_j(x)}{\partial \theta^{(\leq L)}}
    \left(\frac {\partial z^{(L)}_j(x')}{\partial \theta^{(\leq L)}}\right)^T\right ]
    \\
    =
    & \begin{cases} 
    \sws \dot \T(\K^{(L)}(x, x')) \Theta^{(L)}(x, x')   \quad {\rm if } \quad x\neq x'
        \\ 
        \\
        \frac 1 \rho \sws \dot \T(\K^{(L)}(x, x)) \Theta^{(L)}(x, x) 
         \quad {\rm if } \quad x =  x' \,. 
    \end{cases}
\end{align}
In sum, we see that dropout only modifies the diagonal terms %# of $\Theta^{(L+1)}$
\begin{align}
\begin{cases}
\Theta^{(L+1)}_\rho(x, x') = \Theta^{(L+1)}(x, x')
\\
\\
 \Theta^{(L+1)}_\rho(x, x) = \frac 1 \rho \Theta^{(L+1)}(x, x)
+ (1-1/\rho)\sigma_b^2
\end{cases}  
\end{align}
In the ordered phase, 
we see 
\begin{align}
    \lim_{L\to\infty}\Theta^{(L)}_\rho(x, x')  = \pstar, 
    \quad\quad 
    \lim_{L\to\infty}\Theta^{(L)}_\rho(x, x)  = \frac 1 \rho \pstar + (1-\frac 1 \rho) \sbs 
\end{align}
and the condition number 
\begin{align}\label{eq:cond-dropout}
    \lim_{L\to\infty} \kappa^{(L)}_\rho = \frac {(m-1)\pstar + \frac 1 \rho \pstar + (1-\frac 1 \rho) \sbs }{(\frac 1 \rho - 1)(\pstar - \sbs) } = \frac {m\pstar}{(\frac 1 \rho - 1)(\pstar - \sbs) } + 1 
\end{align}
In Fig~\ref{fig:kappa-dropout}, 
we plot the evolution of $ \kappa^{(L)}_\rho$ for $\rho=0.8, 0.95, 0.99$ and $1$, confirming \eqref{eq:cond-dropout}.  
\section{Convolutions}\label{sec:conv-ntk}
In this section, we compute the evolution of $\Thetal$ for CNNs.  

        \textbf{General setup.} For simplicity of presentation we consider 1D convolutional networks with circular padding as in \citet{xiao18a}. We will see that this reduces to the fully-connected case introduced above if the image size is set to one and as such we will see that many of the same concepts and equations carry over schematically from the fully-connected case. The theory of two-or higher-dimensional convolutions proceeds identically but with more indices.

        \textbf{Random weights and biases.} The parameters of the network are the convolutional filters and biases, $\omega^{(l)}_{ij, \beta}$ and $\bparam^{(l)}_{i}$, respectively, with outgoing (incoming) channel index $i$ ($j$) and filter relative spatial location $\beta \in  [\pm k] \equiv  \{- k,\dots, 0, \dots,  k\}$.\footnote{We will use Roman letters to index channels and Greek letters for spatial location. We use letters $i, j, i', j'$, etc to denote channel indices, $\alpha, \alpha'$, etc to denote spatial indices and $\beta, \beta'$, etc for filter indices.} As above, we will assume a Gaussian prior on both the filter weights and biases, 
        \begin{align}
            W^{(l)}_{ij,\beta}  = \frac {\sw}{\sqrt{(2k+1)N^{(l)}}}
            \omega^{(l)}_{ij,\beta}  &
            & b^{(l)}_i &= \sbb \bparam^{(l)}_i, 
            & \omega^{(l)}_{ij,\beta}, \quad \mu^{(l)}_i \sim \N(0, 1) 
        \end{align}
        As above, $\sigma^2_\omega$ and $\sigma^2_b$ are hyperparameters that control the variance of the weights and biases respectively. $N^{(l)}$ is the number of channels (filters) in layer $l$, $2k+1$ is the filter size.
            
        \textbf{Inputs, pre-activations, and activations.} Let $\X$ denote a set of input images. The network has activations $\h^{(l)}(x)$ and pre-activations $z^{(l)}(x)$ for each input image $x\in\X \subseteq \mathbb{R}^{N^{(0)} d}$, with input channel count $N^{(0)} \in \mathbb{N}$, number of pixels $d \in \mathbb{N}$, where
        \begin{align}
           \h^{(l)}_{i,\alpha}(x) &\equiv
            \left\{\begin{array}{ccc}
            	x_{i, \alpha} &  & l=0 \\
            	\phi\left( z^{(l-1)}_{i,\alpha}(x) \right) &  & l > 0
            \end{array}\right.,
            &
            z^{(l)}_{i,\alpha}(x) &\equiv \sum_{j=1}^{N^{(l)}} \sum_{\beta = -k}^{k} W^{(l)}_{ij, \beta} \h^{(l)}_{j, \alpha + \beta}(x) + b^{(l)}_{i}.
            \label{eq weighted sum}
        \end{align}
        $\phi:\R\to\R$ is a point-wise activation function. Since we assume circular padding for all the convolutional layers, the spacial size $d$ remains constant throughout the networks until the readout layer.
        
        For each $l>0$, as $\min\{N^1\dots, N^{(l-1)}\}\to\infty$, for each $i\in\mathbb N$, the pre-activation converges in distribution to $d$-dimensional Gaussian with mean $\bf 0$ and covariance matrix $\Kl$, which can be computed recursively ~\citep{novak2018bayesian, xiao18a} 
        \begin{align}\label{eq:nngp kernel}
            &\Kll = (\sw^2\A + \sbs) \circ \T (\Kl)=   \left((\sw^2\A + \sbs) \circ \T\right)^{l+1} (\K^0) 
        \end{align}
        Here $\Kl \equiv [\Kl_{\alpha, \alpha'}(x, x')]_{\alpha, \alpha' \in[d], x, x'\in \X}$, $\T$ is a non-linear transformation related to its fully-connected counterpart, and $\A$ a convolution acting on $\X d \times \X d$ PSD matrices 
        \begin{align}
                \left[\T\left(\K\right)\right]_{\alpha, \alpha'}\left(x, x'\right) &\equiv \expect{
                        u\sim\N\pp{0, \K}
                    }{\phi\pp{
                    u_{\alpha}(x)
                }\phi\pp{
                    u_{\alpha'}(x')
                }} \label{eq:T_def}
                \\
                \left[\A\left(\K\right)\right]_{\alpha, \alpha'}\left(x, x'\right)
                &\equiv\frac{1}{2k+1}\sum_{\beta}\left[\K\right]_{\alpha +\beta, \alpha'+\beta}\left(x, x'\right).
                \label{eq:A_def}
            \end{align}
        \subsection{The Neural Tangent Kernel}
        To understand how the neural tangent kernel evolves with depth, we define the NTK of the $l$-th hidden layer to be $\hatthetal$
        \begin{align}
            \hatthetal_{\alpha, \alpha'}(x, x') = \nabla_{\theta^{\leq l}} z^{(l)}_{i,\alpha} (x)\left(\nabla_{\theta^{\leq l}} z^{l}_{i,\alpha'}(x') \right)^T
        \end{align}
        where $\theta^{\leq l}$ denotes all of the parameters in layers at-or-below the $l$'th layer. It does not matter which channel index $i$ is used because as the number of channels approach infinity, this kernel will also converge in distribution to a deterministic kernel $\Thetall$ \citep{yang2019scaling}, which can also be computed recursively in a similar manner to the NTK for fully-connected networks as \citep{yang2019scaling, arora2019exact},
        \begin{align}\label{eq: NTK-recursive-CNN}
        &\Thetall = \Kll +\A \circ (\sws  \dT (\Kl)\odot \Thetal),   
        \end{align}
        where $\dT$ is given by \eqref{eq:T_def} with $\phi$ replaced by its derivative $\phi'$. We will also normalize the variance of the inputs to $q^*$ and hence treat $\T$ and $\dT$ as pointwise functions. We will only present the treatment in the chaotic phase to showcase how to deal with the operator $\A$. The treatment of other phases are similar. Note that the diagonal entries of $\Kl$ and $\Thetal$ are exactly the same as the fully-connected setting, which are $\qstar$ and $\pl = l\qstar$, respectively. We only need to consider the off-diagonal terms. Letting $l\to\infty$ in \eqref{eq: NTK-recursive-CNN} we see that all the off-diagonal terms also converge $\pabstar$.
        Note that $\A$ does not mix terms from different diagonals and it suffices to handle each off-diagonal separately. Let $\elab$ and $\dlab$ denote the correction of the $j$-th diagonal of $\Kl$ and $\Thetal$ to the fixed points. Linearizing \eqref{eq:nngp kernel} and \eqref{eq: NTK-recursive-CNN} gives  
        \begin{align}
        \ellab &\approx \chic  \A \elab 
        \\
             \dllab &\approx \chic  \A (\ellab + \frac {\chi_{c^*,2}}{\chic}\pabstar \elab + \dlab) \,.
         \end{align}
         Next let $\{\rho_\alpha\}_\alpha$ be the eigenvalues of $\A$ and $\elaba$ and $\dlaba$ be the projection of $\elab$ and $\dlab$ onto the $\alpha$-th eigenvector of $\A$, respectively. Then for each $\alpha$, 
         \begin{align}
         \ellaba &\approx (\rho_\alpha\chic)^{(l+1)} \epsilon^{(0)}_{ab,\alpha}
         \\
             \dllaba &\approx \rho_\alpha\chic (\ellaba + \frac
            {\chi_{c,2}}{\chic}\pabstar \elaba + \dlaba)\, 
         \end{align}
        which gives
            \begin{align}\label{eq:off-diagonal-correction-cnn}
            \epsilon^{(l)}_{ab,\alpha} &\approx(\rho_\alpha\chic)^l\epsilon^{(0)}_{ab,\alpha}\,\,, 
            \\
            \delta^{(l)}_{ab, \alpha} &\approx (\rho_\alpha \chic)^l\left[\delta^{{(0)}}_{ab, \alpha} + l\left(1 + \frac{\chi_{c, 2}}{\chic}\pab\right)\epsilon^{(0)}_{ab,\alpha}\right]
            \end{align}
        Therefore, the correction $\Thetal - \Theta^*$ propagates independently through different Fourier modes. In each mode, up to a scaling factor $\rho_\alpha^l$, the correction is the same as the correction of FCN. Since the subdominant modes (with $|\rho_\alpha|<1$) decay exponentially faster than the dominant mode (with $\rho_{\alpha}=1$), for large depth, the NTK of CNN is essentially the same as that of FCN. 

    \subsection{The effect of pooling and flattening of CNNs}
    \label{sec:cnn-pooling}
    With the bulk of the theory in hand, we now turn our attention to CNN-F and CNN-P. We have shown that the dominant mode in CNNs behaves exactly like the fully-connected case, however we will see that the readout can significantly affect the spectrum. The NNGP and NTK of the $l$-th hidden layer CNN are 4D tensors 
    $\Kl_{\alpha, \alpha'}(x, x')$ and $\Thetal_{\alpha, \alpha'}(x, x') $, where $\alpha, \alpha' \in [d] \equiv [0, 1, \dots, d-1]$ denote the pixel locations. To perform tasks like image classification or regression, ``flattening'' and ``pooling'' (more precisely, global average pooling) are two popular readout strategies that transform the last convolution layer into the logits layer. The former strategy ``flattens'' an image of size $(d, N)$ into a vector in $\R^{dN}$ and stacks a fully-connected layer on top. The latter projects the $(d, N)$ image into a vector of dimension $N$ via averaging out the spatial dimension and then stacks a fully-connected layer on top. 
    The actions of ``flattening'' and ``pooling'' on the image correspond to computing the mean of the trace and the mean of the pixel-to-pixel covariance on the NNGP/NTK, respectively, i.e.,      
             \begin{align}
                 \Thetalft(x, x') &=\frac 1 d \sum_{\alpha\in [d]}\Thetal_{\alpha, \alpha}(x, x')\, , 
                 \\
                   \Thetalpool(x, x') &=\frac 1 {d^2} \sum_{\alpha, \alpha'\in [d]}\Thetal_{\alpha, \alpha'}(x, x') \, , 
             \end{align}
            where $\Thetalft$ ($\Thetalpool$)  denotes the NTK right after flattening (pooling) the last convolution. We will also use $\Thetalfc$ to denote the NTK of FC.  
            $\Klft$, $\Klpool$ and $\Kl_{\text fc}$ are defined similarly. 
    As discussed above, in the large depth setting, all the diagonals $\Thetal_{\alpha, \alpha}(x, x) = \pl$ (since the inputs are normalized to have variance $\qstar$ for each pixel) and similar to $\Thetalfc$, all the off-diagonals $\Thetal_{\alpha', \alpha}(x, x')$ are almost equal (in the sense they have the same order of correction to $\pabstar$ if exists.) Without loss of generality, we assume all off-diagonals are the same and equal to $\pabl$ (the leading correction of $\qabl$ for CNN and FCN are of the same order.)   
    Applying flattening and pooling, the NTKs become 
    \begin{align}
        &\Thetalft(x, x') = \frac 1 d \sum_{\alpha} \Thetal_{\alpha,  \alpha}(x, x') = \1_{x=x'}\pl +\1_{x\neq x'} \pabl \, , 
    \\
        &\Thetal_{\rm pool}(x, x') = \frac 1 {d^2} \sum_{\alpha, \alpha'} \Thetal_{\alpha,  \alpha'}(x, x') = \frac 1 {d}\1_{x=x'}(\pl - \pabl)
        + \pabl \, , 
    \end{align}
    respectively.  
    As we can see, $\Thetal_{\rm flatten}$ is essentially the same as its FCN counterpart $\Thetal_{\rm fc}$ up to sub-dominant Fourier modes which decay exponentially faster than the dominant Fourier modes. 
    Therefore the spectrum properties of $\Thetal_{\rm flatten}$ and $\Thetal_{\rm fc}$ are essentially the same for large $l$; see Figure \ref{fig: conditioning-complete} (a - c).  
   
    However, pooling alters the NTK/NNGP spectrum in an interesting way.
    Noticeably, the contribution from $\pl$ is discounted by a factor of $d$. 
    On the critical line, asymptotically, the on- and off-diagonal terms are  
    \begin{align}
        \Thetalpool(x, x) &= \frac{2 +d }{3d} l\qstar  + \bigo(1)
        \\
        \Thetalpool(x, x') &= \frac 1 3 l\qstar + \bigo(1) 
    \end{align}
    This implies 
    \begin{align}
     \lmaxl &= (m \colord + 2)q^*l/(3\colord)  +\bigo(1)    
    \\
    \lrestl &= 2\qstar l /(3\colord) +\bigo(1)
    \\
    \kappal &= \frac {m\colord +2}{2} + m\colord \bigo(l^{-1})
    \end{align}
     Here we use blue color to indicate the changes of such quantities against their $\Thetalft$ counterpart. Alternatively, one can consider $\Thetalft$ as a special version (with $\colord=1$) of $\Thetalpool$.  
     Thus pooling decreases $\lrestl$ roughly by a factor of $\colord$ and increases the condition number by a factor of $\colord$ comparing to flattening. In the chaotic phase, pooling does not change the off-diagonals $\qabl =\bigo(1)$ but does slow down the growth of the diagonals by a factor of $d$, i.e. $\pl = \bigo(\chi_1^l\color{blue}{/ d})$. This improves $\Deltal$ by a factor of $\colord$. This suggests, in the chaotic phase, there exists a transient regime of depths, where CNN-F hardly perform while CNN-P performs well. In the ordered phase, the pooling does not affect $\lmaxl$ much but does decrease $\lrestl$ by a factor of $\colord$ and the condition number $\kappal $ grows approximately like $\colord l\chi_1^{-l}$, $\colord$ times bigger than its flattening and fully-connected network counterparts. This suggests the existence of a transient regime of depths, in which CNN-F outperforms CNN-P. This might be surprising because it is commonly believed CNN-P usually outperforms CNN-F. These statements are supported empirically in Figure \ref{fig:pool vs flatten}.

    \section{Figure Zoo}
\begin{figure}[h]
            \begin{subfigure}[b]{.5\textwidth}
                \centering
              \includegraphics[width=.88\textwidth]{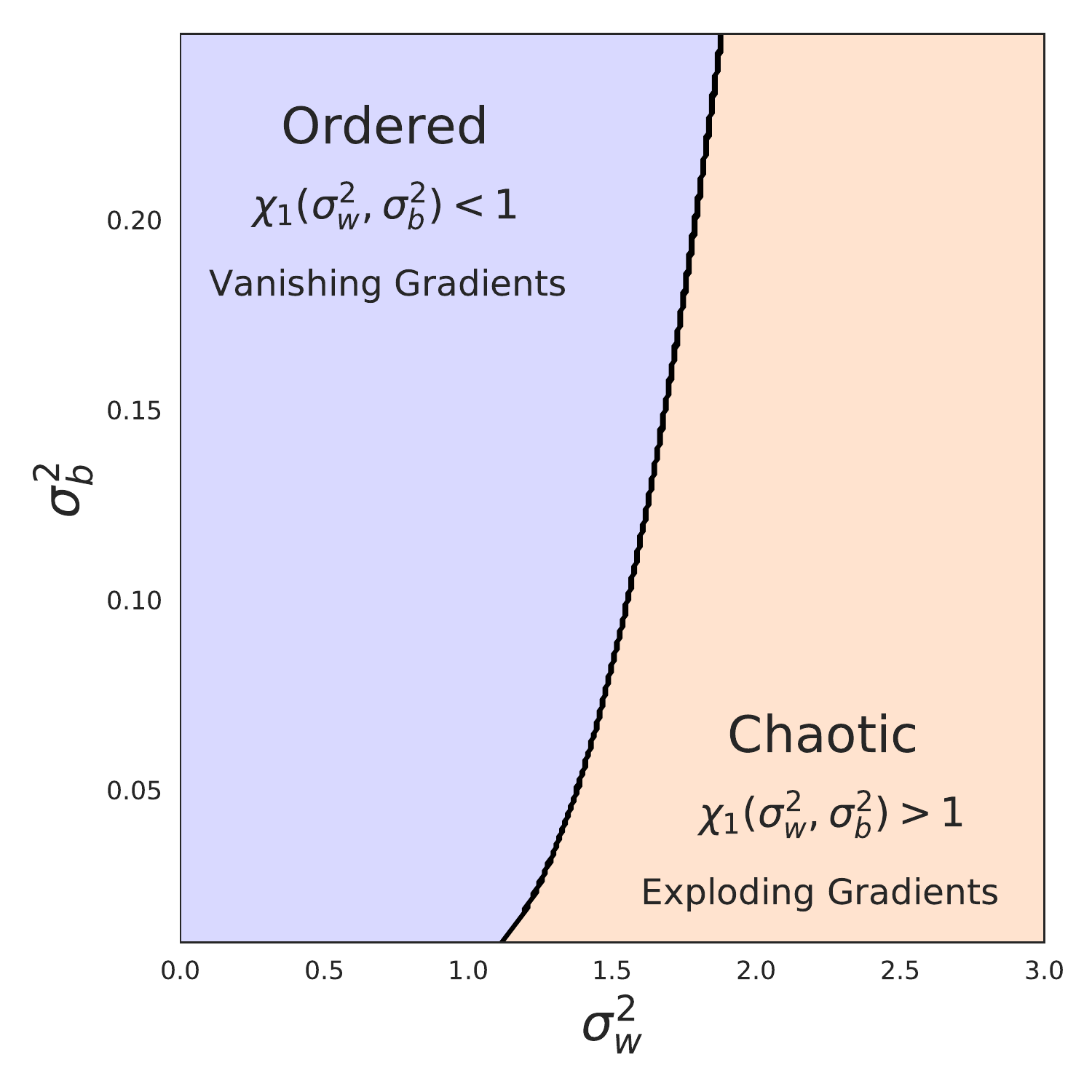}
        \caption{Phase Diagram for $\rm Erf$.} 
        \label{fig:phase-diagram-erf}
        \end{subfigure}%
    \begin{subfigure}[b]{0.5\textwidth}
                \centering
              \includegraphics[width=1.\textwidth]{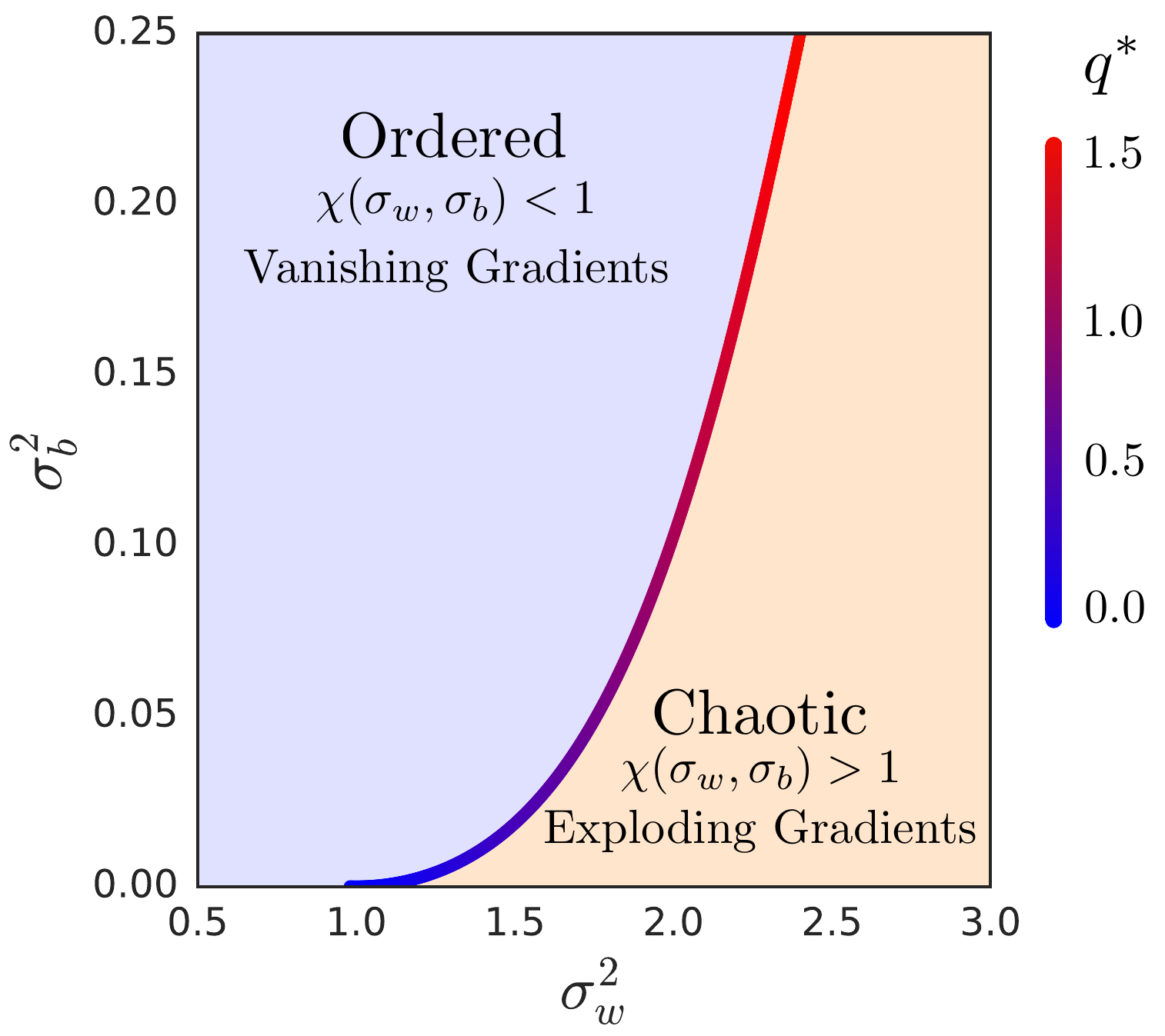}
        \caption{Phase Diagram for $\tanh$. } 
        \label{fig:phase-diagram-tanh}
        \end{subfigure}%
\caption{Phase Diagram for $\tanh$ and Erf (right).} 
        \label{fig:phase-diagram}
    \end{figure}

    \subsection{Phase Diagrams: Figure \ref{fig:phase-diagram}.}
\begin{figure}
        \centering
        \includegraphics[width=1.\textwidth]{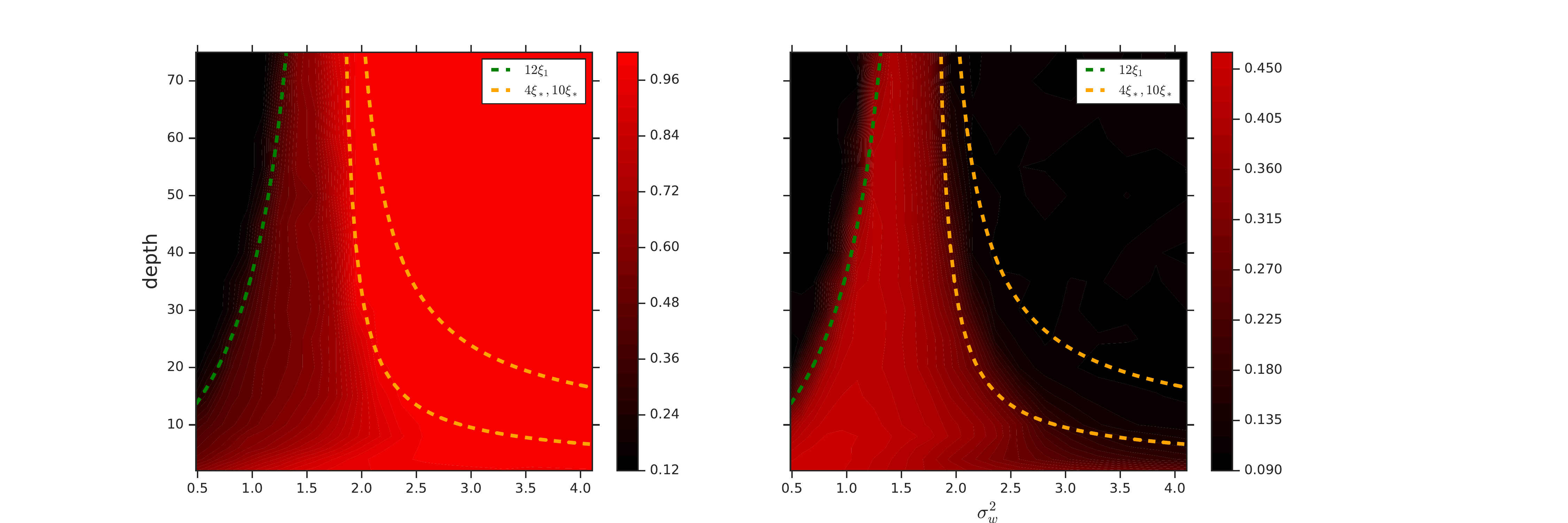}
        \caption{Training and Test Accuracy for FCN for different $(\sws, l)$ configurations.}
        \label{fig:train-test-acc-fcn}
    \end{figure}

    We plot the phase diagrams for the Erf function and the $\tanh$ function (adopted from  \cite{pennington2018emergence}).  
    \subsection{SGD on FCN on Larger Dataset: Figure~\ref{fig:train-test-acc-fcn}.}
    We report the training and test accuracy of FCN trained on a subset (16k training points) of CIFAR-10 using SGD with 20 $\times$ 20 different $(\sws, l)$ configurations. 

\begin{figure*}[h]
        \begin{center}
                    \begin{subfigure}[b]{1.\textwidth}         \includegraphics[width=1.2\textwidth]{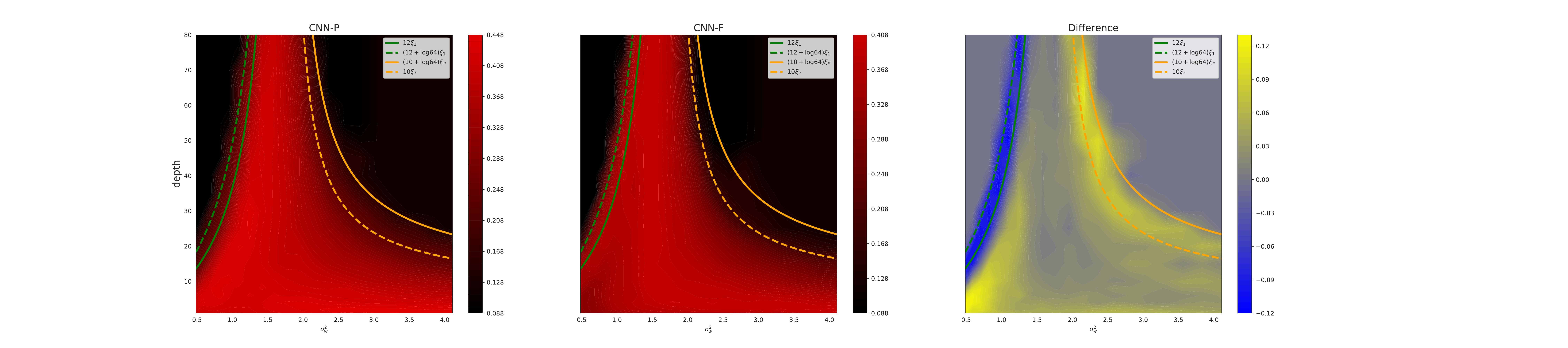}
        \end{subfigure}
        \\
        \begin{subfigure}[b]{1.\textwidth}                          \includegraphics[width=1.2\textwidth]{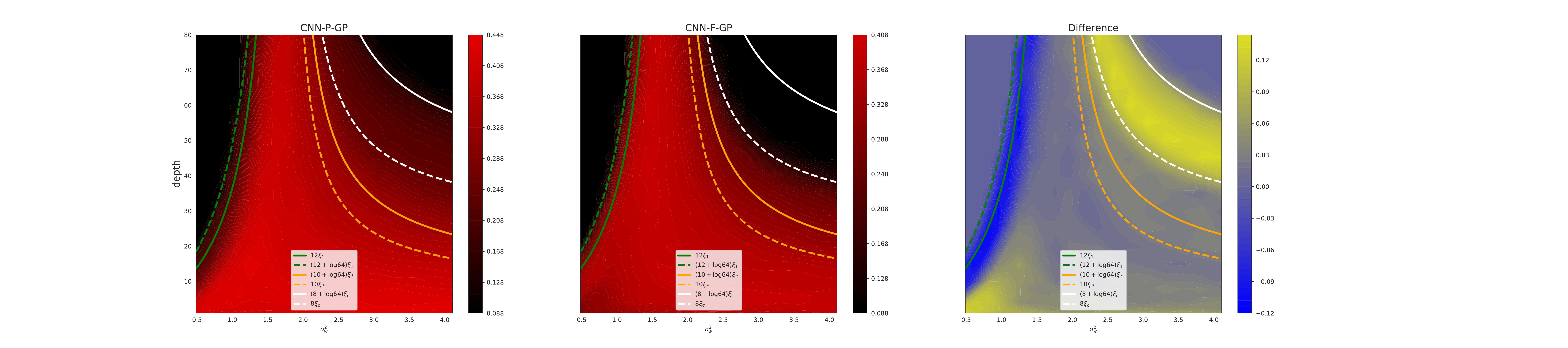}
        \end{subfigure}
                \end{center}
                 \caption{Test Accuracy for NTK (top) and NNGP prediction for different $(\sws, l)$ configurations. First/second column: CNN with/without pooling. Last column: difference between the first and second columns.}
                        \label{fig:gp-ntk-prediction}
    \end{figure*}

    \subsection{NNGP vs NTK prediction: Figure~\ref{fig:gp-ntk-prediction}.}
    Here we compare the test performance of the NNGP and NTK with different $(\sws, l)$ configurations. In the chaotic phase, the generalizable depth-scale of the NNGP is captured by $\xi_{c^*} =-1/ \log(\chic)$. In contrast, the generalizble depth-scale of the NTK is captured by $\xi_* = -1/ (\log(\chic) - \log(\chi_1))$. Since $\chi_1>1$ in the chaotic phase, $\xi_{c^*} > \xi_*$. Thus for larger depth, the NNGP kernel performs better than the NTK.  
    Corrections due to an additional average pooling layer is plotted in the third column of Figure~.\ref{fig:gp-ntk-prediction}

\begin{figure*}[h]
    \begin{center}
        \begin{subfigure}[b]{.49\textwidth}
                % \centering
\includegraphics[width=1.1\textwidth]{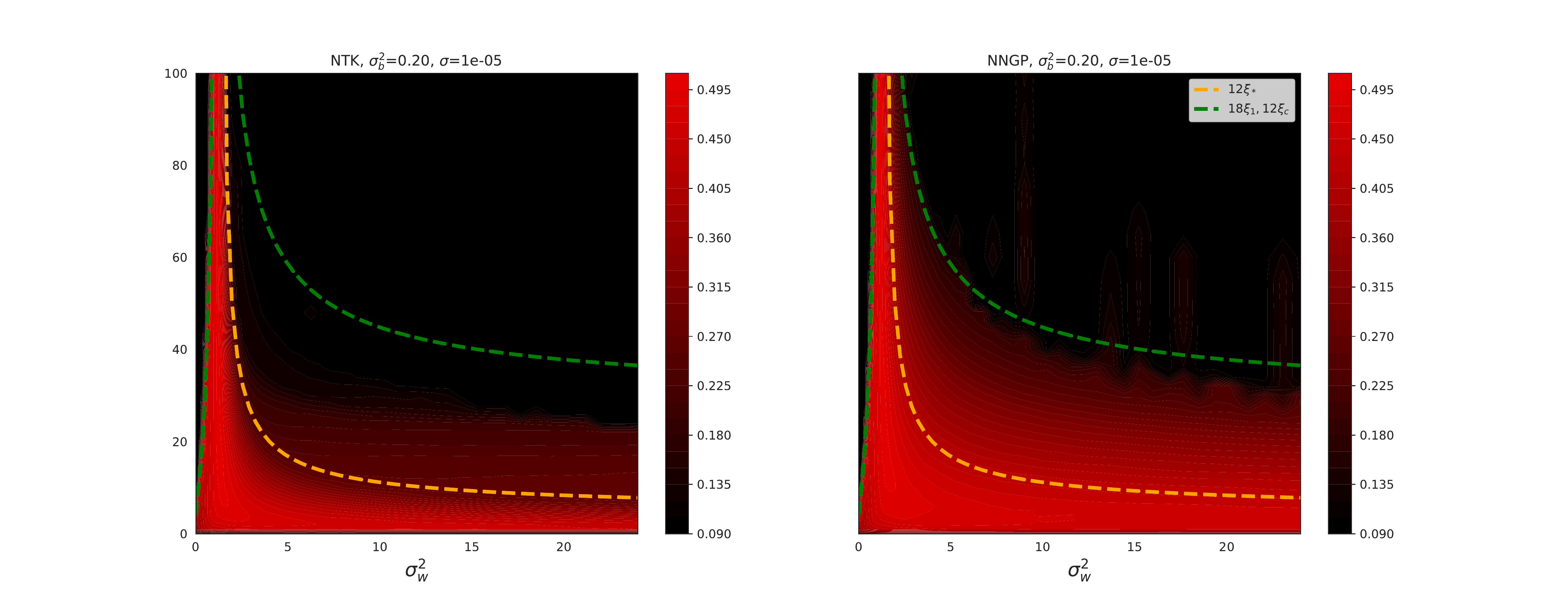}
                % \caption{NNGP Chaotic}
                % \label{fig:nngp-kappa-chaotic-fcn}
        \end{subfigure}
        \begin{subfigure}[b]{.49\textwidth}
                % \centering
               \includegraphics[width=1.1\textwidth]{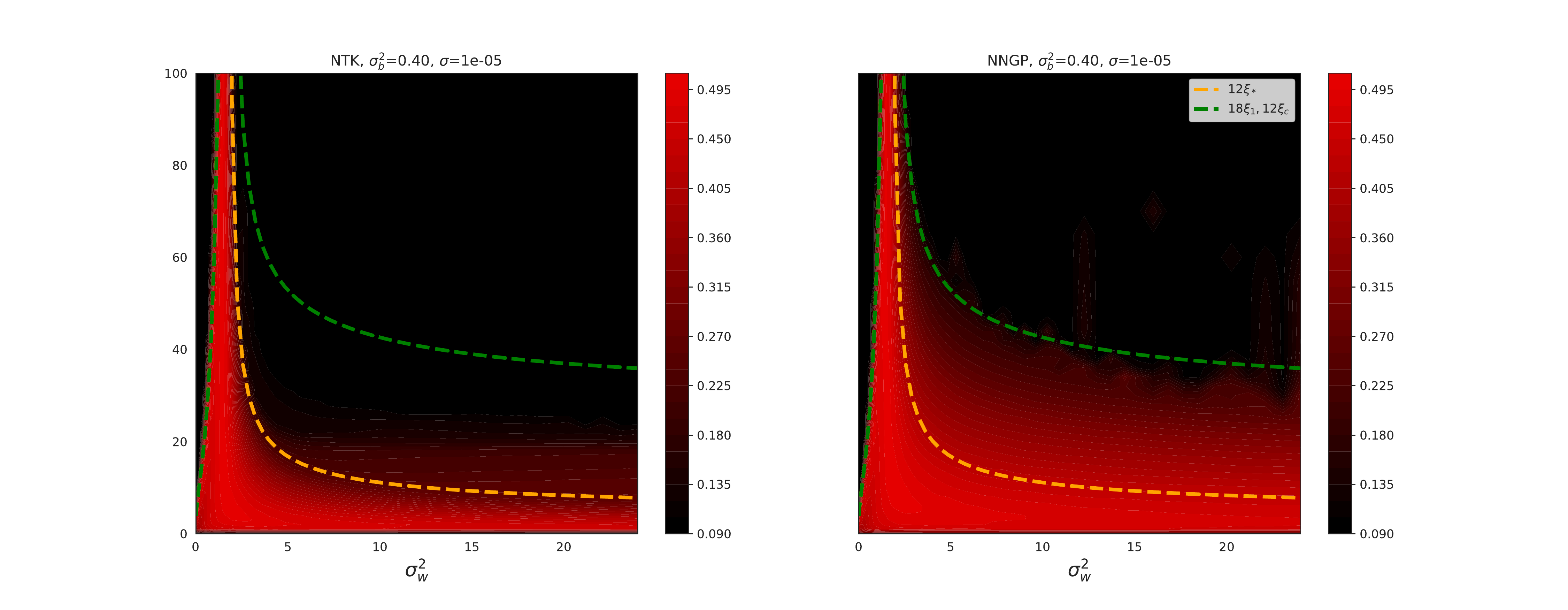}
                % \caption{NNGP Chaotic}
                % \label{fig:nngp-kappa-chaotic-cnn-p}
        \end{subfigure}%
        \\
        \begin{subfigure}[b]{0.49\textwidth}
                % \centering
              \includegraphics[width=1.1\textwidth]{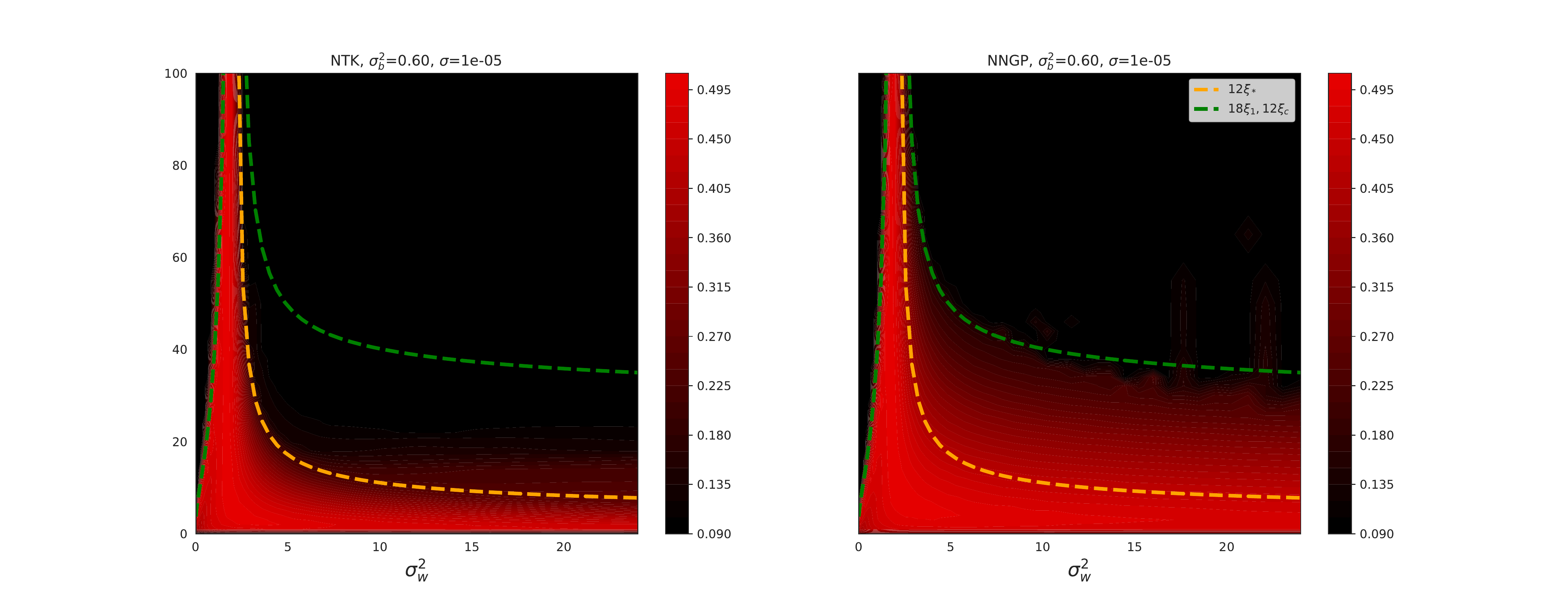}
                % \caption{NNGP Ordered}
                % \label{fig:nngp-kappa-ordered}
        \end{subfigure}%
 \begin{subfigure}[b]{0.49\textwidth}
                % \centering
     \includegraphics[width=1.1\textwidth]{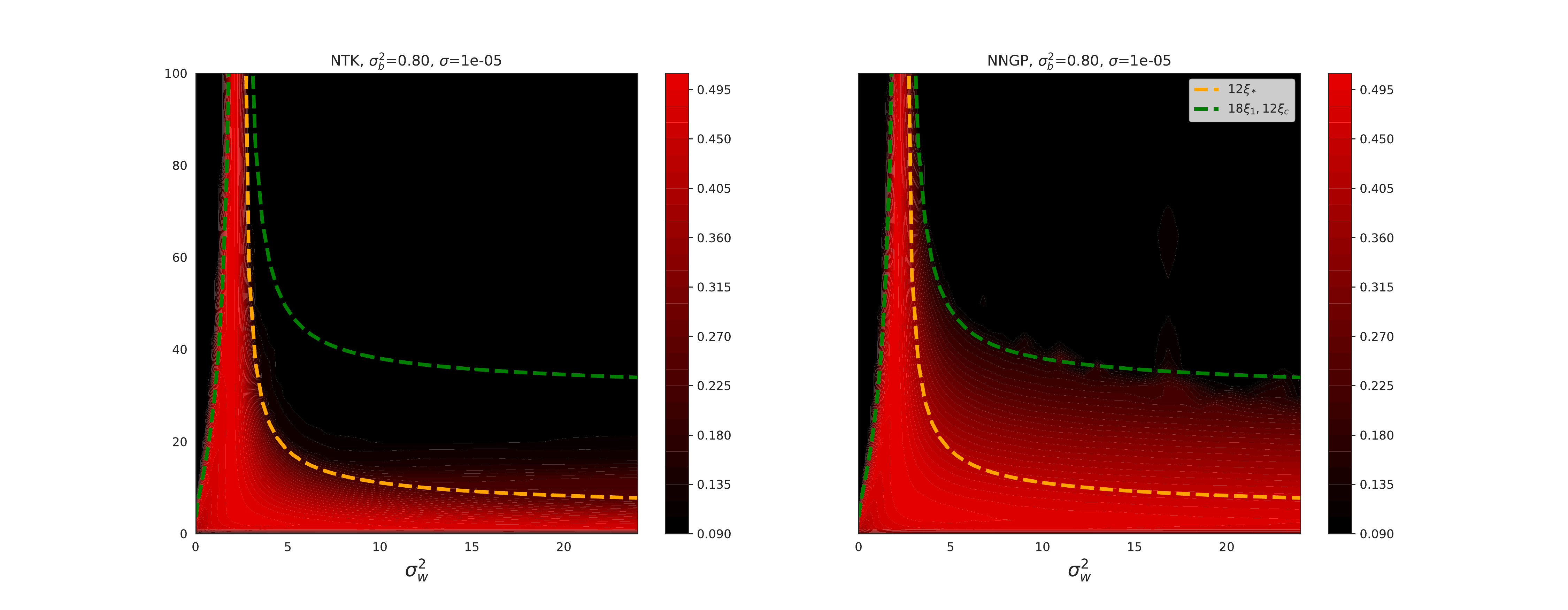}
                % \caption{NNGP Critical}
                % \label{fig:nngp-critical}
        \end{subfigure}%
        \\
        \begin{subfigure}[b]{0.49\textwidth}
                % \centering
              \includegraphics[width=1.1\textwidth]{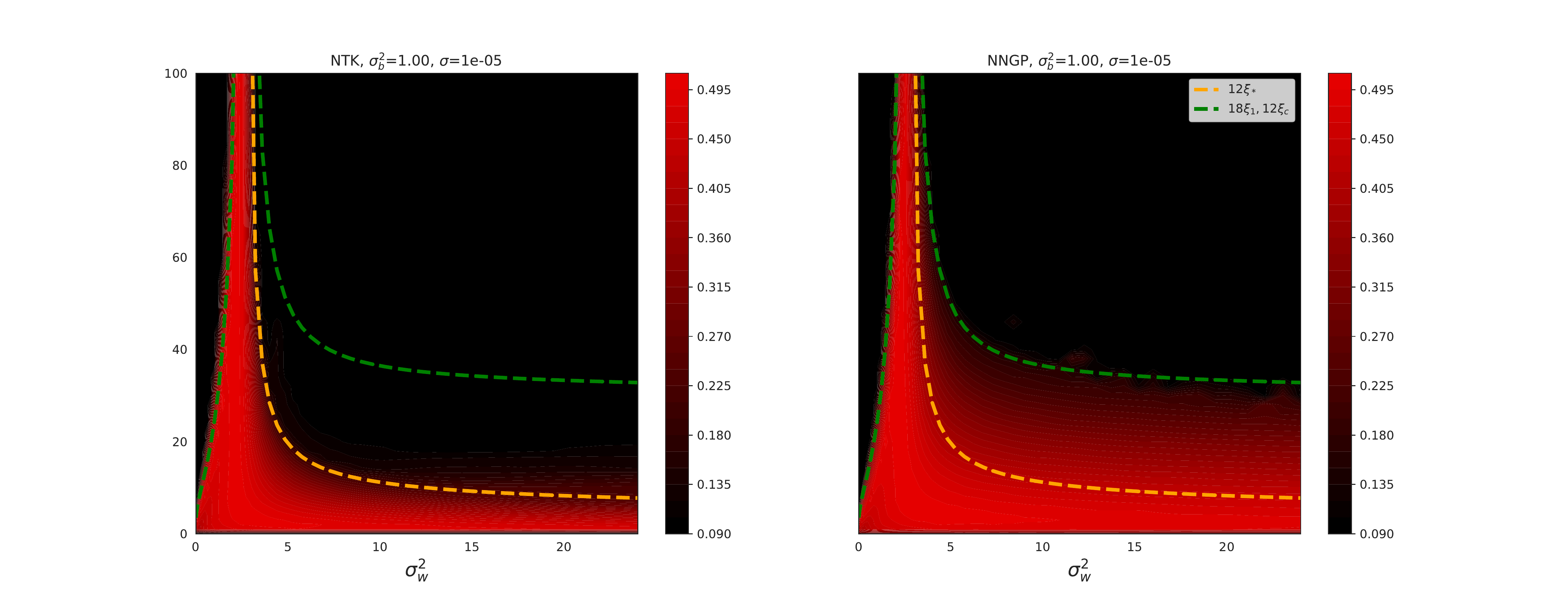}
                % \caption{NNGP RELU}
                % \label{fig:nngp-relu}
        \end{subfigure}%
 \begin{subfigure}[b]{0.49\textwidth}
                % \centering
          \includegraphics[width=1.1\textwidth]{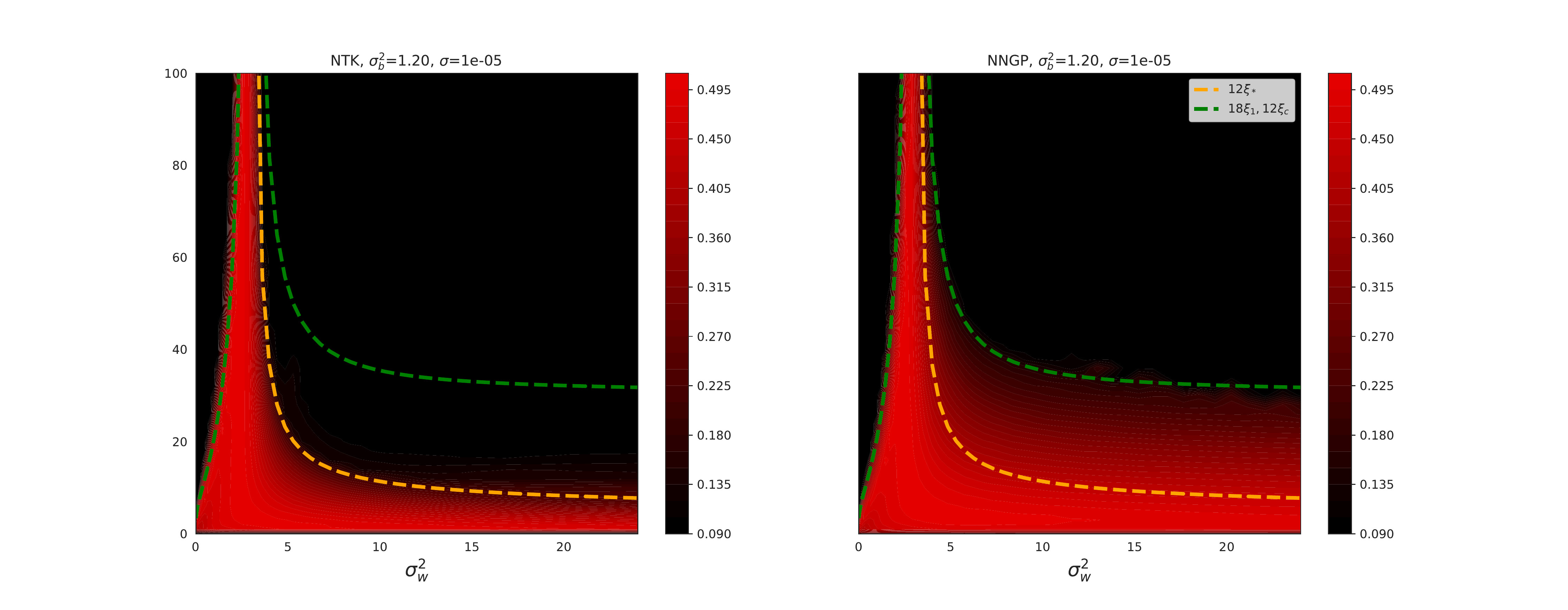}
                % \caption{NTK RELU}
                % \label{fig:relu-ntk}
        \end{subfigure}\\
        \begin{subfigure}[b]{0.49\textwidth}
                % \centering
              \includegraphics[width=1.1\textwidth]{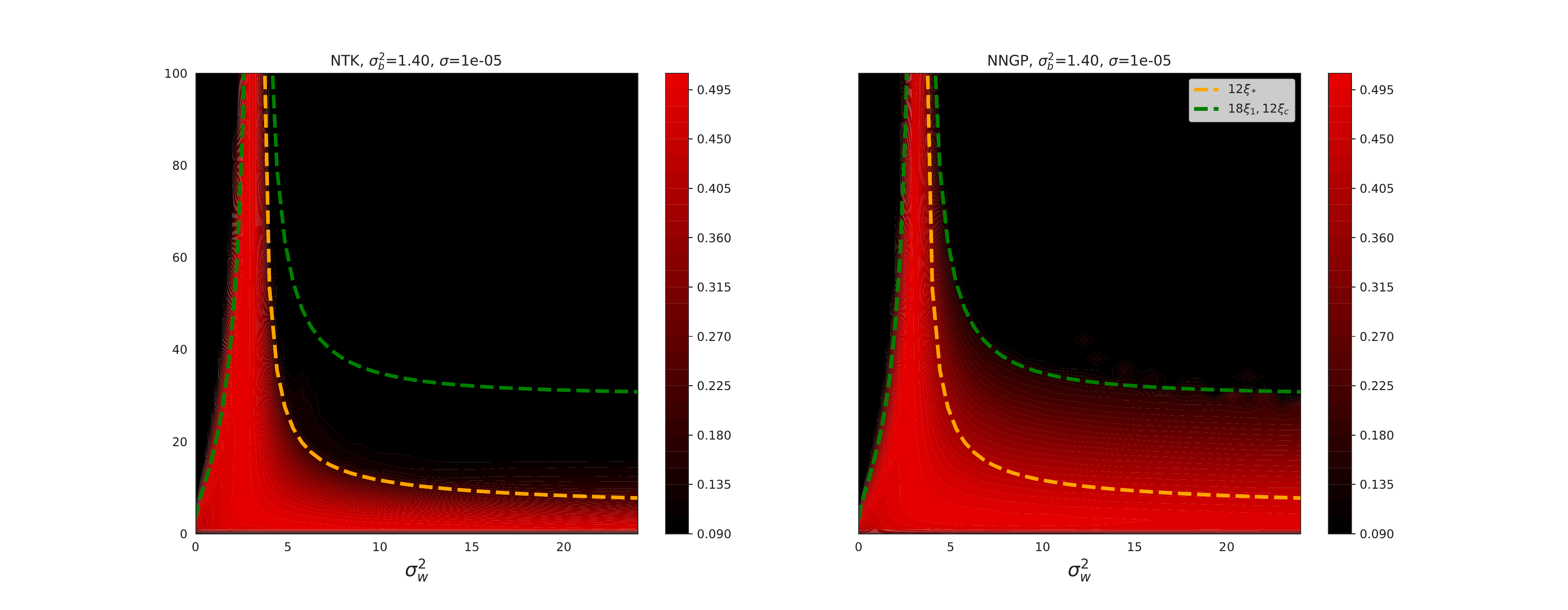}
                % \caption{NNGP RELU}
                % \label{fig:nngp-relu}
        \end{subfigure}%
 \begin{subfigure}[b]{0.49\textwidth}
                % \centering
          \includegraphics[width=1.1\textwidth]{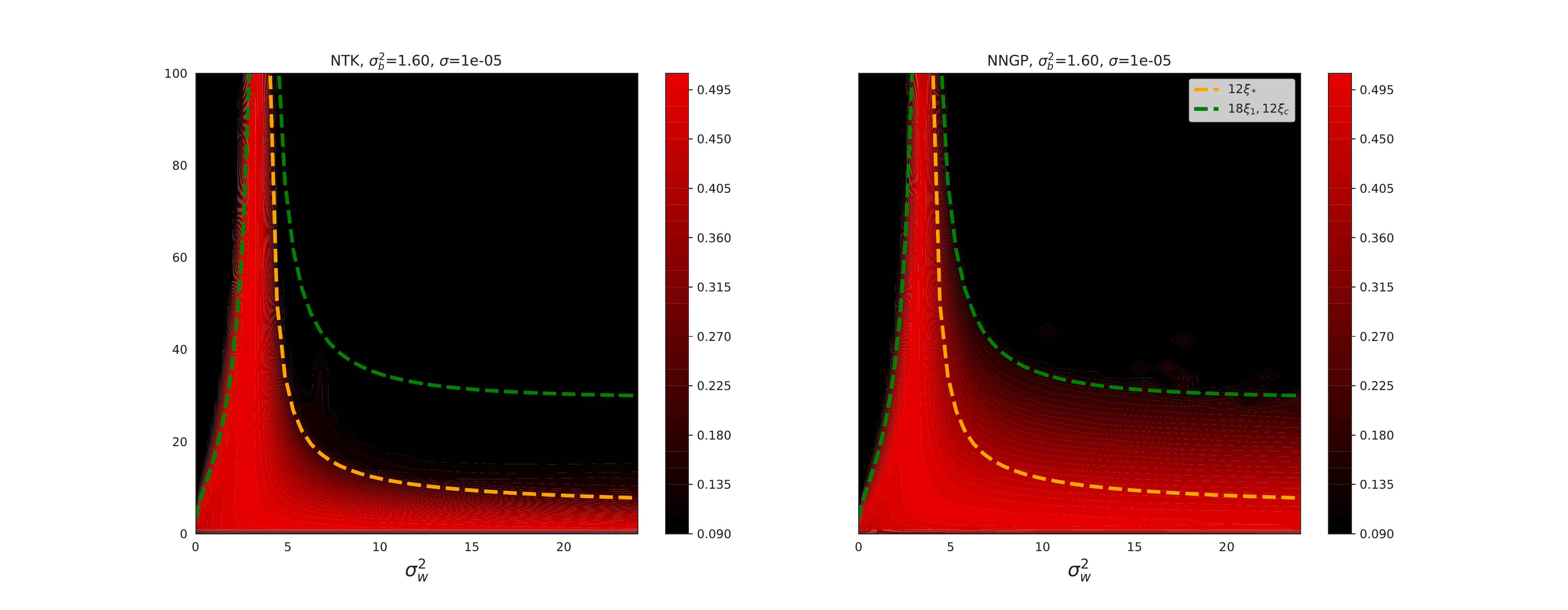}
                % \caption{NTK RELU}
                % \label{fig:relu-ntk}
        \end{subfigure}\\
        \begin{subfigure}[b]{0.49\textwidth}
                \centering
              \includegraphics[width=1.1\textwidth]{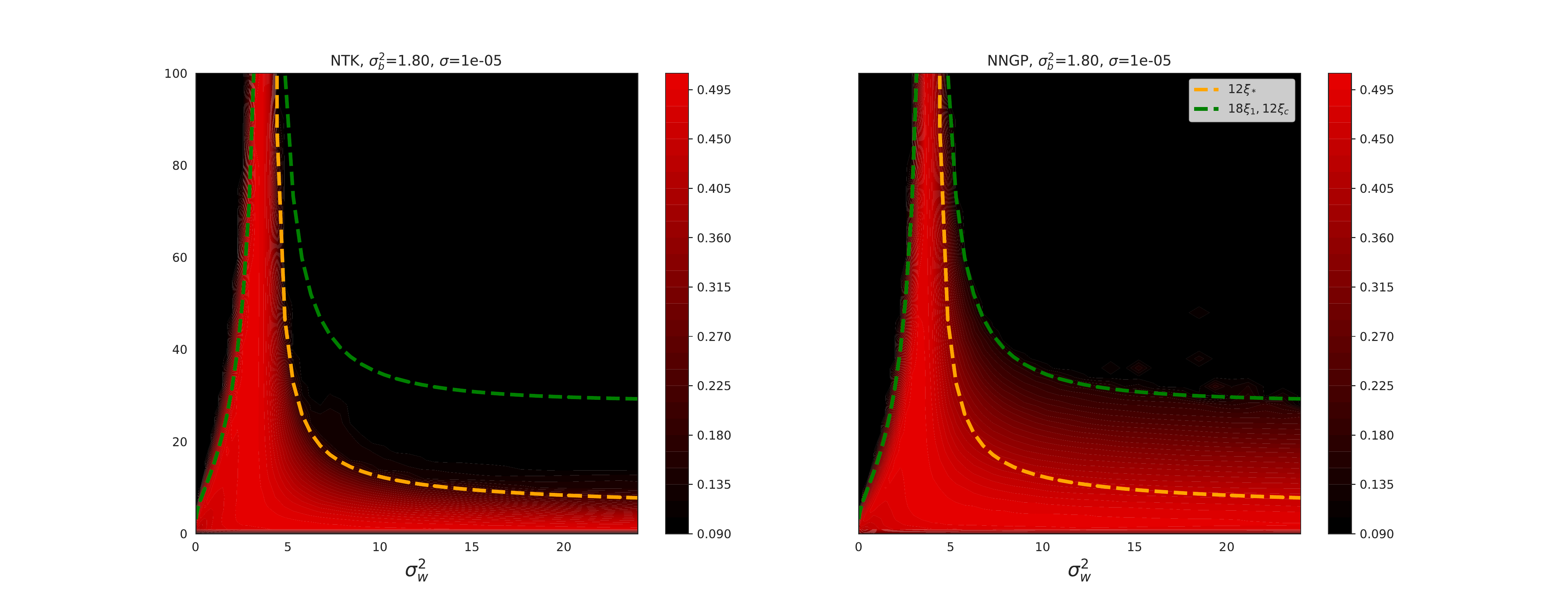}
                % \caption{NNGP RELU}
                % \label{fig:nngp-relu}
        \end{subfigure}%
\end{center}
\caption{{\bf Generalization metrics for NTK/NNGP vs Test Accuracy vs $\sbs$.} 
 }
\label{fig:varying-sb-generalization}
\end{figure*}

\subsection{Densely Sweeping Over $\sbs$: Figure~\ref{fig:varying-sb-generalization}}
\label{sec:varying-sb}
We demonstrate that our prediction for the generalizable depth-scales for the NTK ($\xi_*$) and NNGP ($\xi_c$) are robust across a variety of hyperparameters. We densely sweep over 9 different values of $\sbs\in [0.2, 1.8]$. For each $\sbs$ we compute the NTK/NNGP test accuracy for 20 * 50 different configurations of (l, $\sws$) with $l\in[1, 100]$ and $\sws\in [0.1^2, 4.9^2]$. The training set is a $8$k subset of CIFAR-10.  

\begin{figure*}[h]
    \begin{center}
        \begin{subfigure}[b]{.49\textwidth}
                % \centering
\includegraphics[width=1.1\textwidth]{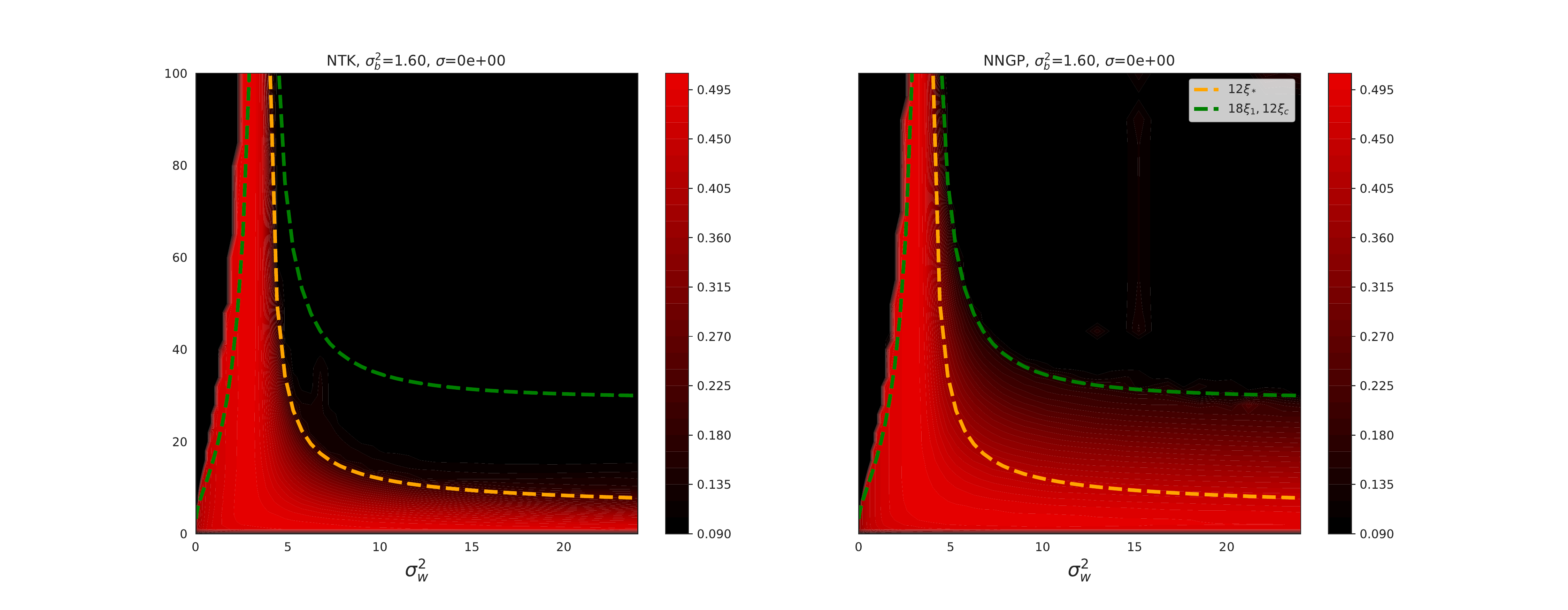}
        \end{subfigure}
        \begin{subfigure}[b]{.49\textwidth}
                % \centering
               \includegraphics[width=1.1\textwidth]{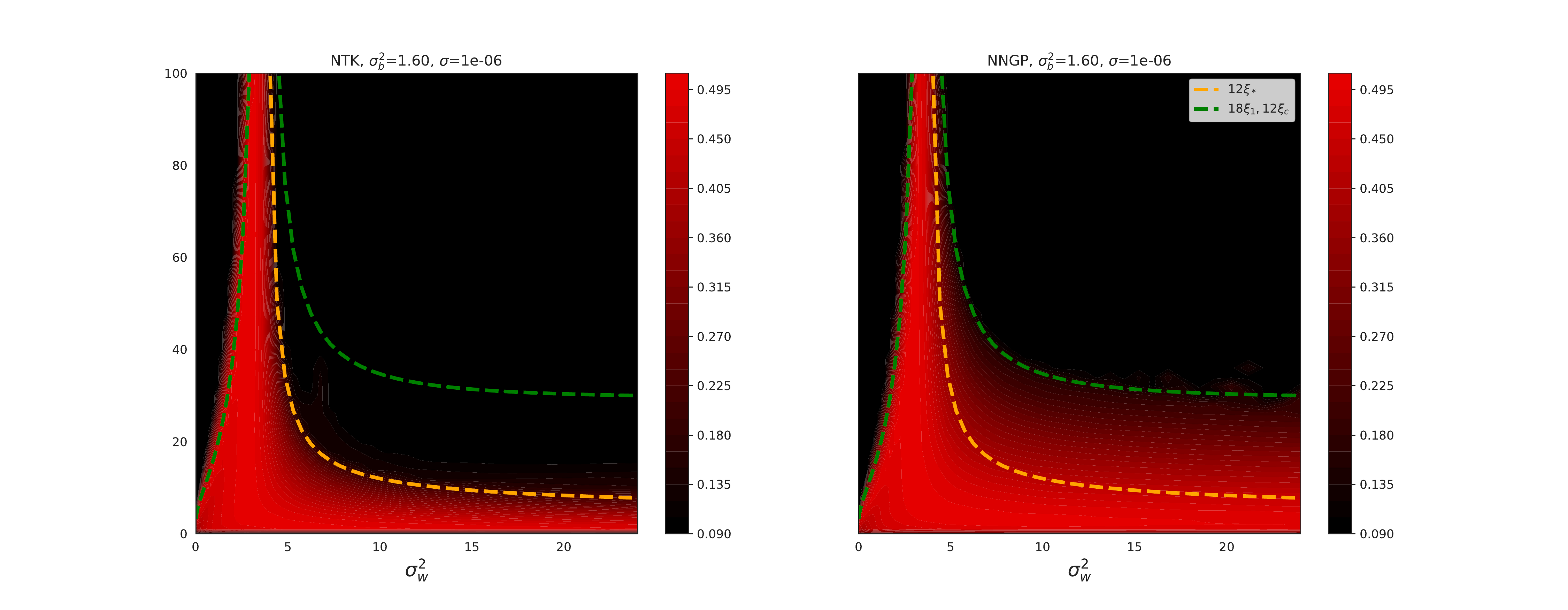}
        \end{subfigure}%
        \\
        \begin{subfigure}[b]{0.49\textwidth}
                % \centering
              \includegraphics[width=1.1\textwidth]{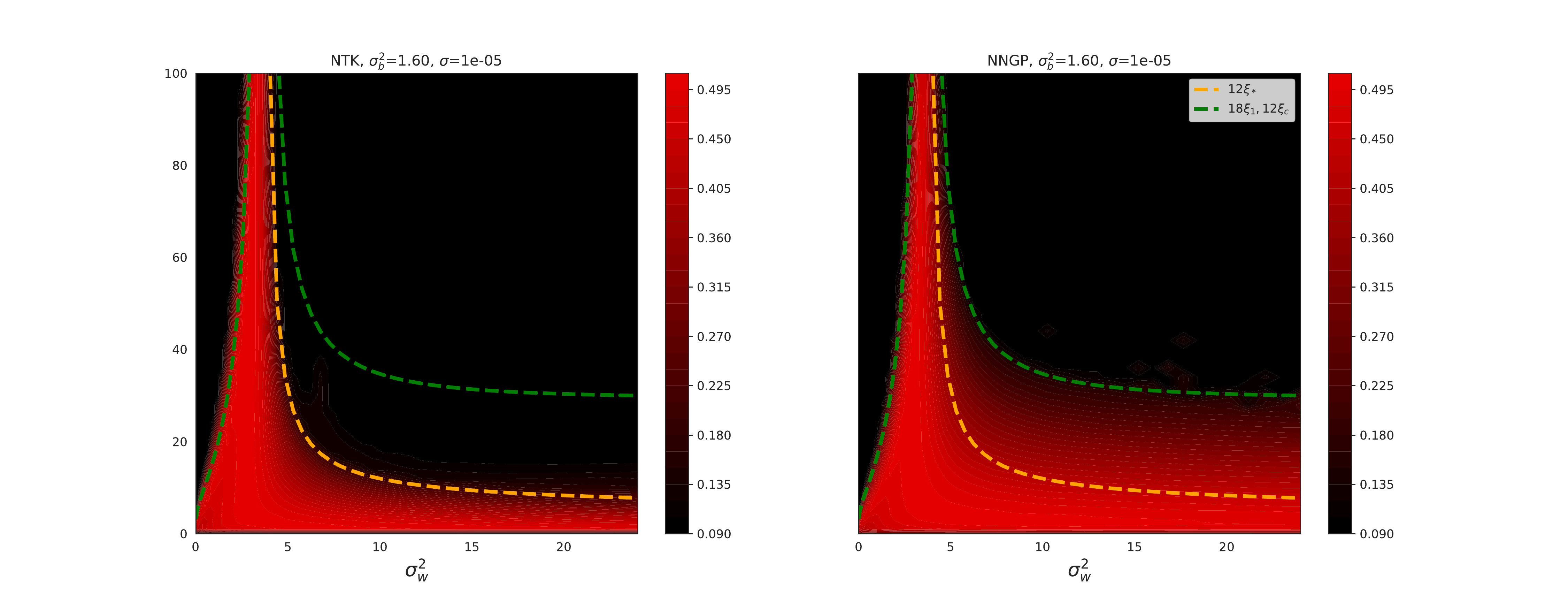}
        \end{subfigure}%
        \begin{subfigure}[b]{0.49\textwidth}
                % \centering
              \includegraphics[width=1.1\textwidth]{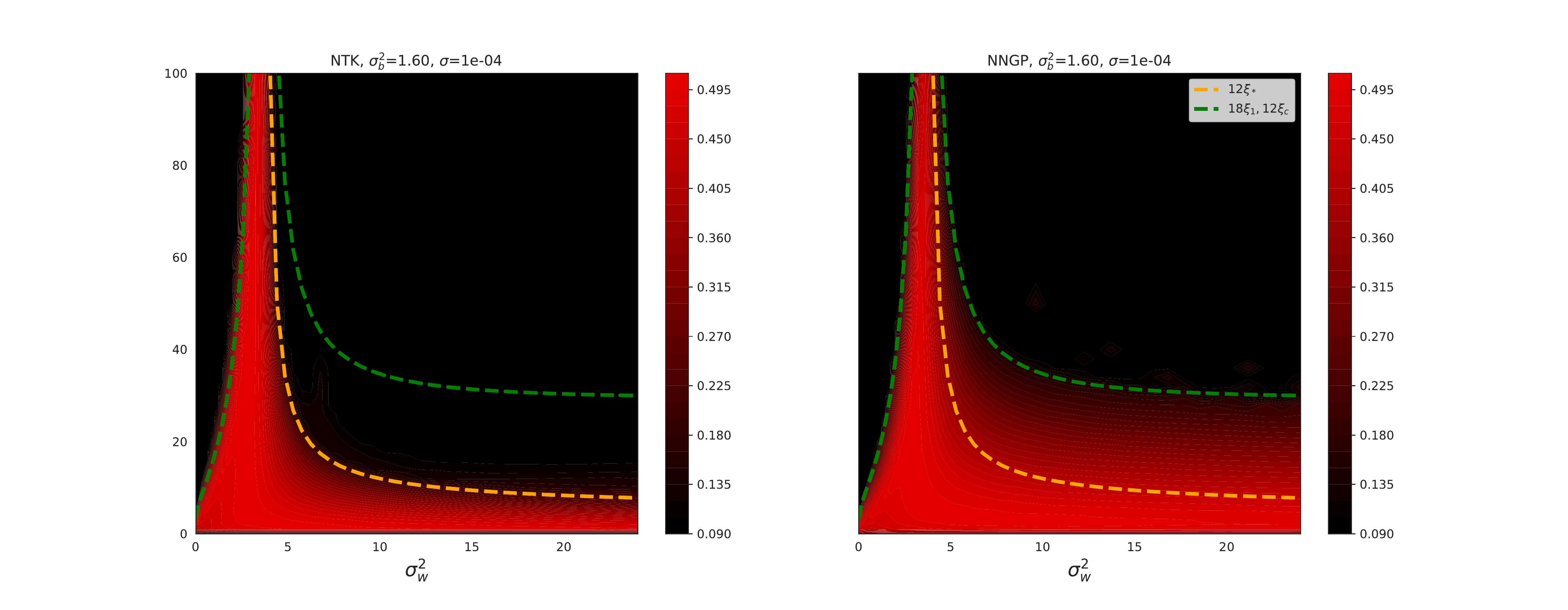}
        \end{subfigure}
        \\
        \begin{subfigure}[b]{0.49\textwidth}
                % \centering
              \includegraphics[width=1.1\textwidth]{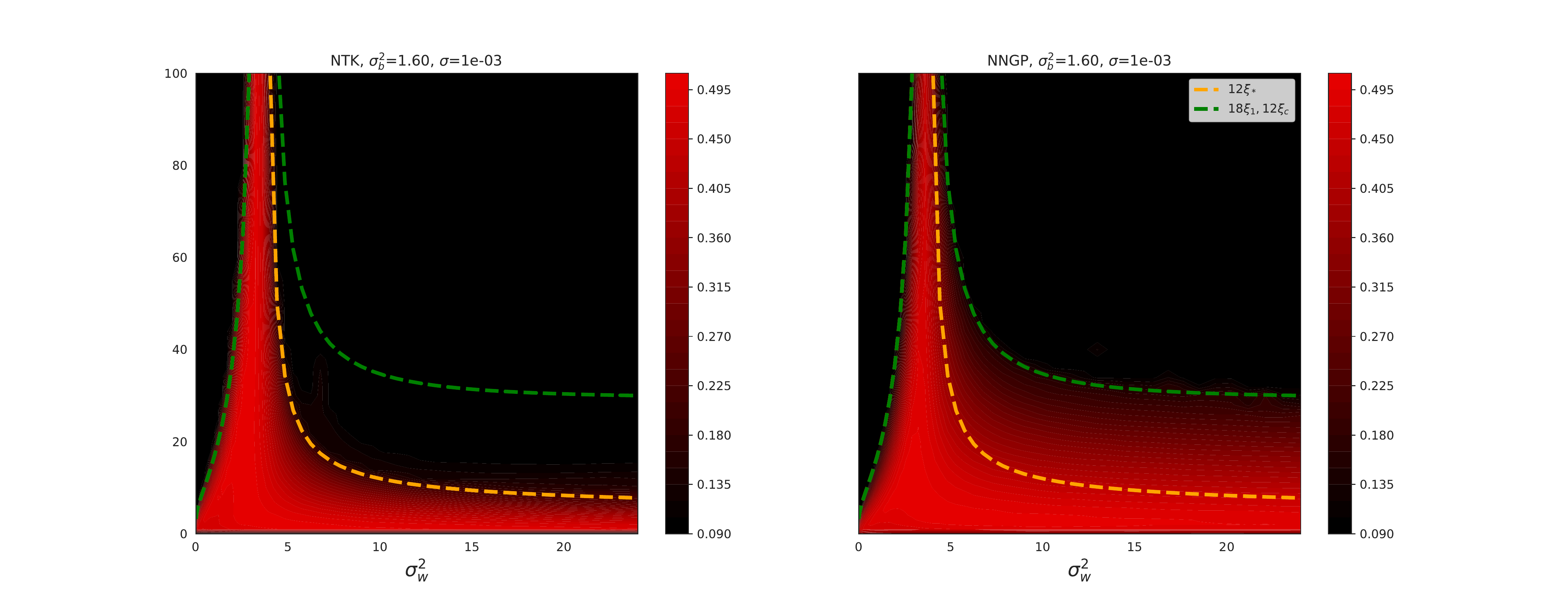}
        \end{subfigure}
 \begin{subfigure}[b]{0.49\textwidth}
                % \centering
              \includegraphics[width=1.1\textwidth]{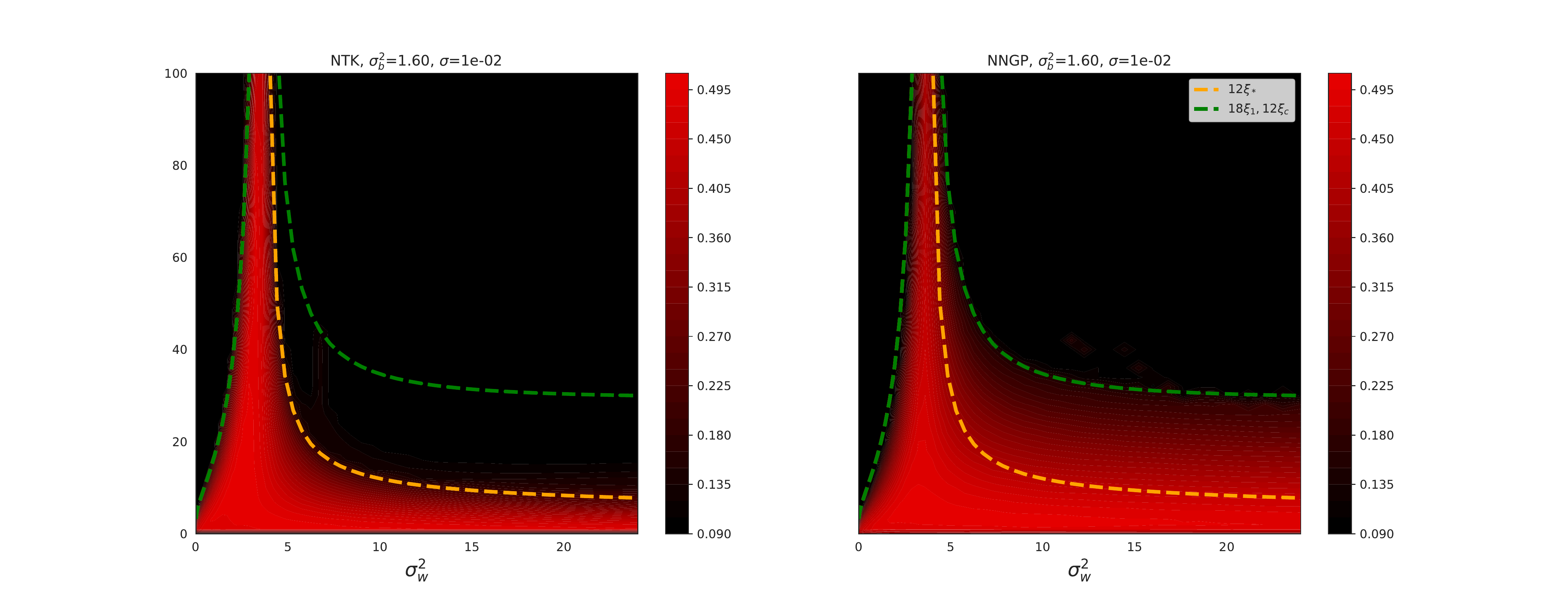}
        \end{subfigure}\\
        \begin{subfigure}[b]{0.49\textwidth}
                % \centering
              \includegraphics[width=1.1\textwidth]{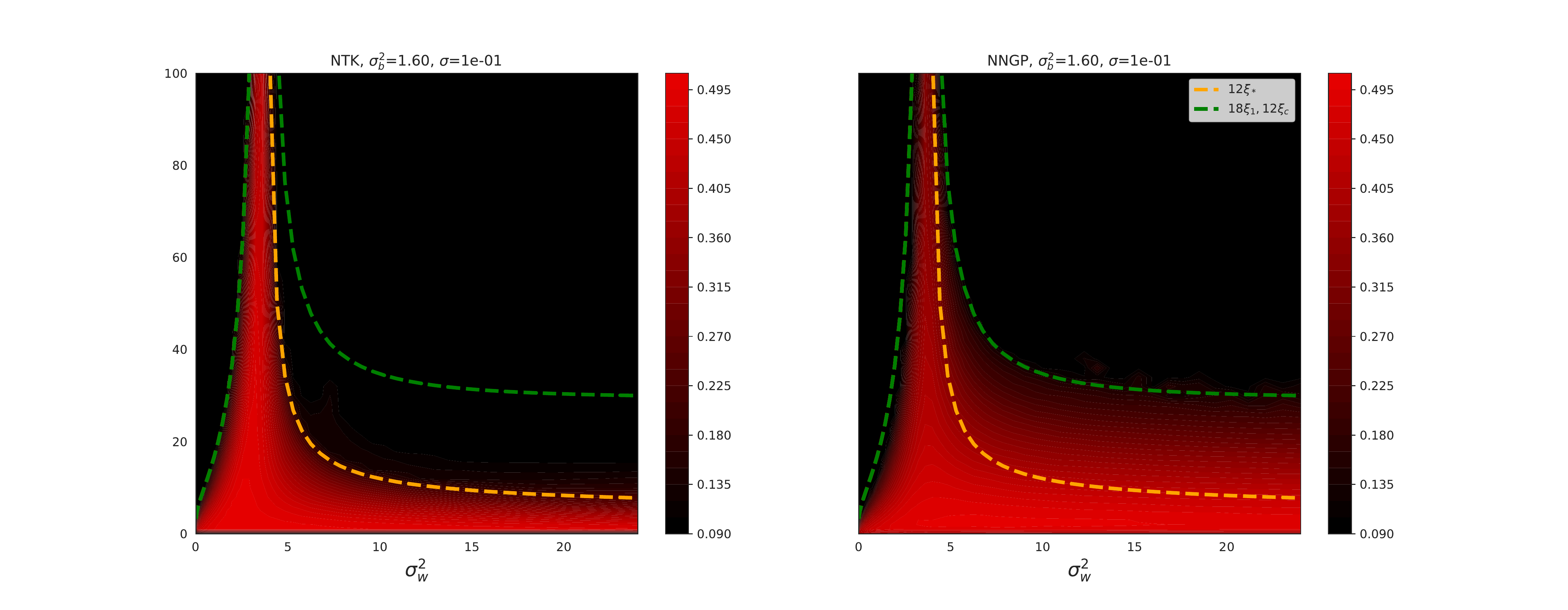}
        \end{subfigure}
\begin{subfigure}[b]{0.49\textwidth}
                % \centering
              \includegraphics[width=1.1\textwidth]{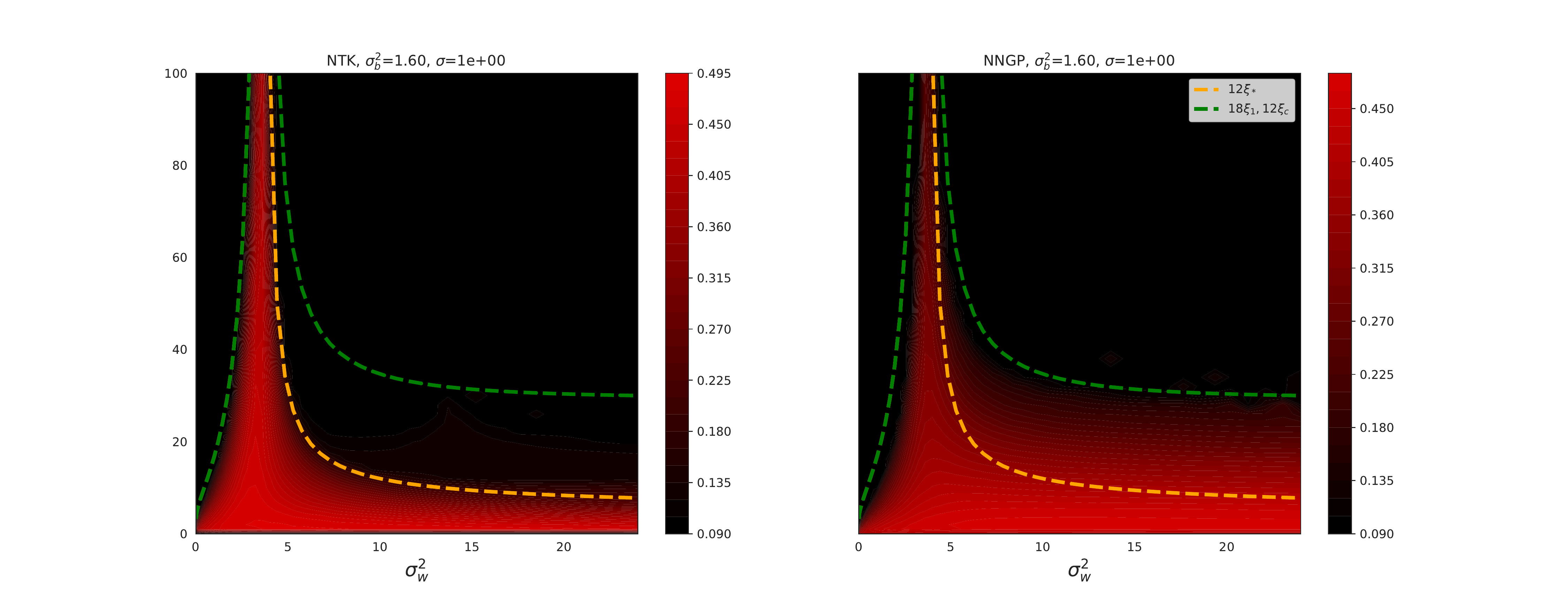}
        \end{subfigure}
\end{center}
\caption{{\bf Generalization metrics for NTK/NNGP vs Test Accuracy vs $\sigma$.} 
}
\label{fig: generalization-varying-sigma}
\end{figure*}

\subsection{Densely Sweeping Over the Regularization Strength $\sigma$: Figure~\ref{fig: generalization-varying-sigma}}
Similar to the above setup, we fixed $\sbs=1.6$ and densely vary $\sigma\in \{0, 10^{-6}, \dots, 10^{0}\}$.
